%% file: main.tex
\documentclass[10pt,journal,compsoc]{IEEEtran}
\ifCLASSOPTIONcompsoc
  \usepackage[nocompress]{cite}
\else
  \usepackage{cite}
\fi
\ifCLASSINFOpdf
  \usepackage[pdftex]{graphicx}
\else
\fi
\usepackage{subfigure}
\usepackage{amsmath}
\usepackage{multirow}
\usepackage{amssymb}  
\newcommand{\etal}{\textit{et al.}}
\usepackage{color}
\newcommand{\REBUT}[1]{\textcolor{black}{#1}}
\newcommand{\GUR}[1]{\textcolor{blue}{#1}}

\usepackage{booktabs} 
\usepackage{hyperref}
\begin{document}
%

\title{ROAD: The ROad event Awareness Dataset for Autonomous Driving}
%
%

\author{Gurkirt~Singh${}^{3}$,
        Stephen~Akrigg${}^{1}$,
        Manuele~Di~Maio${}^5$,
        Valentina~Fontana${}^{2}$, 
        Reza Javanmard Alitappeh${}^{4}$
        Salman Khan${}^{1}$,
        Suman~Saha${}^3$,
        Kossar Jeddisaravi${}^{4}$
        Farzad Yousefi${}^{4}$
        Jacob Culley${}^{1}$,
        Tom Nicholson${}^{1}$,
        Jordan Omokeowa${}^{1}$,
        Stanislao~Grazioso${}^2$,
        Andrew~Bradley${}^1$,
        Giuseppe~Di~Gironimo${}^2$,
        Fabio~Cuzzolin${}^1$
\IEEEcompsocitemizethanks{\IEEEcompsocthanksitem ${}^1$VAIL, Oxford Brookes University, UK, ${}^2$ University of Naples Federico II, Italy, ${}^3$ CVL, ETH Zurich, ${}^4$ University of Science and Technology of Mazandaran, Behshahr, Iran, ${}^5$Siemens SpA, Bologna, Italy.
This collaborative work took place at Oxford Brookes University when V. Fontana and M. Di Maio were visiting students, and S. Akrigg and G. Singh were students there.
E-mail: gurkirt.singh@vision.ee.ethz.ch, fabio.cuzzolin@brookes.ac.uk.}
\thanks{Manuscript received XXX, 2020.}}

\markboth{IEEE Transaction on Pattern Analysis and Machine Intelligence,~Vol.~X, No.~Y, 20XX}%
{Singh \MakeLowercase{\textit{et al.}}: Bare Demo of IEEEtran.cls for Computer Society Journals}

\IEEEtitleabstractindextext{%
\begin{abstract}
\input{text/abstract}

\end{abstract}

\begin{IEEEkeywords}
Autonomous driving, action detection, road agents, situation awareness, decision making.
\end{IEEEkeywords}}

\maketitle

\IEEEdisplaynontitleabstractindextext

\IEEEpeerreviewmaketitle

\input{text/intro}
\input{text/soa}
\input{text/dataset}
\input{text/baseline_methods}

\input{text/experiments}

\input{text/conclusion}

\ifCLASSOPTIONcompsoc
  \section*{Acknowledgments}
\else
  \section*{Acknowledgment}
\fi

This project has received funding from the European Union's Horizon 2020 research and innovation programme, under grant agreement No. 964505 (E-pi).
The authors would like to thank Petar Georgiev, Adrian Scott, Alex Bruce and Arlan Sri Paran for their contribution to video annotation. The project was also partly funded by the Leverhulme Trust under the Research Project Grant RPG-2019-243.
\REBUT{We also wish to acknowledge the members of the ROAD challenge's winning teams: Chenghui Li, Yi Cheng, Shuhan Wang, Zhongjian Huang, Fang Liu, Lijun Yu, Yijun Qian, Xiwen Chen, Wenhe Liu, Alexander G. Hauptmann, Yujie Hou and Fengyan Wang.}




\bibliographystyle{IEEEtran}
\bibliography{ref}


\vspace{-14mm}
\clearpage
\appendices
\input{text/supp}

\end{document}

%% file: text/abstract.tex
Humans drive in a holistic fashion which entails, in particular, understanding dynamic road events and their evolution. Injecting these capabilities in autonomous vehicles can thus take situational awareness and decision making closer to human-level performance. To this purpose, we introduce the ROad event Awareness Dataset (ROAD) for Autonomous Driving, to our knowledge the first of its kind. ROAD is designed to test an autonomous vehicle’s ability to detect road events, defined as triplets composed by an active agent, the action(s) it performs and the corresponding scene locations. ROAD comprises videos originally from the Oxford RobotCar Dataset, annotated with bounding boxes showing the location in the image plane of each road event. We benchmark various detection tasks, proposing as a baseline a new incremental algorithm for online road event awareness termed 3D-RetinaNet.
\REBUT{We also report the performance on the ROAD tasks of Slowfast and YOLOv5 detectors, as well as that of the winners of the ICCV2021 ROAD challenge,} which highlight the challenges faced by situation awareness in autonomous driving. ROAD is designed to allow scholars to investigate exciting tasks such as complex (road) activity detection, future event anticipation and continual learning.
\REBUT{The dataset is available at https://github.com/gurkirt/road-dataset; the baseline can be found at https://github.com/gurkirt/3D-RetinaNet.}

%% file: text/intro.tex
\IEEEraisesectionheading{\section{Introduction}}~\label{sec:intro}

\IEEEPARstart{I}{n} recent years, \textit{autonomous driving} (or \textit{robot-assisted driving}) has emerged as a fast-growing research area. 
\noindent The race towards fully autonomous vehicles pushed many large companies, such as Google, Toyota and Ford, to develop their own concept of \emph{robot-car}~\cite{winn2006layout,KirstenNov2017,gpandey-2011a}. 
While self-driving cars are widely considered to be a  major development and testing ground for the real-world application of artificial intelligence, 
major reasons for concern remain 
in terms of safety, ethics, cost, and reliability~\cite{maurer2016autonomous}.
From a safety standpoint, in particular, smart cars need to robustly interpret the behaviour of the humans (drivers, pedestrians or cyclists) they share the environment with, in order to cope with their decisions.
\emph{Situation awareness} and the ability to 
understand the behaviour of other road users are thus crucial for the safe deployment of autonomous vehicles (AVs).

The latest generation of robot--cars is equipped with a range of different sensors (i.e., laser rangefinders, radar, cameras, GPS) to provide data on what is happening on the road~\cite{broggi2016intelligent}. The information so extracted is then fused to suggest how the vehicle should move~\cite{8500500,8500413,8500556,8500605}.
Some authors, 
however, maintain that vision is a sufficient sense for AVs to navigate their environment, supported by humans' ability to do just so. 
Without enlisting ourselves as supporters of the latter point of view, in this paper we consider the context of \emph{vision-based} autonomous driving~\cite{BERTOZZI20001} from video sequences captured by cameras mounted on the vehicle in a streaming, online fashion. 

While detector networks~\cite{YOLO9000} are routinely trained 
to facilitate object and actor recognition in road scenes, this simply allows the vehicle to 'see' what is around it. 
\REBUT{The philosophy of this work is that robust self-driving capabilities require a deeper, more human-like understanding of dynamic road environments (and of the evolving behaviour of other road users over time) in the form of semantically meaningful concepts, as a stepping stone for intention prediction and automated decision making.}
One advantage of this approach is that it allows the autonomous vehicle to focus on a much smaller amount of relevant information when learning how to make its decisions, in a way arguably closer to how decision making takes place in humans.

\begin{figure*}[ht!] 
  \centering
  \includegraphics[scale=0.8]{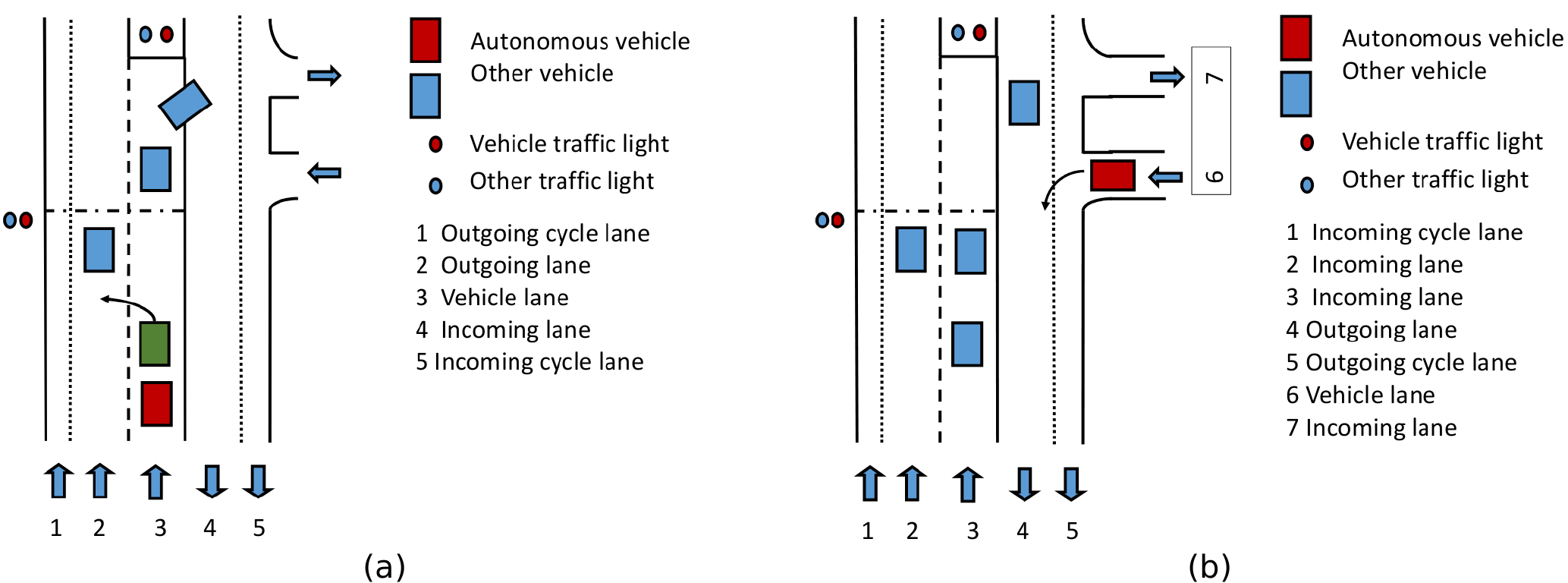}
  \caption{Use of labels in ROAD to describe typical road scenarios. (a) A green car is in front of the AV while changing lanes, as depicted by the arrow symbol. The associated event will then carry the following labels: \textit{in vehicle lane} (location), \textit{moving left} (action). Once the event is completed, the location label will change to: \textit{in outgoing lane}. (b) Autonomous vehicle turning left from lane 6 into lane 4: lane 4 will be the \textit{outgoing lane} as the traffic is moving in the same direction as the AV. However, if the AV turns right from lane 6 into lane 4 (a wrong turn), then lane 4 will become the \textit{incoming lane} as the vehicle will be moving into the incoming traffic. The overall philosophy of ROAD is to use suitable combinations of multiple label types to fully describe a road situation, and allow a machine learning algorithm to learn from this information.}
  \label{fig:typical} 
\end{figure*}

On the opposite side of the spectrum lies 
end-to-end reinforcement learning. There, the behaviour of a human driver in response to road situations is used to train, in an {imitation learning} setting~\cite{codevilla2018end}, an autonomous car to respond in a more ‘human-like’ manner to road scenarios. This, however, requires an astonishing amount of data from 
a myriad of road situations. For highway driving only, a relatively simple task when compared to city driving, Fridman et al. in~\cite{Fridman2017arxiv} had to use a whole fleet of vehicles to collect 45 million frames. 
Perhaps more importantly, 
in this approach the network learns a mapping from the scene to control inputs,
\REBUT{without attempting to model the significant facts taking place in the scene or the reasoning of the agents therein.}
As discussed in~\cite{Cuzzolin2020tom}, many authors~\cite{Rasouli2020,Rudenko2019} have recently highlighted the insufficiency
of models which directly map observations to actions \cite{Armstrong2018}, specifically in the self-driving cars scenario. 

\subsection{ROAD: a multi-label, multi-task dataset}~\label{subsec:intro-road}

\REBUT{\emph{Concept}. This work aims to propose a new framework for situation awareness and perception, 
departing from the disorganised collection of object detection, semantic segmentation or pedestrian intention tasks which is the focus of much current work. 
We propose to do so in a “holistic”, multi-label approach in which agents, actions and their locations are all ingredients in the fundamental concept of \emph{road event} (RE). Road events are defined as triplets $E = (Ag, Ac, Loc)$ composed by an active road agent $Ag$, the action(s) $Ac$ it performs (possibly more than one at the same time), and the location(s) $Loc$ in which this takes place (which may vary from the start to the end of the event itself), as seen from the point of the view of an autonomous vehicle. 
\\
This takes the problem to a higher conceptual level, in which AVs are tested on their \emph{understanding of what is going on} in a dynamic scene rather than their ability to describe what the scene \emph{looks like}, putting them in a position to use that information to make decisions and a plot course of action. Modelling dynamic road scenes in terms of road events can also allow us to model the causal relationships between what happens; these causality links can then be exploited to predict further future consequences.}

\REBUT{To transfer this conceptual paradigm into practice, this paper introduces ROAD, the first \emph{ROad event Awareness in Autonomous Driving Dataset}, as an entirely new type of dataset designed to allow researchers in autonomous vehicles to test the 
situation awareness capabilities of their stacks in a manner impossible until now. Unlike all existing benchmarks, ROAD provides ground truth for the action performed by all road agents, not just humans.
In this sense ROAD 
is unique in the richness and sophistication of its annotation, designed to support the proposed conceptual shift. We are confident this contribution will be very useful moving forward for both the autonomous driving and the computer vision community.}

\REBUT{\emph{Features}.
ROAD is built upon (a fraction of) the Oxford RobotCar Dataset \cite{maddern20171}, by carefully annotating 22 carefully selected, relatively long-duration 
videos. Road events are represented as '{tubes}', i.e., time series of frame-wise bounding box detections.}
ROAD is a dataset of significant size, \REBUT{most notably in terms of the richness and complexity of its annotation rather than the raw number of video frames. A total of} $122K$ video frames are labelled for a total of $560K$ detection bounding boxes in turn associated with $1.7M$ unique individual labels, 
broken down into $560K$ agent labels, $640K$ action labels and $499K$ location labels.

The dataset was designed according to the following principles.
\begin{itemize}
    \item A \emph{multi-label} benchmark: each road event is composed by the label of the (moving) agent responsible, the label(s) of the type of action(s) 
    being performed, and labels describing where the action is located. 
    \item Each event can be assigned \emph{multiple instances} of the same label type whenever relevant (e.g., an RE can be an instance of both \textit{moving away} and \textit{turning left}).
    \item The labelling is done \emph{from the point of view of the AV}: the final goal is for the autonomous vehicle to use this information to make the appropriate decisions.
    \item The meta-data is intended to contain all the information required to fully describe a road scenario: an illustration of this concept is given in Figure \ref{fig:typical}. After closing one's eyes, the set of labels associated with the current video frame should be sufficient to recreate the road situation in one's head (or, equivalently, sufficient for the AV to be able to make a decision).
\end{itemize}

\REBUT{In an effort to take action detection into the real world}, 
ROAD moves away from human body actions almost entirely, to consider (besides pedestrian behaviour) actions performed by humans as drivers of various types of vehicles, shifting the paradigm from \emph{actions performed by human bodies} to \emph{events caused by agents}.
As shown in our experiments, ROAD is more challenging than current action detection benchmarks due to 
the complexity of road events happening in real, non-choreographed driving conditions, the number of active agents present and the variety of weather conditions encompassed.


\REBUT{\emph{Tasks}.}
ROAD allows one to validate manifold tasks associated with situation awareness for self-driving, each associated with a label type (agent, action, location) or combination thereof: \emph{spatiotemporal}
(i) \emph{agent detection},
(ii) \emph{action detection}, 
(iii) \emph{location detection}, 
(iv) \emph{agent-action detection}, 
(v) \emph{road event detection}, as well as the
(vi) \emph{temporal segmentation of AV actions}. 
For each task one can assess both \emph{frame-level} detection, which outputs independently for each video frame the bounding box(es) (BBs) of the instances there present and the relevant class labels, and \emph{video-level} detection, which consists in regressing
the whole series of temporally-linked bounding boxes (i.e., in current terminology, a ’tube’) associated with an instance, together with the relevant class label. In this paper we conduct tests on both.
All tasks come with both the necessary annotation and a shared baseline, 
which is described in Section \ref{sec:baseline-method}. 

\subsection{Contributions}

\REBUT{The major contributions of the paper are thus the following.}
\begin{itemize}
\item
\REBUT{A conceptual shift in situation awareness centred on} a formal definition of the notion of road event, as a triplet composed by a road agent, the action(s) it performs and the location(s) of the event, seen from the point of view of the AV.
\item A new ROad event Awareness Dataset for Autonomous Driving (ROAD), the first of its kind, designed to \REBUT{support this paradigm shift and} allow the testing of a range of tasks related to situation awareness for autonomous driving: agent and/or action detection, event detection, ego-action classification.
\end{itemize}
\REBUT{Instrumental to the introduction of ROAD as the benchmark of choice for semantic situation awareness, we propose a robust baseline for online action/agent/event detection (termed \emph{3D-RetinaNet}) which combines state-of-the-art single-stage object detector technology with an online tube construction method~\cite{singh2017online}, with the aim of linking detections over time to create \emph{event tubes}~\cite{saha2016deep,Georgia-2015a}. Results for two additional baselines based on a Slowfast detector architecture \cite{feichtenhofer2019slowfast} and YOLOv5\footnote{\url{https://github.com/ultralytics/yolov5}.} (for agent detection only) are also reported and critically assessed.}

We are confident that this work will lay the foundations upon which much further research in this area can be built.

\subsection{Outline}

The remainder of the paper is organised as follows. Section \ref{sec:related-work} reviews related work concerning existing datasets, both for autonomous driving (Sec. \ref{subsec:av-datasets}) and action detection (Sec. \ref{sec:soa-detection-datasets}), as well as action detection methods (Sec. \ref{sec:soa-detection-methods}).
Section \ref{sec:road-dataset} presents our ROAD dataset in full detail, including: its multi-label nature (Sec. \ref{sec:dataset-multilabel}), data collection (Sec. \ref{sec:dataset-collection}), annotation (Sec. \ref{subsec:dataset-annotation}), the tasks it is designed to validate (Sec. \ref{sec:dataset-tasks}), and a quantitative summary (Sec. \ref{sec:dataset-summary}). 
Section \ref{sec:baseline-method} \REBUT{presents an overview of the proposed 3D-RetinaNet baseline, and recalls the ROAD challenge organised by some of us at ICCV 2021 to disseminate this new approach to situation awareness within the autonomous driving and computer vision communities, using ROAD as the benchmark.} 
Experiments are described in Section \ref{sec:experiments}, 
where \REBUT{a number of ablation studies} are reported and critically analysed in detail, \REBUT{together with the results of the ROAD challenge's top participants}. 
Section \ref{sec:extensions} outlines additional exciting tasks the dataset can be used as a benchmark for in the near future, \REBUT{such as future event anticipation, decision making and machine theory of mind \cite{Cuzzolin2020tom}}.
Conclusions and future work are outlined in Section \ref{sec:conclusion}.

\REBUT{The Supplementary material reports detailed class-wise results, a qualitative analysis of success and failure cases, and a link to a 30-minute footage visually illustrating the baseline's predictions versus the ground truth.}

%% file: text/soa.tex
\section{RELATED WORK} \label{sec:related-work}

\subsection{Autonomous driving datasets} \label{subsec:av-datasets}

In recent years a multitude of  AV datasets have been released, 
mostly focusing on object detection and scene segmentation.
We can categorise them into two main bins: (1) RGB without range data (single modality) and (2) RGB with range data (multimodal).

\emph{Single-modality datasets}.
Collecting and annotating RGB data only is relatively less time-consuming and expensive than building multimodal datasets including range data from LiDAR or radar. 
Most single-modality datasets
\cite{brostow2008segmentation, cordts2016cityscapes, neuhold2017mapillary, yu2018bdd100k, wang2019apolloscape,che2019d}
provide 2D bounding box and scene segmentation labels for RGB images. {Examples include Cityscapes \cite{cordts2016cityscapes}, Mapillary Vistas \cite{neuhold2017mapillary},
BDD100k \cite{yu2018bdd100k} and Apolloscape \cite{wang2019apolloscape}.}
To allow the studying of how vision algorithms generalise to different unseen data, 
\cite{neuhold2017mapillary,che2019d,yu2018bdd100k}
collect RGB images under different illumination and weather conditions.
Other datasets only provide pedestrian detection annotation  
\cite{dalal2005histograms, ess2008mobile, wojek2009multi, enzweiler2008monocular, zhang2017citypersons, neumann2018nightowls, covello1993evaluation}.
Recently, MIT and Toyota have released DriveSeg, which comes with pixel-level semantic labelling for 12 agent classes \cite{mmke-dv03-20}.

\emph{Multimodal datasets}.
KITTI \cite{geiger2012we} was the first-ever multimodal dataset. It provides depth labels from front-facing stereo images and dense point clouds from LiDAR alongside GPS/IMU (inertial) data.
It also provides bounding-box annotations to facilitate improvements in 3D object detection. 
H3D \cite{patil2019h3d} and KAIST \cite{choi2018kaist} are two more examples of multimodal datasets. H3D provides 3D box annotations, using real-world LiDAR-generated 3D coordinates, in crowded scenes.
Unlike KITTI, H3D comes with object detection annotations in a full 360$^o$ view.  
KAIST provides thermal camera data alongside RGB, stereo, GPS/IMU and LiDAR-based range data.
Among other notable multimodal datasets \cite{blanco2014malaga, maddern20171}
only consist of raw data without semantic labels, 
whereas \cite{jung2016multi} and \cite{chen2018lidar} provide labels for location category and driving behaviour, respectively. 
The most recent multimodal large-scale AV datasets
\cite{chang2019argoverse, kesten2019lyft, sun2019scalability, pham20193d, geyer2019a2d2, Liao2021ARXIV}
are significantly larger in terms of both data (also captured under varying weather conditions, e.g. by night or in the rain) 
and annotations (RGB, LiDAR/radar, 3D boxes). 
For instance, Argovers \cite{chang2019argoverse} doubles the number of sensors in comparison to KITTI \cite{geiger2012we} and nuScenes \cite{caesar2020nuscenes}, providing 3D bounding boxes with tracking information for 15 objects of interest. 
Similarly, Lyft \cite{kesten2019lyft} provides 3D bounding boxes for cars and location annotation including lane segments, pedestrian crosswalks, stop signs, parking zones, speed bumps, and speed humps. \REBUT{In a setup similar to KITTI's \cite{geiger2012we}, in KITTI-360 \cite{Liao2021ARXIV} two fisheye cameras and a pushbroom laser scanner are added to have a full $360{^o}$ field of view. KITTI-360 contains semantic and instance annotations for both 3D point clouds and 2D images, which include 19 objects. IMU/GPS sensors are added for localisation purposes.} Both 3D bounding boxes based on LiDAR data and 2D annotation on camera data for 4 objects classes are provided in Waymo \cite{sun2019scalability}. 
In \cite{pham20193d}, using
similar 3D annotation for 5 objects classes, the authors provide a more challenging dataset by adding more night-time scenarios using a faster-moving car.
Amongst large-scale multimodal datasets, nuScenes \cite{caesar2020nuscenes}, Lyft L5 \cite{kesten2019lyft}, Waymo Open \cite{sun2019scalability} and
A*3D \cite{pham20193d} are the most dominant ones in terms of number of instances, the use of high-quality sensors with different types of data (e.g., point clouds or 360$^{\circ}$ RGB videos), and richness of the annotation providing both semantic information and 3D bounding boxes.
Furthermore, nuScenes \cite{caesar2020nuscenes}, Argoverse \cite{chang2019argoverse}
Lyft L5 \cite{kesten2019lyft} and KITTI-360~\cite{Liao2021ARXIV} provide contextual knowledge through human-annotated rich semantic maps, an important prior for scene understanding.

\emph{Trajectory prediction}.
Another line of work considers the problem of pedestrian trajectory prediction in the autonomous driving setting, and rests on several influential RGB-based datasets.
To compile these datasets, RGB data were captured using either stationary surveillance cameras 
\cite{lerner2007crowds, pellegrini2010improving, oh2011large} or drone-mounted ones \cite{robicquet2016learning} for aerial view.
\cite{yao2019egocentric, chandra2019traphic} use RGB images capturing an egocentric view from a moving car for future trajectory forecasting. 
Recently, the multimodal 3D point cloud-based datasets  
\cite{geiger2012we, patil2019h3d, kesten2019lyft, caesar2020nuscenes, sun2019scalability, chang2019argoverse},
initially introduced for the benchmarking of 3D object detection and tracking, have been taken up for trajectory prediction as well. A host of interesting recent papers
\cite{rasouli2017they, minguez2018pedestrian, rasouli2019pie,malla2020titan}
do propose datasets to study the intentions and actions of agents using cameras mounted on vehicles.
However, they encompass a limited set of action labels (e.g. walking, standing, looking or crossing), wholly insufficient for a thorough study of road agent behaviour.
Among them, TITAN~\cite{malla2020titan} is arguably the most promising. 
\\
Our ROAD dataset is similar to TITAN in the sense that both consider actions performed by humans present in the road scene and provide spatiotemporal localisation for each person using multiple action labels. \REBUT{However, TITAN's action labels are restricted to humans (pedestrians), rather than extending to all road agents (with the exception of vehicles with `stopped' and `moving' actions). The dataset is a collection of much shorter videos which only last 10-20 seconds, and does not not contemplate agent location (a crucial source of information). Finally, the size of its vocabulary in terms of number of agents and actions is much smaller (see Table \ref{tab:datasets}).}

\begin{table}[t]
	\centering
	\caption{\REBUT{ Comparison of ROAD with similar datasets for perception in autonomous driving in terms of diversity of labels. The comparison is based on the number of classes portrayed and the availability of action annotations and tube tracks for both pedestrians and vehicles, as well as location information. Most competitor datasets do not provide action annotation for either pedestrians or vehicles.}}
	\label{tab:datasets}
	\setlength{\tabcolsep}{4pt}
	    {\footnotesize
		\scalebox{0.80}{
\begin{tabular}{ccccccc}
	\toprule
	\multicolumn{1}{c}{\multirow{2}{*}{\textbf{Dataset}}} & \multicolumn{1}{c}{\multirow{2}{*}{\textbf{Class Num.}}} & \multicolumn{1}{c}{\multirow{2}{*}{\textbf{Location label}}} & \multicolumn{2}{c}{\textbf{Action Ann}}            & \multicolumn{2}{c}{\textbf{Tube Ann}}              \\ \cline{4-7} 
	\multicolumn{1}{c}{} & \multicolumn{1}{c}{} & \multicolumn{1}{c}{} & \multicolumn{1}{c}{Ped.} & \multicolumn{1}{c}{Veh.} & \multicolumn{1}{c}{Ped.} & \multicolumn{1}{c}{Veh.} \\ \hline
	\multicolumn{1}{c}{SYNTHIA\cite{ros2016synthia}} & \multicolumn{1}{c}{13}& \multicolumn{1}{c}{pixelwise ann.} & \multicolumn{1}{c}{-}& \multicolumn{1}{c}{-}& \multicolumn{1}{c}{-}& \multicolumn{1}{c}{-}\\ 
	\multicolumn{1}{c}{SemKITTI \cite{behley2019semantickitti}} & \multicolumn{1}{c}{28}& \multicolumn{1}{c}{3D sem. seg.} & \multicolumn{1}{c}{-}& \multicolumn{1}{c}{-}& \multicolumn{1}{c}{-}& \multicolumn{1}{c}{-}\\ 
	\multicolumn{1}{c}{Cityscapes \cite{cordts2016cityscapes}} & \multicolumn{1}{c}{30}& \multicolumn{1}{c}{pixel level sem.} & \multicolumn{1}{c}{-}& \multicolumn{1}{c}{-}& \multicolumn{1}{c}{-}& \multicolumn{1}{c}{-}\\ 
	\multicolumn{1}{c}{ A2D2\cite{geyer2019a2d2}} & \multicolumn{1}{c}{14}& \multicolumn{1}{c}{3D sem. seg.} & \multicolumn{1}{c}{-}& \multicolumn{1}{c}{-}& \multicolumn{1}{c}{-}& \multicolumn{1}{c}{-}\\ 
	\multicolumn{1}{c}{ Waymo \cite{sun2019scalability}} & \multicolumn{1}{c}{4}& \multicolumn{1}{c}{-} & \multicolumn{1}{c}{-}& \multicolumn{1}{c}{-}& \multicolumn{1}{c}{\checkmark}& \multicolumn{1}{c}{\checkmark}\\ 
	\multicolumn{1}{c}{ Apolloscape \cite{wang2019apolloscape}} & \multicolumn{1}{c}{25}& \multicolumn{1}{c}{pixel level sem.} & \multicolumn{1}{c}{-}& \multicolumn{1}{c}{-}& \multicolumn{1}{c}{\checkmark}& \multicolumn{1}{c}{\checkmark}\\ 
	\multicolumn{1}{c}{ PIE \cite{rasouli2019pie}} & \multicolumn{1}{c}{6}& \multicolumn{1}{c}{-} & \multicolumn{1}{c}{\checkmark}& \multicolumn{1}{c}{-}& \multicolumn{1}{c}{\checkmark}& \multicolumn{1}{c}{-}\\ 
	\multicolumn{1}{c}{ TITAN \cite{malla2020titan}} & \multicolumn{1}{c}{50}& \multicolumn{1}{c}{-} & \multicolumn{1}{c}{\checkmark}& \multicolumn{1}{c}{\checkmark}& \multicolumn{1}{c}{\checkmark}& \multicolumn{1}{c}{\checkmark}\\ 
	\multicolumn{1}{c}{ KITTI360 \cite{Liao2021ARXIV}} & \multicolumn{1}{c}{19}& \multicolumn{1}{c}{sem. ann.} & \multicolumn{1}{c}{-}& \multicolumn{1}{c}{-}& \multicolumn{1}{c}{-}& \multicolumn{1}{c}{-}\\ 
	\multicolumn{1}{c}{  A*3D \cite{pham20193d}} & \multicolumn{1}{c}{7}& \multicolumn{1}{c}{-} & \multicolumn{1}{c}{-}& \multicolumn{1}{c}{-}& \multicolumn{1}{c}{-}& \multicolumn{1}{c}{-}\\ 
	\multicolumn{1}{c}{  H3D \cite{patil2019h3d}} & \multicolumn{1}{c}{8}& \multicolumn{1}{c}{-} & \multicolumn{1}{c}{-}& \multicolumn{1}{c}{-}& \multicolumn{1}{c}{\checkmark}& \multicolumn{1}{c}{\checkmark}\\ 
	\multicolumn{1}{c}{  Argoverse \cite{chang2019argoverse}} & \multicolumn{1}{c}{15}& \multicolumn{1}{c}{-} & \multicolumn{1}{c}{-}& \multicolumn{1}{c}{-}& \multicolumn{1}{c}{\checkmark}& \multicolumn{1}{c}{\checkmark}\\ 
	\multicolumn{1}{c}{ NuScense \cite{caesar2020nuscenes}} & \multicolumn{1}{c}{23}& \multicolumn{1}{c}{3D sem. seg.} & \multicolumn{1}{c}{-}& \multicolumn{1}{c}{-}& \multicolumn{1}{c}{\checkmark}& \multicolumn{1}{c}{\checkmark}\\ 
	\multicolumn{1}{c}{ DriveSeg \cite{mmke-dv03-20}} & \multicolumn{1}{c}{12}& \multicolumn{1}{c}{sem. ann.} & \multicolumn{1}{c}{-}& \multicolumn{1}{c}{-}& \multicolumn{1}{c}{-}& \multicolumn{1}{c}{-}\\ 
	\midrule
	\multicolumn{5}{c}{Spatiotemporal action detection datasets} & \\
	\midrule
	\multicolumn{1}{c}{ UCF24 \cite{jiang2014thumos}} & \multicolumn{1}{c}{24}& \multicolumn{1}{c}{-} & \multicolumn{1}{c}{\checkmark}& \multicolumn{1}{c}{-}& \multicolumn{1}{c}{\checkmark}& \multicolumn{1}{c}{-}\\ 
	\multicolumn{1}{c}{ AVA \cite{ava2017gu}} & \multicolumn{1}{c}{80}& \multicolumn{1}{c}{-} & \multicolumn{1}{c}{\checkmark}& \multicolumn{1}{c}{-}& \multicolumn{1}{c}{\checkmark}& \multicolumn{1}{c}{-}\\ 
	\multicolumn{1}{c}{ Multisports \cite{li2021multisports}} & \multicolumn{1}{c}{66} & \multicolumn{1}{c}{-} & \multicolumn{1}{c}{\checkmark}& \multicolumn{1}{c}{-}& \multicolumn{1}{c}{\checkmark}& \multicolumn{1}{c}{-}\\ 
	\midrule
	\multicolumn{1}{c}{ \textbf{ROAD} (ours)} & \multicolumn{1}{c}{$43$}& \multicolumn{1}{c}{\checkmark} & \multicolumn{1}{c}{\checkmark}& \multicolumn{1}{c}{\checkmark}& \multicolumn{1}{c}{\checkmark}& \multicolumn{1}{c}{\checkmark}\\ 
	\bottomrule 
	\multicolumn{7}{l}{Ped. Pedestrian, Veh. Vehicle, ann. annotation, sem. seg. semantic segmentation} \\
\end{tabular}
}
}
\label{table:ROADtubecomparison}
\end{table}

%
	
\REBUT{As mentioned, our ROAD dataset is built upon the multimodal Oxford RobotCar dataset, which contains both visual and 3D point cloud data. Here, however, we only process a number of its videos to describe and annotate road events. Note that it is indeed possible to map the 3D point clouds from RobotCar’s LiDAR data onto the 2D images to enable true multi-modal action detection. However, a considerable amount would be required to do this, and will be considered in future extensions.}

\REBUT{ROAD departs substantially from all previous efforts, as: (1) it is designed to formally introduce} the notion of \emph{road event} as a combination of three semantically-meaningful labels such as agent, action and location; (2) it provides both bounding-box-level \emph{and} tube-level annotation (to validate methods that exploit the dynamics of motion patterns) on long-duration videos (thus laying the foundations for future work on event anticipation and continual learning); (3) it provides temporally dense annotation; (4) it labels the actions not only of physical humans but also of other relevant road agents such as vehicles of different kinds.

\REBUT{Table \ref{table:ROADtubecomparison} compares our ROAD dataset with the other state-of-the-art datasets in perception for autonomous driving, in terms of the number and type of labels. As it can be noted in the table, the unique feature of ROAD is its diversity in terms of the types of actions and events portrayed, for all types of road agents in the scene. With 12 agent classes, 30 action classes and 15 location classes ROAD provides (through a combination of these three elements) a much more refined description of road scenes.
}

\subsection{Action detection datasets} \label{sec:soa-detection-datasets}

Providing annotation for action detection datasets is a painstaking process. {Specifically, the requirement to track actors through the temporal domain makes the manual labelling of a dataset an extremely time consuming exercise, requiring frame-by-frame annotation.} As a result, action detection benchmarks are fewer and smaller than, say, image classification, action recognition or object detection datasets.

Action recognition research can aim for robustness thanks to the availability of truly large scale datasets such as Kinetics~\cite{kay2017kinetics}, Moments\cite{monfortmoments} and others, which are the de-facto benchmarks in this area. The recent 'something-something' video database focuses on more complex actions performed by humans using everyday objects \cite{goyal2017something}, exploring a fine-grained list of 174 actions.
More recently, temporal activity detection datasets like ActivityNet~\cite{caba2015activitynet} and Charades~\cite{sigurdsson2018charadesego} have come to the fore. 
Whereas the latter still do not address the spatiotemporal nature of the action detection problem, however, datasets such as J-HMDB-21 \cite{J-HMDB-Jhuang-2013}, UCF24 \cite{soomro2012ucf101}, LIRIS-HARL \cite{liris-harl-2012}, DALY \cite{daly2016weinzaepfel} or the more recent AVA \cite{ava2017gu} have been designed to provide spatial and temporal annotations for human action detection. 
\\
In fact, most action detection papers are validated on the rather dated and small LIRIS-HARL\cite{liris-harl-2012}, J-HMDB-21~\cite{J-HMDB-Jhuang-2013}, and UCF24~\cite{soomro2012ucf101}, whose level of challenge in terms of presence of different source domains and nuisance factors is quite limited. Although recent additions such as DALY~\cite{daly2016weinzaepfel} and AVA~\cite{ava2017gu} have somewhat improved the situation in terms of variability and number of instances labelled, the realistic validation of action detection methods is still an outstanding issue.
AVA is currently the biggest action detection dataset with $1.6M$ label instances, but it is annotated rather sparsely (at a rate of one frame per second).

Overall, the main objective of these datasets is to validate the localisation of human actions in short, untrimmed videos. ROAD, in opposition, goes beyond the detection of actions performed by physical humans to extend the notion of other forms of intelligent agents (e.g., human- or AI-driven vehicles on the road). 
Furthermore, in contrast with the short clips considered in, e.g., J-HMDB-21 and UCF24, our new dataset is composed of 22 very long videos (around 8 minutes each), thus stressing the dynamical aspect of events and the relationship between distinct but correlated events. 
Crucially, it is geared towards online detection rather than traditional offline detection, as these videos are streamed in using a vehicle-mounted camera. 

\subsection{Online action detection} \label{sec:soa-detection-methods}

We believe advances in the field of human action recognition\cite{carreira2017quo,nonlocal2018wang,feichtenhofer2019slowfast,feichtenhofer2020x3d} can be useful when devising a general approach to the situation awareness problem. 
We are particularly interested in the \emph{action detection} problem~\cite{Georgia-2015a,ava2017gu,girdhar2018better,wu2019long}, in particular \emph{online} action detection~\cite{singh2017online}, given the incremental processing needs of an autonomous vehicle.
Recent work in this area \cite{soomro2016predicting,singh2017online,behl2017incremental,kalogeiton2017action,li2020actionsas,yang2019step}  demonstrates very competitive performance compared to (generally more accurate) offline action detection methods~\cite{ava2017gu,nonlocal2018wang,saha2016deep,saha2017amtnet,peng2016eccv,zhao2019dance,singh2018tramnet,singh2018predicting} on UCF-101-24~\cite{soomro2012ucf101}. 
As mentioned, UCF-101-24 is the main benchmark for online action detection research, as it provides annotation in the form of action tubes and every single frame of the untrimmed videos in it is annotated (unlike AVA~\cite{ava2017gu}, in which videos are only annotated at one frame per second).

A short review of the state-of-the-art in online action detection is in place. 
Singh~\etal~\cite{singh2017online}'s method was perhaps the first to propose an online, real-time solution to action detection in untrimmed videos, validated on UCF-101-24, and based on an innovative incremental tube construction method.
Since then, many other papers~\cite{kalogeiton2017action,li2020actionsas,singh2018tramnet} have made use of the online tube-construction method in~\cite{singh2017online}. 
A common trait of many recent online action detection methods is the reliance on 'tubelet'~\cite{saha2017amtnet,kalogeiton2017action,li2020actionsas} predictions from a stack of frames. This, however, leads to processing delays proportional to the number of frames in the stack, 
making these methods not quite applicable in pure online settings. In the case of~\cite{saha2017amtnet,kalogeiton2017action,li2020actionsas} the frame stack is usually 6-8 frames long, leading to a latency of more than half a second. 

For these reasons, inspired by the frame-wise (2D) nature of~\cite{singh2017online} and the success of the latest single-stage object detectors (such as RetinaNet~\cite{lin2017focal}), \REBUT{here} we propose a simple extension of \cite{singh2017online} \REBUT{termed '3D-RetinaNet' as a baseline algorithm for ROAD tasks}. The latter is completely online when using a 2D backbone network. One, however, can also insert a 3D backbone to make it even more accurate, while keeping the prediction heads online. 
We benchmark our proposed 3D-RetinaNet architecture against the above-mentioned online and offline action detection methods on the UCF-101-24 dataset to show its effectiveness, twinned with its simplicity and efficiency. \REBUT{We also compare it on our new ROAD dataset against the state-of-the-art action detection Slowfast~\cite{feichtenhofer2019slowfast} network. We omit, however, to reproduce other state-of-the-art action detectors such as \cite{pan2021actor} and \cite{tang2020asynchronous}, for \cite{pan2021actor} is affected by instability at training time which makes it difficult to reproduce its results, whereas~\cite{tang2020asynchronous} is too complicated to be suitable as a baseline because of its sparse tracking and memory banks features. Nevertheless, both methods rely on the Slowfast detector as a backbone and baseline action detector.}

%% file: text/dataset.tex
\section{THE DATASET} \label{sec:road-dataset}

\subsection{A multi-label benchmark} \label{sec:dataset-multilabel}

The ROAD dataset is specially designed from the perspective of self-driving cars, and thus includes actions performed not just by humans but by all road agents in specific locations, to form \emph{road events} (REs). 
REs are annotated by drawing a bounding box around each active road agent present in the scene, and linking these bounding boxes over time to form 'tubes'. 
As explained, to this purpose three different types of labels are introduced, namely: 
(i) the category of \emph{road agent} involved (e.g. \textit{Pedestrian}, \textit{Car}, \textit{Bus}, \textit{Cyclist}); 
(ii) the \emph{type of action} being performed by the agent (e.g. \textit{Moving away}, \textit{Moving towards}, \textit{Crossing} and so on), and
(iii) the \emph{location} of the road user relative the autonomous vehicle perceiving the scene (e.g. \textit{In vehicle lane}, \textit{On right pavement}, \textit{In incoming lane}).
In addition, ROAD labels the actions performed by the vehicle itself. Multiple agents might be present at any given time, and each of them may perform multiple actions simultaneously (e.g. a \textit{Car} may be \textit{Indicating right} while \textit{Turning right}). Each agent is always associated with at least one action label. 

\REBUT{The full lists of agent, action and location labels are given in the Supplementary material, Tables 1, 2, 3 and 4}.

\emph{Agent labels}. 
Within a road scene, the objects or people able to perform actions which can influence the decision made by the autonomous vehicle are termed \textit{agents}.
We only annotate \emph{active} agents (i.e., a parked vehicle or a bike or a person visible to the AV but located away from the road are not considered to be 'active' agents).
Three types of agent are considered to be of interest, in the sense defined above, to the autonomous vehicle: people, vehicles and traffic lights.
For simplicity, the AV itself is considered just like another agent: this is done by labelling the vehicle's bonnet.
People are further subdivided into two sub-classes: pedestrians and cyclists.
\REBUT{The vehicle category is subdivided into six sub--classes: car, small--size motorised vehicle, medium--size motorised vehicle, large--size motorised vehicle, bus, motorbike, emergency vehicle.
Finally, the `traffic lights' category is divided into two sub--classes: \textit{Vehicle traffic light}
(if they apply to the AV) and \textit{Other traffic light} (if they apply to other road users). }
Only one agent label can be assigned to each active agent present in the scene at any given time.

\emph{Action labels}. 
Each {agent} can perform one or more \textit{actions} at any given time instant. For example, a traffic light can only carry out a single action: it can be either red, amber, green or `black'. A car, instead, can be associated with two action labels simultaneously, e.g., \textit{Turning right} and \textit{Indicating right}. Although some road agents are inherently multitasking, some action combinations can be suitably described by a single label: for example, pushing an object (e.g. a pushchair or a trolley-bag) while walking can be simply labelled as \textit{Pushing object}. The latter was our choice. 

\emph{AV own actions}. 
Each video frame is also labelled with the action label associated with what the AV is doing. To this end, a bounding box is drawn on the bonnet of the AV. The AV can be assigned one of the following seven action labels: \textit{AV-move}, \textit{AV-stop}, \textit{AV-turn-left}, \textit{AV-turn-right}, \textit{AV-overtake}, \textit{AV-move-left} and \textit{AV-move-right}. 
\REBUT{The full list of AV own action classes is given in the Supplementary material, Table 4.} 
Note that these are separate classes only applicable to the AV, with a different semantics than the similar-sounding classes. 
For instance,
the regular \textit{Moving} action label means 'moving in the perpendicular direction to the AV', whereas \textit{AV-move} means that the AV is on the move along its normal direction of travel. 
These labels mirror those used for the autonomous vehicle in the Honda Research Institute Driving Dataset (HDD) \cite{Ramanishka2018}. 

\emph{Location labels}. 
Agent \textit{location} is crucial for deciding what action the AV should take next. 
As the final, long-term objective of this project is to assist autonomous decision making, we propose to label the location of each agent from the perspective of the autonomous vehicle. For example, a pedestrian can be found on the right or the left pavement, in the vehicle's own lane, while crossing or at a bus stop. The same applies to other agents and vehicles as well. 
There is no location label for the traffic lights as they are not movable objects, but agents of a static nature and well-defined location.
To understand this concept, Fig. \ref{fig:typical} illustrates two scenarios in which the location of the other vehicles sharing the road is depicted from the point of view of the AV. 
\REBUT{\emph{Traffic light} is the only agent type missing location labels, all the other agent classes are associated with at least one location label. A complete table with location classes and their description is provided in Supplementary material.}

\subsection{Data collection} \label{sec:dataset-collection}

ROAD is composed of 22 videos from the publicly available Oxford RobotCar Dataset~\cite{maddern20171} (OxRD) released in 2017 by the Oxford Robotics Institute\footnote{\url{http://robotcar-dataset.robots.ox.ac.uk/}}, covering diverse road scenes under various weather conditions. The OxRD dataset, collected from the narrow streets of the historic city of Oxford, was selected because it presents challenging scenarios for an autonomous vehicle due to the diversity and density of various road users and road events. \REBUT{The OxRD dataset was gathered using 6 cameras, as well as LIDAR (Light Detection and Ranging), GPS (Global Positioning System) and INS (Inertial Navigation System) sensors mounted on a Nissan LEAF vehicle~\cite{maddern20171}. To construct ROAD we only annotated videos from the frontal camera view.}

Note, however, that our labelling process (described below) is not limited to OxRD. In principle, other autonomous vehicle datasets (e.g.~\cite{yu2018bdd100k,geiger2013vision}) may be labelled in the same manner to further enrich the ROAD benchmark,: 
we plan to do exactly so in the near future.

\emph{Video selection}. 
Within OxRD, videos were selected with the objective of ensuring diversity in terms of weather conditions, times of the day and types of scenes recorded. 
Specifically, the 22 videos have been recorded both during the day (in strong sunshine, rain or overcast conditions, sometimes with snow present on the surface) and at night. 
Only a subset of the large number of videos available in OxRD was selected. The presence of semantically meaningful content was the main selection criterion. This was done by manually inspecting the videos in order to cover all types of labels and label classes and to avoid 'deserted' scenarios as much as possible. {Each of the 22 videos is 8 minutes and 20 seconds long, \REBUT{barring three videos whose duration is 6:34, 4:10 and 1:37}, respectively. In total, ROAD comprises 170 minutes of video content.} 

\emph{Preprocessing}. 
Some preprocessing was conducted.
First, the original sets of video frames were downloaded and demosaiced, in order to convert them to red, green, and blue (RGB) image sequences. Then, they were encoded into proper video sequences using \texttt{ffmpeg}\footnote{\href{https://www.ffmpeg.org/}{https://www.ffmpeg.org/}} at the rate of 12 frames per second (fps). 
Although the original frame rate in the considered frame sequences varies from 11 fps to 16 fps, we uniformised it to keep the annotation process consistent.
As we retained the original time stamps, however, the videos in ROAD can still be synchronised with the LiDAR and GPS data associated with them in the OxRD dataset, allowing future work on multi-modal approaches.

\subsection{Annotation process} \label{subsec:dataset-annotation}

\emph{Annotation tool}. 
Annotating tens of thousands of frames rich in content is a very intensive process; therefore, a tool is required which \REBUT{can make this process both fast and intuitive}. 
For this work, we adopted Microsoft's VoTT\footnote{\url{https://github.com/Microsoft/VoTT/}}. The most useful feature of this annotation tool is that it can copy annotations (bounding boxes and their labels) from one frame to the next, while maintaining a unique identification for each box, so that boxes across frames are automatically linked together. 
Moreover, VoTT also allows for multiple labels, thus lending itself well to ROAD's multi-label annotation concept. %
\REBUT{A number of examples of annotated frames from the two videos using the VOTT tool is provided in supplementary material.}

\begin{figure*}[t]
  \centering{
      \includegraphics[width=0.99\textwidth]{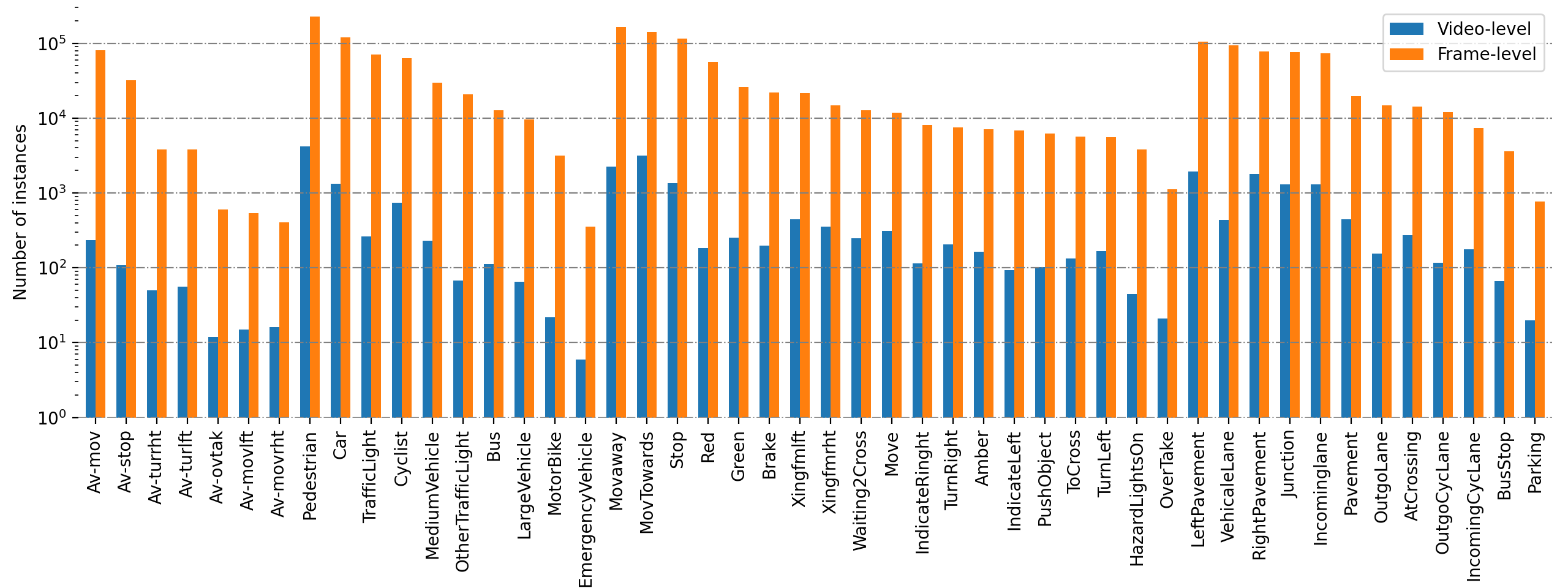}
  \caption{Number of instances of each class of individual label-types, in logarithmic scale. \label{fig:stats}}
  }
\end{figure*}

\REBUT{\emph{Annotation protocol}. All salient objects and actors within the frame were labelled, with the exception of inactive participants (mostly parked cars) and objects / actors at large distances from the ego vehicle, as the latter were judged to be irrelevant to the AV’s decision making. This can be seen in the attached 30-minute video\footnote{\url{https://www.youtube.com/watch?v=CmxPjHhiarA}.} portraying ground truth and predictions.
As a result, pedestrians, cyclists and traffic lights were always labelled. Vehicles, on the other hand, were only labelled when active (i.e., moving, indicating, being stopped at lights or stopping with hazard lights on on the side of road). As mentioned, only parked vehicles were not considered active (as they do not arguably influence the AV's decision making), and were thus not labelled.}


\emph{Event label generation}. 
Using the annotations manually generated for actions and agents in the multi-label scenario as discussed above it is possible to generate \emph{event-level} labels about agents, e.g. \textit{Pedestrian / Moving towards} the AV \textit{On right pavement} or \textit{Cyclist / Overtaking / In vehicle lane}. 
Any combinations of location, action and agent labels are admissible.
If location labels are ignored, the resulting event labels become location-invariant. 
\\
In addition to event tubes, in this work we do explore
\textit{agent-action} pair instances (see Sec. \ref{sec:experiments}).
Namely, given an agent tube and the continuous temporal sequence of action labels attached to its constituent bounding box detections, we can generate action tubes by looking for changes in the action label series associated with each agent tube. For instance, a \textit{Car} appearing in a video might be first \textit{Moving away} before \textit{Turning left}. The agent tube for the car will then be formed by two contiguous agent-action tubes: a first tube with label pair \textit{Car / Moving away} and a second one with pair \textit{Car / Turning left}.

\begin{table}[ht!]
     \centering
     \caption{\REBUT{ROAD tasks and attributes.}}     
     \label{tab:task_types}
     \begin{tabular}{lccc}
         Task type & Problem type & Output & Multiple labels \\
         \midrule
         Active agent & Detection & Box\&Tube & No \\
         Action & Detection & Box\&Tube & Yes \\
         Location & Detection & Box\&Tube & Yes \\
         Duplex & Detection & Box\&Tube & Yes \\
         Event & Detection & Box\&Tube & Yes \\
         AV-action & Temp segmentation & Start/End & No \\
         \bottomrule
     \end{tabular}
 \end{table}

\subsection{Tasks} \label{sec:dataset-tasks}

ROAD is designed as a sandbox for validating the six tasks relevant to situation awareness in autonomous driving outlined in Sec. \ref{subsec:intro-road}.
Five of these tasks are detection tasks, while the last one is a frame-level action recognition task sometimes referred to as 'temporal action segmentation'~\cite{sigurdsson2018charadesego}, \REBUT{Table~\ref{tab:task_types} shows the main attributes of these tasks}.

All detection tasks are evaluated both at frame-level and at video- (tube-)level. \emph{Frame-level detection} refers to the problem of identifying in each video frame the bounding box(es) of the instances there present, together with the relevant class labels. \emph{Video-level detection} consists in regressing a whole series of temporally-linked bounding boxes (i.e., in current terminology, a '{tube}') together with the relevant class label. 
In our case, the bounding boxes will mark a specific active agent in the road scene. The labels may issue (depending on the specific task) either from one of the individual label types described above (i.e., agent, action or location) or from one of the meaningful combinations described in~\ref{subsec:dataset-annotation} (i.e., either agent-action pairs or events). 

Below we list all the tasks for which we currently provide a baseline, with a short description.

\begin{enumerate}
\item{\emph{Active agent detection} (or \emph{agent detection}) 
aims at localising an active agent using a bounding box (frame-level) or a tube (video-level) and \REBUT{assigning a class label to it.}
} 
\item{\emph{Action detection} 
seeks to localise an active agent occupied in performing \REBUT{a specific} action from the list of action classes.} 
\item
\REBUT{
In \emph{agent location detection} (or \emph{location detection}) a label from the relevant list of locations (as seen from the AV) is sought and attached to the relevant bounding box or tube.} 
\item In \emph{agent-action detection} the bounding box or tube is \REBUT{assigned} a pair agent-action 
as explained in~\ref{subsec:dataset-annotation}. We sometimes refer to this task as 'duplex detection'.
\item{\emph{Road event detection} (or \emph{event detection}) consist in assigning to each box or tube a triplet of class labels.} 
\item{\emph{Autonomous vehicle temporal action segmentation} is a frame-level action classification task in which each video frame is assigned a label from the list of possible AV own actions. 
We refer to this task as 'AV-action segmentation', similarly to \cite{sigurdsson2018charadesego}.}
\end{enumerate}
 
\subsection{Quantitative summary} \label{sec:dataset-summary}

Overall, $122K$ frames extracted from 22 videos were labelled, in terms of both AV own actions (attached to the entire frame) and bounding boxes with attached one or more labels of each of the three types: agent, action, location.
In total, ROAD includes $560K$ bounding boxes with $1.7M$ instances of individual labels. The latter figure can be broken down into $560K$ instances of agent labels, $640K$ instances of action labels, and $499K$ instances of location labels.
Based on the manually assigned individual labels, we could identify $603K$ instances of duplex (agent-action) labels and $454K$ instances of triplets (event labels).

The number of instances for each individual class from the three lists is shown in Fig.~\ref{fig:stats} (frame-level, in orange). 
The $560K$ bounding boxes make up $7,029$, $9,815$, $8,040$, $9,335$ and $8,394$ tubes for the label types agent, action, location, agent-action and event, respectively.
Figure~\ref{fig:stats} also shows the number of tube instances for each class of individual label types as number of video-level instances (in blue). 

%% file: text/baseline_methods.tex
\section{\REBUT{BASELINE AND CHALLENGE}} \label{sec:baseline-method}

Inspired by the success of recent 3D CNN architectures~\cite{carreira2017quo} for video recognition and of feature-pyramid networks (FPN)~\cite{lin2017feature} with focal loss~\cite{lin2017focal}, we propose a simple yet effective 3D feature pyramid network (3D-FPN) with focal loss
as a baseline method for ROAD's detection tasks.
We call this architecture \emph{3D-RetinaNet}.

\begin{figure*}[t!]
    \centering{
        \includegraphics[width=\textwidth]{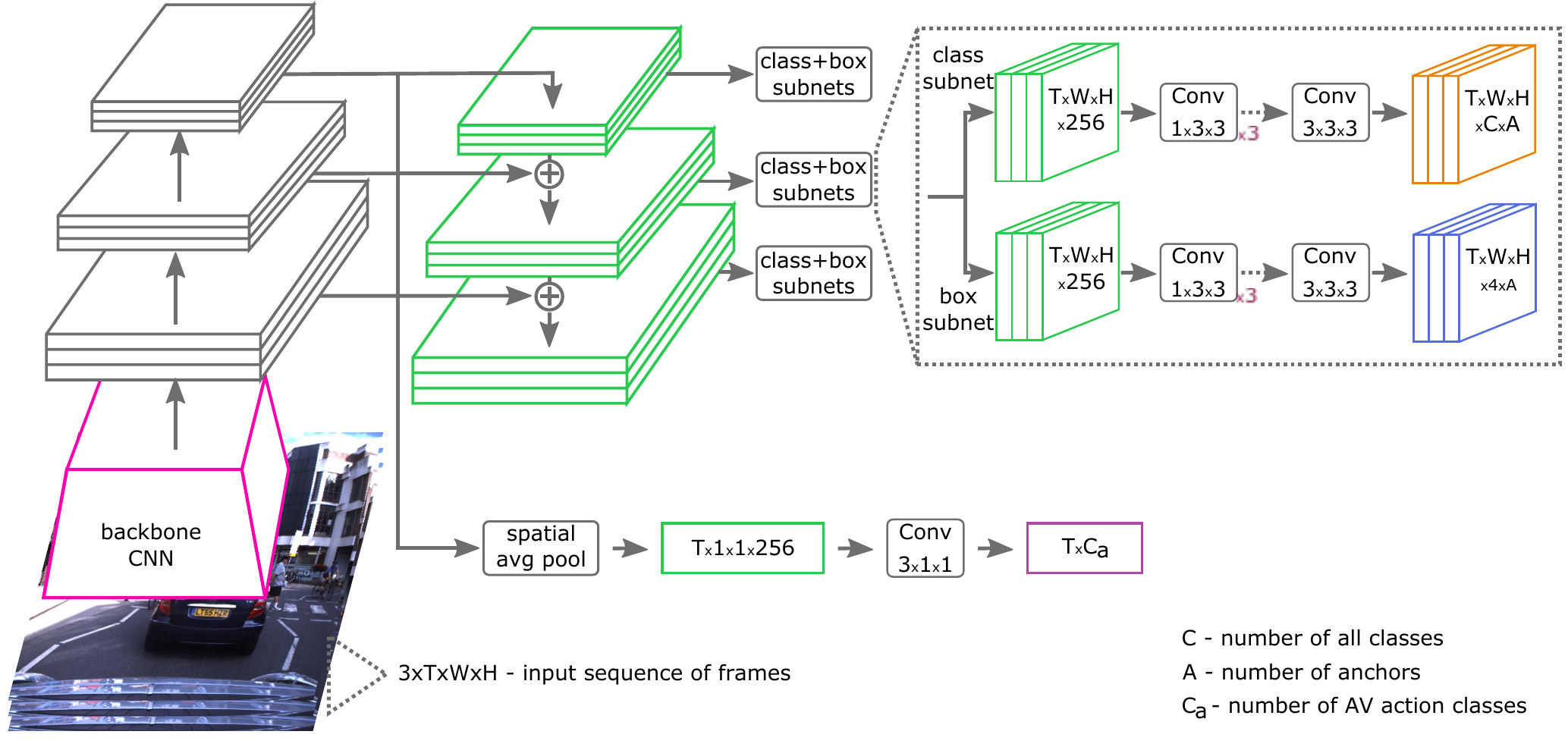}
    \caption{Proposed 3D-RetinaNet architecture for online video processing.}
    \label{fig:3dfpn}
    }
\end{figure*}

\subsection{3D-RetinaNet architecture}

The data flow of 3D-RetinaNet is shown in Figure~\ref{fig:3dfpn}.
The input is a sequence of $T$ video frames. As in classical FPNs~\cite{lin2017feature}, the initial block of 3D-RetinaNet consists of a backbone network outputting a series of forward feature pyramid maps, and of lateral layers producing the final feature pyramid composed by $T$ feature maps. 
The second block is composed by two sub-networks which process these features maps to produce both bounding boxes ($4$ coordinates) and $C$ classification scores for each anchor location (over $A$ possible locations). 
In the case of ROAD, the integer $C$ is the sum of the numbers of agent, action, location, action-agent (duplex) and agent-action-location (event) classes, plus one reserved for an \emph{agentness} score. The extra class agentness is used to describe the presence or absence of an active agent.
As in FPN~\cite{lin2017feature}, we adopt ResNet50~\cite{he2016deep} as the backbone network. 

\emph{2D versus 3D backbones}. 
\REBUT{In our experiments we show results obtained using three different backbones: frame-based ResNet50 (2D), inflated 3D (I3D)~\cite{carreira2017quo} and Slowfast~\cite{feichtenhofer2019slowfast}, in the manner also explained in ~\cite{nonlocal2018wang,feichtenhofer2019slowfast}. 
Choosing a 2D backbone makes the detector completely online~\cite{singh2017online}, with a delay of a single frame. Choosing an I3D or a Slowfast backbone, instead, causes a 4-frame delay at detection time.
Note that, as Slowfast and I3D networks makes use of a max-pool layer with stride 2, the initial feature pyramid in the second case contains $T/2$ feature maps. Nevertheless, in this case we can simply linearly upscale the output to $T$ feature maps.}

\emph{AV action prediction heads}. 
In order for the method to also address the prediction of the AV's own actions (e.g. whether the AV is stopping, moving, turning left etc.),
we branch out the last feature map of the pyramid (see Fig. \ref{fig:3dfpn}, bottom)
and apply spatial average pooling, followed by a temporal convolution layer. The output is a score for each of the $C_a$ classes of AV actions, for each of the $T$ input frames. 

\emph{Loss function}. 
As for the choice of the loss function, we adopt a binary cross-entropy-based focal loss~\cite{lin2017focal}. We choose a binary cross entropy because our dataset is multi-label in nature. 
The choice of a focal-type loss is motivated by the expectation that it may help the network deal with long tail and class imbalance (see Figure~\ref{fig:stats}).

\subsection{Online tube generation via agentness score} \label{subsec:tube-generation}

The autonomous driving scenario requires any suitable method for agent, action or event tube generation to work in an \emph{online} fashion, by incrementally updating the existing tubes as soon as a new video frame is captured.
For this reason, this work adopts a recent algorithm proposed by Singh \etal \cite{singh2017online}, which incrementally builds action tubes in an online fashion and at real-time speed. To be best of our knowledge, \cite{singh2017online} was the first online multiple action detection approach to appear in the literature, and was later adopted by almost all subsequent works~\cite{kalogeiton2017action,li2020actionsas,singh2018tramnet} on action tube detection. 

\emph{Linking of detections}. 
We now briefly review the tube-linking method of Singh~\etal~\cite{singh2017online}, and show how it can be adapted to build agent tubes based on an 'agentness' score, rather than build a tube separately for each class as proposed in the original paper.
This makes the whole detection process faster, since the total number of classes is much larger than in the original work~\cite{singh2017online}.
The proposed 3D-RetinaNet is used to regress and classify detection boxes in each video frame potentially containing an active agent of interest. Subsequently, detections whose score is lower than $0.025$ are removed and non-maximal suppression is applied based on the agentness score. 

At video start, each detection initialises an agentness tube. From that moment on, at any time instance $t$ the highest scoring tubes in terms of mean agentness score up to $t-1$ are linked to the detections with the highest agentness score in frame $t$ which display an Intersection-over-Union (IoU) overlap with the latest detection in the tube above a minimum threshold $\lambda$. The chosen detection is then removed from the pool of frame-$t$ detections. This continue until the tubes are either assigned or not assigned a detection from current frame. Remaining detections at time $t$ are used to initiate new tubes.  A tube is terminated after no suitable detection is found for $n$ consecutive frames. 
As the linking process takes place, each tube carries scores for all the classes of interest for the task at hand (e.g., action detection rather than event detection), as produced by the classification subnet of 3D-RetinaNet. We can then label each agentness tube using the $k$ classes that show the highest mean score over the duration of the tube.

\emph{Temporal trimming}. 
Most tubelet based methods~\cite{kalogeiton2017action,li2020actionsas,simonyan2014two} do not perform any temporal trimming of the action tubes generated in such a way (i.e., they avoid deciding when they should start or end). Singh~\etal~\cite{singh2017online} proposed to pose the problem in a label consistency formulation solved via dynamic programming. However, as it turns out, temporal trimming~\cite{singh2017online} does not actually improve performance, as shown in \cite{singh2018tramnet}, except in some settings, for instance in the DALY ~\cite{daly2016weinzaepfel} dataset. 

The situation is similar for our ROAD dataset as opposed to what happens on UCF-101-24, for which temporal trimming based on solving the label consistency formulation in terms of the actionness score, rather than the class score, does help improve localisation performance. Therefore, in our experiments we only use temporal trimming on the UCF-101-24 dataset but not on ROAD.

\subsection{\REBUT{The ROAD challenge}} \label{sec:challenge}

\REBUT{To introduce the concept of road event, our new approach to situation awareness and the ROAD dataset to the computer vision and AV communities, some of us have organised in October 2021 the workshop "The ROAD challenge: Event Detection for Situation Awareness in Autonomous Driving"\footnote{\url{https://sites.google.com/view/roadchallangeiccv2021/}.}.
For the challenge, we selected (among the tasks described in Sec. \ref{sec:dataset-tasks}) only three tasks: agent detection, action detection and event detection, which we identified as the most relevant to autonomous driving.}

\REBUT{As standard in action detection, evaluation was done in terms of video mean average precision (video-mAP).
3D-RetinaNet was proposed as the baseline for all three tasks.
Challenge participants had 18 videos available for training and validation. The remaining 4 videos were to be used to test the final performance of their model. 
This split was applied to all the three challenges (split 3 of the ROAD evaluation protocol, see Section \ref{subsec:exp:road}).}

\REBUT{The challenge opened for registration on April 1 2021, with the training and validation folds released on April 30, the test fold released on July 20 and the deadline for submission of results set to September 25. 
For each stage and each Task the maximum number of submissions was capped at 50, with an additional constraint of 5 submissions per day.
The workshop, co-located with ICCV 2021, took place on October 16 2021.}

\REBUT{In the validation phase we had between three and five teams submit between 15 and 17 entries to each of three challenges. In the test phase, which took place after the summer, we noticed a much higher participation with 138 submissions from 9 teams to the agent challenge, 98 submissions from 8 teams to the action challenge, and 93 submission from 6 teams to the event detection challenge.}

\REBUT{The methods proposed by the winners of each challenge are briefly recalled in Section \ref{sec:experiments-challenge}.}

\REBUT{\emph{Benchmark maintenance}. After the conclusion of the ROAD @ ICCV 2021 workshop, the challenge has been re-activated to allow for submissions indefinitely.
The ROAD benchmark will be maintained by withholding the test set from the public on the \url{eval.ai} platform\footnote{\url{https://eval.ai/web/challenges/challenge-page/1059/overview}}, where teams can submit their predictions for evaluation. 
Training and validation sets can be downloaded from \url{https://github.com/gurkirt/road-dataset}.}

%% file: text/experiments.tex
\section{EXPERIMENTS} \label{sec:experiments}

In this section we present results on the various task the ROAD dataset is designed to benchmark (see Sec. \ref{sec:dataset-tasks}), as well as the action detection results delivered by our 3D-RetinaNet model on UCF-101-24~\cite{jiang2014thumos,soomroucf101}. 

We first present the evaluation metrics and implementation details specific to ROAD in Section~\ref{subsec:details}.
In Section~\ref{subsec:exp:ucf-becnhmark} we benchmark our 3D-RetinaNet model for the action detection problem on UCF-101-24. The purpose is to show that this baseline model is competitive with the current state of the art in action tube detection while only using RGB frames as input, and to provide a sense of how challenging ROAD is when compared to standard action detection benchmarks.
Indeed, the complex nature of the real-world, non-choreographed road events, often involving large numbers of actors simultaneously responding to a range of scenarios in a variety of weather conditions makes ROAD a dataset which poses significant challenges when compared to other, simpler action recognition benchmarks.

In Section~\ref{subsec:exp:road} we illustrate and discuss the baseline results on ROAD for the different tasks (Sec. \ref{sec:experiments-tasks}), \REBUT{using a 2D ResNet50, an I3D and a Slowfast backbone, as well as the agent detection performance of the standard YOLOv5 model.} Different training/testing splits encoding different weather conditions \REBUT{are examined using the I3D backbone} (Sec. \ref{sec:experiments-weather}). In particular, in Sec. \ref{sec:experiments-joint} we show the results one can obtain when predicting composite labels as products of single-label predictions as opposed to training a specific model for them, as this can provide a crucial advantage in terms of efficiency, as well as give the system the flexibility to be extended to new composite labels without retraining.
Finally, in Sec. \ref{sec:experiments-av} we report our baseline results on the temporal segmentation of AV actions.

\subsection{Implementation details} \label{subsec:details}

The results are evaluated in terms of both frame-level bounding box detection and of tube detection. In the first case, the evaluation measure of choice is \emph{frame mean average precision} (f-mAP). We set the Intersection over Union (IoU) detection threshold to $0.5$ (signifying a 50\% overlap between predicted and true bounding box). For the second set of results we use \emph{video mean average precision} (video-mAP), as information on how the ground-truth BBs are temporally connected is available. These evaluation metrics are standard in action detection~\cite{Saha2016,Weinzaepfel-2015,kalogeiton2017action,singh2017online,li2018recurrent}.
\\
We also evaluate actions performed by AV, 
as described in \ref{sec:dataset-multilabel}. Since this is a temporal segmentation problem, we adopt the mean average precision metric computed at frame-level, as standard on the Charades~\cite{sigurdsson2018charadesego} dataset.

\begin{table}[t]
    \centering
    \setlength{\tabcolsep}{4pt}
    \caption{Comparison of the action detection performance (frame-mAP@0.5 (f-mAP) and video-mAP at different IoU thresholds) of the proposed 3D-RetinaNet baseline model with the state-of-the-art on the UCF-101-24 dataset.}
    {\footnotesize
    \scalebox{0.90}{
    \begin{tabular}{lccccc}
    \toprule
    Methods \// $\delta$ =         &                 f-mAP & 0.2  & 0.5 & 0.75 & 0.5:0.9 \\\midrule
    \multicolumn{6}{l}{RGB + FLOW methods} \\\midrule
    MR-TS Peng~\etal~\cite{peng2016eccv}            & --      & 73.7 & 32.1 & 00.9  & 07.3        \\
    FasterRCNN Saha~\etal \cite{Saha2016}           & --      & 66.6 & 36.4 & 07.9  & 14.4  \\
    SSD + OJLA Behl~\etal~\cite{behl2017incremental}${}^*$ & -- & 68.3 & 40.5 & 14.3 & 18.6 \\
    SSD Singh~\etal~\cite{singh2017online}${}^*$    & --      & 76.4 & 45.2  & 14.4  & 20.1 \\
    AMTnet Saha~\etal~\cite{saha2017amtnet}${}^*$         & -- & 78.5  & 49.7 & 22.2 & 24.0 \\
    ACT Kalogeiton~\etal~\cite{kalogeiton2017action}${}^*$  & --  & 76.5 & 49.2  & 19.7  & 23.4  \\
    TraMNet Singh~\etal~\cite{singh2018tramnet}${}^*$  & --  &  79.0 & 50.9 & 20.1 & 23.9  \\
    Song~\etal~\cite{song2019tacnet}  & 72.1  & 77.5 & 52.9 & 21.8 & 24.1  \\
    Zhao~\etal~\cite{zhao2019dance}  & --  & 78.5 & 50.3 & 22.2 & 24.5  \\
    I3D Gu~\etal~\cite{gu2018ava} & 76.3  & -- & \textbf{59.9} & -- & -- \\
    Li~\etal~\cite{li2020actionsas}${}^*$   & \textbf{78.0}  & \textbf{82.8} & 53.8 & \textbf{29.6} & \textbf{28.3}  \\
    \midrule
    \multicolumn{6}{l}{RGB only methods} \\
    \midrule
    RGB-SSD Singh~\etal~\cite{singh2017online}${}^*$  & 65.0 & 72.1 & 40.6 &	14.1 &	18.5 \\
    RGB-AMTNet Saha~\etal~\cite{saha2017amtnet}${}^*$  & -- & 75.8 & 45.3 & 19.9 & 22.0 \\
    3D-RetinaNet / 2D (ours)${}^*$  & 65.2 & 73.5 & 48.6 & 22.0 & 22.8 \\
    3D-RetinaNet / I3D (ours) & \textbf{75.2} & \textbf{82.4} & \textbf{58.2} & \textbf{25.5} & \textbf{27.1} \\
    \bottomrule
    \multicolumn{5}{l}{ ${}^*$ online methods}\\
    \end{tabular}
    }
    }
    \label{tab:ucf24} 
\end{table}

We use sequences of $T = 8$ frames as input to 3D-RetinaNet. 
Input image size is set to $512\times682$. This choice of 
$T$ is the result of GPU memory constraints; however, at test time, we unroll our convolutional 3D-RetinaNet for sequences of 32 frames, showing that it can be deployed in a streaming fashion.
We initialise the backbone network with weights pretrained on Kinetics~\cite{kay2017kinetics}. For training we use an SGD optimiser with step learning rate. The initial learning rate is set to $0.01$ and drops by a factor of $10$ after $18$ and $25$ epochs, up to an overall $30$ epochs. For tests on the UCF-101-24 dataset the learning rate schedule is shortened to a maximum $10$ epochs, and the learning rate drop steps are set to $6$ and $8$. 

The parameters of the tube-building algorithm (Sec. \ref{subsec:tube-generation}) are set by cross validation. 
For ROAD we obtain $\lambda = 0.5$ and $k = 4$. 
For UCF-101-24, we get $\lambda = 0.25$ and $k = 4$. 
Temporal trimming is only performed on UCF-101-24.

\subsection{Baseline performance on UCF-101-24}~\label{subsec:exp:ucf-becnhmark}

Firstly, we benchmarked 3D-RetinaNet on UCF-101-24~\cite{jiang2014thumos,soomroucf101}, using the corrected annotations from \cite{singh2017online}. 
We evaluated both frame-mAP and video-mAP and provided a comparison with state-of-the-art approaches in Table~\ref{tab:ucf24}. It can be seen that our baseline is competitive with the current state-of-the-art~\cite{li2020actionsas,gu2018ava}, even as those methods use both RGB and optical flow as input, as opposed to ours. As shown in the bottom part of Table~\ref{tab:ucf24}, 3D-RetinaNet outperforms all the methods solely relying on appearance (RGB) by large margins.
The model retains
the simplicity of single-stage methods, while sporting, as we have seen, the flexibility of being able to be reconfigured by changing the backbone architecture. \REBUT{Note that its performance} could be further boosted using the simple optimisation technique proposed in~\cite{kong2019consistent}.

\subsection{Experimental results on ROAD} \label{subsec:exp:road}

\begin{table}[t!]
    \centering
    \setlength{\tabcolsep}{4pt}
    \caption{Splits of training, validation and test sets for the ROAD dataset with respect to weather conditions. The table shows the number of videos in each set or split. For splits, the first figure is the number of training videos, the second number that of validation videos.} 
    {\footnotesize
    \scalebox{0.90}{
    \begin{tabular}{lcccc}
    \toprule
    Condition & sunny & overcast & snow & night \\ \midrule
    Training and validation & 7 & 7 & 1 & 3 \\
    \textit{Split-1} & 7/0 & 4/3 & 1/0 & 3/0 \\
    \textit{Split-2} & 7/0 & 7/0 & 1/0 & 0/3 \\
    \textit{Split-3} & 4/3 & 7/0 & 1/0 & 3/0 \\ \midrule
    Testing & 1 & 1 & 1 & 1 \\ \bottomrule
    \end{tabular}
    }
    }
    \label{tab:splits} 
\end{table}

\subsubsection{Three splits: modelling weather variability}~\label{subsec:splits}
For the benchmarking of the ROAD tasks, we divided the dataset into two sets.
The first set contains 18 videos for training and validation purposes, while the second set contains 4 videos for testing, equally representing the four types of weather conditions encountered.

The group of training and validation videos is further subdivided into three different ways ('splits'). In each split, 15 videos are selected for training and 3 for validation. Details on the number of videos for each set and split are shown in Table~\ref{tab:splits}. 
All 3 validation videos for Split-1 are overcast; 4 overcast videos are also present in the training set. As such, Split-1 is designed to assess the effect of different overcast conditions.
Split-2 has all 3 night videos in the validation subset, and none in the training set. It is thus designed to test model robustness to day/night variations. Finally, Split-3 contains 4 training and 3 validation videos for sunny weather: it is thus designed to evaluate the effect of different sunny conditions, as camera glare can be an issue when the vehicle is turning or facing the sun directly.

Note that there is no split to simulate a bias towards snowy conditions, as the dataset only contains one video of that kind.
The test set (bottom row) is more uniform, as it contains one video from each environmental condition.

\subsubsection{Results on the various tasks} \label{sec:experiments-tasks}

Results are reported for the tasks discussed in Section~\ref{sec:dataset-tasks}. 

\emph{{Frame-level} results across the five detection tasks} are summarised in Table~\ref{tab:frame-avg-results} using the frame-mAP (f-mAp) metric, for 
a detection threshold of $\delta = 0.5$. 
The reported figures are averaged across the three splits described above, {in order to assess the overall robustness of the detectors to domain variations.}
Performance within each split is evaluated on both the corresponding validation subset and test set. 
\REBUT{Each row in the Table shows the result of a particular combination of backbone network (2D, I3D, or Slowfast)} and test-time sequence length (in number of frames, { 8 and 32).
Frame-level results vary between 16.8\% (events) and 65.4\% (agentness) \REBUT{for I3D, and between 23.9\% and 69.2\% for Slowfast}.
Clearly, for each detection task except agentnness (which amounts to agent detection on ROAD) the performance is quite lower than the 75.2\% achieved by our I3D baseline network on UCF-101-24 (Table \ref{tab:ucf24}, last row). This is again due to the numerous nuisance factors present in ROAD, such as significant camera motion, weather conditions, etc. For a fair comparison, note that there are only 11 agent classes, as opposed to e.g. 23 action classes and 15 location classes.}

\emph{Video-level results} 
are reported in terms of video-mAP in Table~\ref{tab:video-avg-results}. As for the frame-level results, tube detection performance  (see Sec.~\ref{subsec:tube-generation}) is averaged across the three splits. One can appreciate the similarities between frame- and video-level results, which follow a similar trend \REBUT{albeit at a much lower absolute level}. {Again, results are reported for different backbone networks and sequence lengths. 
\REBUT{Not considering the YOLOv5 numbers,} video-level results at detection threshold $\delta = 0.2$ vary between a minimum of \REBUT{20.5\% (actions) to a maximum of 33.0\% (locations)}, compared to the 82.4\% achieved on UCF-101-24. For a detection threshold $\delta$ equal to 0.5, the video-level results lie between 4.7\% (actions) and 11\% (locations) compared to the 58.2\% achieved on UCF-101-24 for the same IoU threshold.
The difference is quite dramatic, and highlights the order of magnitude of the challenge involved by perception in autonomous driving compared to a standard benchmark portraying only human actions.} Furthermore, we can notice a few important facts.

\begin{table}[t!]
    \centering
    \setlength{\tabcolsep}{4pt}
    \caption{Frame-level results (mAP $\%$) averaged across the three splits of ROAD. 
    \REBUT{The considered models differ in terms of backbone network (2D, I3D, and Slowfast) and clip length (08 vs 32). The performance of YOLOv5 on agent detection is also reported.} Detection threshold $\delta = 0.5$.
    Both validation and test performance are reported for each entry. }
    {\footnotesize
    \scalebox{0.85}{
    \begin{tabular}{lcccccc}
    \toprule
    Model & {Agentness}  & Agents & Actions & Locations & Duplexes & {Events} \\ \midrule
    2D-08     & 51.8/63.4 & 30.9/39.5 & 15.9/22.0 & 23.2/30.8 & 18.1/25.1 & 10.6/12.8\\
    2D-32    & 52.4/64.2 & 31.5/39.8 & 16.3/22.6 & 23.6/31.4 & 18.7/25.8 & 10.8/13.0\\
    I3D-08    & 52.3/65.1 & 32.2/39.5 & 19.3/25.4 & 24.5/34.9 & 21.5/30.8 & 12.3/16.5\\
    I3D-32    & 52.7/65.4 & 32.3/39.2 & 19.7/25.9 & 24.7/35.3 & 21.9/31.0 & 12.6/16.8\\
    \REBUT{Slowfast-08} &   68.8/\textbf{69.2} & 41.9/47.5 & 26.9/31.1     & 34.6/\textbf{37.3} & 31.6/36.0 & \textbf{18.1}/23.7\\
    \REBUT{Slowfast-32} &   \textbf{69.3}/68.7 &  42.6/43.7 & \textbf{27.3}/\textbf{31.7} & \textbf{34.8}/36.4 & \textbf{32.0}/\textbf{36.1} & 18.0/\textbf{23.9}\\
    \REBUT{YOLOv5} &  - & \textbf{57.9}/\textbf{56.9} &   - & - & - & -\\
   \bottomrule
    \end{tabular}
    }
    }
    \label{tab:frame-avg-results} 
\end{table}

\begin{table}[t!]
  \centering
  \setlength{\tabcolsep}{4pt}
  \caption{Video-level results (mAP $\%$) averaged across the three ROAD splits. 
  \REBUT{The models differ in terms of backbone network (2D, I3D and Slowfast) and test time clip length (08 vs 32). The performance of YOLOv5 on agent detection is also reported.}
  Both validation and test performance are reported for each entry.}
  {\footnotesize
  \scalebox{0.85}{
  \begin{tabular}{lccccc}
  \toprule
  Model & Agents & Actions & Locations & Duplexes & {Events}\\ \midrule
  \midrule
  \multicolumn{6}{l}{Detection threshold $\delta = 0.2$} \\
  \midrule
  2D-08     & 22.2/25.1 & 10.3/13.9 & 18.2/24.8 & 16.1/21.9 & 12.8/14.7\\
  2D-32    & 22.6/25.0 & 11.2/14.5 & 18.5/25.9 & 16.2/22.7 & 13.0/15.3\\
  I3D-08    & 23.2/26.5 & 14.1/15.8 & 20.8/25.8 & 21.1/24.0 & 14.9/17.4\\
  I3D-32    & 24.4/26.9 & 14.3/17.5 & 21.3/27.1 & 21.4/25.5 & 15.9/17.9\\
  \REBUT{Slowfast-08} & 24.1/{29.0} & \textbf{16.0}/\textbf{20.5} & 28.3/\textbf{33.0} & 24.0/\textbf{27.3} & 18.9/22.4\\
  \REBUT{Slowfast-32} & 24.2/28.6 & \textbf{16.0}/19.55 & \textbf{29.0}/29.7 & \textbf{24.3}/26.1 & \textbf{19.1}/\textbf{22.5}\\
  \REBUT{YOLOv5} & \textbf{38.8}/\textbf{43.3} &  - & - & - & - \\
  \midrule
  \multicolumn{6}{l}{Detection threshold $\delta = 0.5$} \\ 
  \midrule
  2D-08     & 8.9/7.5 & 2.3/3.0 & 5.2/6.1 & 6.5/6.1 & 5.1/5.3\\ 
  2D-32    & 8.3/8.0 & 2.7/3.3 & 5.6/7.1 & 6.3/6.8 & 5.0/5.7\\ 
  I3D-08    & 9.2/9.6 & \textbf{4.0}/4.3 & 5.8/6.9 & 7.2/7.4 & 4.6/5.4\\ 
  I3D-32    & 9.7/10.2 & \textbf{4.0}/4.6 & 6.4/7.7 & 7.1/8.3 & 4.8/6.1\\ 
  \REBUT{Slowfast-08}    & 7.1/8.9 & 3.9/\textbf{4.7} & 7.1/\textbf{11.0} & \textbf{7.3}/7.7 & \textbf{6.5}/6.6\\ 
  \REBUT{Slowfast-32}    & 8.3/9.8 & 3.7/4.4 & \textbf{8.4}/10.0 & 7.1/\textbf{9.0} & 5.3/\textbf{7.3}\\ 
    \REBUT{YOLOv5} &  \textbf{18.7}/\textbf{13.9} &  - & - & - & - \\
  \bottomrule
  \end{tabular}
  }
  }
  \label{tab:video-avg-results} 
\end{table}


\begin{table*}[t]
  \centering
  \setlength{\tabcolsep}{4pt}
  \caption{Number of video- and frame-level instances for each label (individual or composite), left. 
  Corresponding frame-/video-level results (mAP@$\%$) for each of the three ROAD splits (right). {Val-$n$ denotes the validation set for Split $n$}. Results produced by an I3D backbone.} 
  {\footnotesize
  \scalebox{0.9}{
  \begin{tabular}{lccccc | cccccc}
  \toprule
  & \multicolumn{5}{c}{Number of instance}  & \multicolumn{6}{c}{Frame-mAP@$0.5$/Video-mAP@$0.2$} \\ 
  \midrule
  Train subset & \multicolumn{5}{c}{\#Boxes/\#Tubes} & \multicolumn{2}{c}{Train-1}  & \multicolumn{2}{c}{Train-2}  & \multicolumn{2}{c}{Train-3} \\ 
  \midrule 
  Eval subset & All & Val-1 & Val-2  & Val-3 & Test  & Val-1 & Test  & Val-2 & Test  & Val-3 & Test \\ 
  \midrule
Agent & 559142/7029  & 60103/781    & 79119/761    & 83750/809    & 82465/1138   & 44.5/30.1    & 34.0/25.7    & 17.2/16.0    & 40.9/27.4    & 35.3/27.1    & 42.6/27.5    \\ 
 Action & 639740/9815  & 69523/1054   & 89142/1065   & 95760/1111   & 94669/1548   & 26.2/17.0    & 26.6/17.4    & 11.7/11.4    & 25.3/17.3    & 21.2/14.6    & 25.7/17.9    \\ 
 Location & 498566/8040  & 56594/851    & 67116/864    & 77084/914    & 70473/1295   & 34.9/28.6    & 35.2/26.4    & 13.7/12.1    & 33.9/26.3    & 25.4/23.2    & 36.7/28.6    \\ 
 Duplex & 603274/9335  & 60000/965    & 85730/1032   & 88960/1050   & 89080/1471   & 28.2/25.3    & 28.7/23.4    & 13.6/17.3    & 31.4/24.8    & 23.9/21.6    & 33.0/28.4    \\ 
 Event & 453626/8394  & 43569/883    & 65965/963    & 72152/967    & 64545/1301   & 17.7/18.6    & 15.9/15.8    & 6.4/11.8     & 16.4/18.9    & 13.7/17.2    & 18.1/18.9    \\ 
 \midrule
 \multicolumn{6}{c}{Number of instances
 } & \multicolumn{6}{c}{Frame-AP} \\ 
 \midrule
 AV-action
 & 122154/490   & 17929/67     & 18001/56     & 16700/85     & 20374/82     & 57.9     & 45.7     & 33.5     & 43.6     & 43.7    & 48.2 \\
 \bottomrule
  \end{tabular}
  }
  }
\label{tab:classwise-splitwise-primary-labels} 
\end{table*}


\emph{Streaming deployment}. Increasing test sequence length from 8 to 32 does not much impact performance. 
This indicates that, even though the network is trained on 8-frame clips, being fully convolutional (including the heads in the temporal direction), it can be easily unrolled to process longer sequences at test time, making it easy to deploy in a streaming fashion. 
Being deployable in an incremental fashion is a must for autonomous driving applications; this is a quality that other tubelet-based online action detection methods~\cite{kalogeiton2017action,singh2018tramnet,li2020actionsas} fail to exhibit, as they can only be deployed in a sliding window fashion. \REBUT{Interestingly, the latest work on streaming object detection~\cite{li2020towards} 
proposes an approach that integrates latency and accuracy into a single metric for real-time online perception, termed `streaming accuracy'. We will consider adopting this metric in the future evolution of ROAD.}

\REBUT{\emph{Impact of the backbone}. Broadly speaking, the Slowfast~\cite{feichtenhofer2019slowfast} and I3D~\cite{carreira2017quo} versions of the backbone perform as expected, much better than the 2D version.
A Slowfast backbone can particularly help with tasks which require the system to `understand' movement, e.g. when detecting actions, agent-actions pairs and road events, at least at 0.2 IoU. Under more stringent localisation requirements ($\delta = 0.5$), it is interesting to notice how Slowfast's advantage is quite limited, with the I3D version often outperforming it. This shows that by simply switching backbone one can improve on performance or other desirable properties, such as training speed (as in or X3D~\cite{feichtenhofer2020x3d}). The 3D CNN encoding can be made intrinsically online, as in RCN~\cite{singh2019recurrent}. Finally, even stronger backbones using transformers~\cite{liu2021video,fan2021multiscale} can be plugged in.}

\emph{Level of task challenge}. The overall results on event detection (last column in both Table~\ref{tab:frame-avg-results} and Table~\ref{tab:video-avg-results}) are encouraging, \REBUT{but they remain in the low 20s at best, showing how challenging situation awareness is in road scenarios}.



\emph{Comparison across tasks}. From a superficial comparison of the mAPs obtained, action detection seems to perform worse than agent-action detection or even event detection. 
However, the headline figures are not really comparable since, as we know, the number of class per task varies. More importantly, within-class variability is often lower for composite labels. For example, the score for \textit{Indicating right} 
is really low, whereas \textit{Car / Indicating-right} 
has much better performance (see Supplementary material, Tables 11--13 for class-specific performance). This is because the within-class variability of the pair \textit{Car / Indicating-right} is much lower than that of \textit{Indicating right}, which puts together instances of differently-looking types of vehicles (e.g. buses, cars and vans) all indicating right.
Interestingly, results on agents are comparable among the four baseline models (especially for f-mAP and v-mAP at 0.2, see Tables \ref{tab:frame-avg-results} and \ref{tab:video-avg-results}).

\REBUT{\emph{YOLOv5 for Agent detection}. For completeness, we also trained YOLOv5\footnote{https://github.com/ultralytics/yolov5} for the detection of active agents. The results are shown in the last row of both Table~\ref{tab:frame-avg-results} and Table~\ref{tab:video-avg-results}. Keeping is mind that YOLOv5 is trained only on single input frames, it shows a remarkable improvement over the other baseline methods for active agent detection. We believe that is because YOLOv5 is better at the regression part of the detection problem -- namely, Slowfast has a recall of 71\% compared to the 94\% of YOLOv5, so that Slowfast has a 10\% lower mAP for active agent detection. We leave the combination of YOLOv5 for bounding box proposal generation and Slowfast for proposal classification as a promising future extension, which could lead to a general improvement across all tasks.}

\emph{Validation vs test results}. Results on the test set are, on average, superior to those on the validation set. This is because the test set includes data from all weather/visibility conditions (see Table \ref{tab:splits}), whereas for each split the validation set only contains videos from a single weather condition. E.g., in Split 2 all validation videos are nighttime ones.



\subsubsection{Results under different weather conditions} \label{sec:experiments-weather}

Table~\ref{tab:classwise-splitwise-primary-labels} shows, instead, the results obtained under the three different splits we created on the basis of the weather/environmental conditions of the ROAD videos, discussed in Section~\ref{subsec:splits} and summarised in Table~\ref{tab:splits}. 
Note that the total number of instances (boxes for frame-level results or tubes for video-level ones) of the five detection tasks is comparable for all the three splits.

We can see how Split-2 (for which all three validation videos are taken at night and no nighttime videos are used for training, see Table~\ref{tab:splits})
has the lowest validation results, as seen in Table~\ref{tab:classwise-splitwise-primary-labels} (Train-2, Val-2). 
When the network trained on Split-2's training data is evaluated on the (common) test set, instead, its performance is similar to that of the networks trained on the other splits (see Test columns). 
Split-1 has three overcast videos in the validation set, but also four overcast videos in the training set. The resulting network has the best performance across the three validation splits. Also, under overcast conditions one does not have the typical problems with night-time vision, nor glares issues as in sunny days. 
Split-3 is in a similar situation to Split-1, 
as it has sunny videos in both train and validation sets. 

These results seem to attest a certain robustness of the baseline to weather variations, for no matter the choice of the validation set used to train the network parameters (represented by the three splits), the performance on test data (as long as the latter fairly represents a spectrum of weather conditions) is rather stable.

\subsubsection{Joint versus product of marginals} \label{sec:experiments-joint}

One of the crucial points we wanted to test is weather the manifestation of composite classes (e.g., agent-action pairs or road events) can be estimated by separately training models for the individual types of labels, to then combine the resulting scores by simple multiplication (under an implicit, naive assumption of independence).
This would have the advantage of not having to train separate networks on all sort of composite labels, an obvious positive in terms of efficiency, especially if we imagine to further extend in the future the set of labels to other relevant aspects of the scene, such as attributes (e.g. vehicle speed). This would also give the system the flexibility to be extended to new composite events in the future without need for retraining.

\begin{table}[t]
  \centering
  \setlength{\tabcolsep}{4pt}
  \caption{Comparison of joint vs product of marginals approaches with I3D backbone.
  Number of video-/frame-level instances for each composite label ('No instances' column) and corresponding frame-/video-level results (mAP@$\%$) averaged across all three splits, on both validation and test sets.}
  {\footnotesize
  \scalebox{0.85}{
  \begin{tabular}{lccccc}
  \toprule
  & \multicolumn{1}{c}{No instances}  & \multicolumn{4}{c}{Frame-mAP@$0.5$/Video-mAP@$0.2$} \\ 
  \midrule
  Eval-method & \multicolumn{1}{c}{} & \multicolumn{2}{c}{Joint} & \multicolumn{2}{c}{Prod. of marginals}  \\ 
  \midrule 
  Eval subset & All & Val & Test  & Val & Test \\ 
  \midrule
 Duplexes & 603274/9335  & 21.9/21.4    & 31.0/25.5    & 21.6/21.2    & 30.8/24.3    \\ 
 Event                & 453626/8394  & 12.6/15.9    & 16.8/17.9   & 13.7/15.4    & 16.3/16.1    \\ 
 \bottomrule
  \end{tabular}
  }
  }
\label{tab:classwise-marginals}  \vspace{-3mm}
\end{table}

For instance, we may want to test the hypothesis that the score for the pair \textit{Pedestrian / Moving away} can be approximated as 
$P_{Ag}(\text{Pedestrian}) \times P_{Ac}(\text{Moving away})$, where $P_{Ag}$ and $P_{Ac}$ are the likelihood functions associated with the individual agent and action detectors\footnote{Technically the networks output scores, not probabilities, but those can be easily calibrated to probability values.}. 
This boils down to testing whether we need to explicitly learn a model for the joint distribution of the labels, or we can approximate that joint as a product of marginals. 
Learning-wise, the latter task involves a much smaller search space, so that marginal solutions (models) can be obtained more easily.

Table~\ref{tab:classwise-marginals} compares the detection performance on composite (duplex or event) labels obtained by expressly training a detection network for those ('Joint' column) as opposed to simply multiplying the detector scores generated by the networks trained on individual labels ('Prod. of marginals').
The results clearly validate the hypothesis that it is possible to model composite labels using predictions for individual labels without having to train on the former.
In most cases, the product of marginals approach achieves results similar or even better than those of joint prediction, although in some case (e.g. \textit{Traffic light red} and \textit{Traffic light red}, see Supplementary material again) we can observe a decrease in performance.
We believe this to be valuable insight for further research.

\subsubsection{Results of AV-action segmentation} \label{sec:experiments-av}

Finally, Table~\ref{tab:classwise-av-actions} shows the results of using 3D-RetinaNet to temporally segment AV-action classes, averaged across all three splits on both validation and test set. As we can see, the results for classes \textit{AV-move} and \textit{AV-stop} are very good, we think because these two classes are predominately present in the dataset. The performance of the 'turning' classes is reasonable, but the results for the bottom three classes 
are really disappointing. We believe this is mainly due the fact that the dataset is very heavily biased (in terms of number of instances) towards  the other classes. 
As we do intend to further expand this dataset in the future by including more and more videos, we hope the class imbalance issue can be mitigated over time. A measure of performance weighing mAP using the number of instances per class could be considered, but this is not quite standard in the action detection literature. At the same time, ROAD provides an opportunity for testing methods designed to address class imbalance.

\begin{table}[t]
  \centering
  \setlength{\tabcolsep}{4pt}
  \caption{AV-action temporal segmentation results (frame mAP$\%$) averaged across all three splits. 
  }
  {\footnotesize
  \scalebox{0.85}{
  \begin{tabular}{lccccc}
  \toprule
  & \multicolumn{1}{c}{No instances}  & \multicolumn{4}{c}{Frame-mAP@$0.5$ 
  } \\ 
  \midrule
  Model & \multicolumn{1}{c}{} & \multicolumn{2}{c}{I3D} & \multicolumn{2}{c}{2D}
  \\ 
  \midrule 
  Eval subset & All & Val & Test  & Val & Test \\ 
 \midrule
  Av-move                    & 81196/233    & 92.0     & 96.6  & 83.0     & 87.8     \\ 
  Av-stop                   & 31801/108    & 92.2     & 98.5  & 65.3     & 68.4     \\ 
  Av-turn-right                 & 3826/50      & 46.1     & 63.0  & 35.0     & 57.7     \\ 
  Av-turn-left                 & 3787/56      & 69.0     & 59.8  & 55.1     & 42.9     \\ 
  Av-overtake                  & 599/12       & 4.9      & 1.1   & 2.7      & 2.5      \\ 
  Av-move-left                 & 537/15       & 0.5      & 0.8   & 0.5      & 0.5      \\ 
  Av-move-right                 & 408/16       & 10.5     & 0.6   & 4.0      & 2.0      \\ 
  Total/Mean                & 122154/490   & 45.0     & 45.8  & 35.1     & 37.4     \\ 
 \bottomrule
  \end{tabular}
  }
  }
\label{tab:classwise-av-actions} 
\end{table}

\subsection{\REBUT{Challenge Results}} \label{sec:experiments-challenge}

\REBUT{Table~\ref{tab:challenge_results} compares the results of the top teams participating in our ROAD @ ICCV 2021 challenge with those of the Slowfast and YOLOv5 baselines, at a tube detection threshold of 0.2. The challenge server remains open at \url{https://eval.ai/web/challenges/challenge-page/1059/overview}, where one can consult the latest entries.}

\begin{table}[t]
    \centering
    \caption{\REBUT{Results (in video-mAP) of the winning entries to the ICCV 2021 ROAD challenge compared with the Slowfast and YOLOv5 baselines, at a detection threshold of $0.2$.}}
    \label{tab:challenge_results}
    \begin{tabular}{llccc}
        \toprule
         Task & Top team & Slowfast & YOLOv5 & Winners \\ \midrule
         Agent detection & Xidian & 29.0 & 43.3 & \textbf{52.4} \\
         Action detection & CMU-INF & 20.5 & - & \textbf{25.6} \\
         Event detection & IFLY & 22.4 & - & \textbf{24.7} \\
         \bottomrule
    \end{tabular}
\end{table}

\REBUT{\emph{Agent detection}. The agent detection challenge was won by a team formed by Chenghui Li, Yi Cheng, Shuhan Wang, Zhongjian Huang, Fang Liu of Xidian University, with an entry using YOLOv5 with post-processing. In their approach, agents are linked by evaluating their similarity between frames and grouping them into a tube. Discontinuous tubes are completed through frame filling, using motion information. Also, the authors note that YOLOv5 generates some incorrect bounding boxes, scattered in different frames, and take advantage of this by filtering out the shorter tubes. As shown in Table \ref{tab:challenge_results}, the postprocessing applied by the winning entry significantly outperforms our off-the-shelf implementation of YOLOv5 on agent detection.}

\REBUT{\emph{Action detection}. The action detection challenge was won by Lijun Yu, Yijun Qian, Xiwen Chen, Wenhe Liu and Alexander G. Hauptmann of team CMU-INF, with an entry called ``ArgusRoad: Road Activity Detection with Connectionist Spatiotemporal Proposals", based on their Argus++ framework for real-time activity recognition in extended videos in the NIST ActEV (Activities in Extended Video ActEV) challenge\footnote{\url{https://actev.nist.gov/}.}. The had to adapt their system to be run on ROAD, e.g. to construct tube proposals rather than frame-level proposals. The approach is a rather complex cascade of object tracking, proposal generation, activity recognition and temporal localisation stages \cite{liu2020argus}. Results show a significant (5\%) improvement over the Slowfast baseline, which is close to state-of-the-art in action detection, but still at a relatively low level (25.6\%)}

\REBUT{\emph{Event detection}. The event detection challenge was won by team IFLY (Yujie Hou and Fengyan Wang, from the University of Science and Technology of China and IFLYTEK). The entry consisted in a number of amendments to the 3D-RetinaNet baseline, namely: bounding box interpolation, tuning of the optimiser, ensemble feature extraction with RCN, GRU and LSTM units, together with some data augmentation. Results show an improvement of above 2\% over Slowfast, which suggests event better performance could be achieved by applying the ensemble technique to the latter.}

%% file: text/conclusion.tex
\section{FURTHER EXTENSIONS} \label{sec:extensions}

By design, ROAD is an open project which we expect to evolve and grow over time.

\emph{Extension to other datasets and environments}. 
In the near future we will work towards completing the multi-label annotation process for a larger number of frames coming from videos spanning an even wider range of road conditions.
Further down the line, we plan to extend the benchmark to other cities, countries and sensor configurations, to slowly grow towards an even more robust, 'in the wild' setting.
\REBUT{
In particular, we will initially target the Pedestrian Intention Dataset (PIE, \cite{rasouli2019pie}) and Waymo \cite{sun2020scalability}. 
The latter one comes with spatiotemporal tube annotation for pedestrian and vehicles, much facilitating the extension of ROAD-like event annotation there.
}

\emph{Event anticipation/intent prediction}.
ROAD is an oven-ready playground for action and event anticipation algorithms, a topic of growing interest in the vision community \cite{kong2017deep,kong2018adversarial}, as it already provides the kind of annotation that
allows researchers to test predictions of both future event labels and future event locations, both spatial and temporal.
Anticipating the future behaviour of other road agents is crucial to empower the AV to react timely and appropriately. The output of this Task should be in the form of one or more future tubes, with the scores of the associated class labels and the future bounding box locations in the image plane \cite{singh2018predicting}.
We will shortly propose a baseline method for this Task, but we encourage researchers in the area to start engaging with the dataset from now.

\emph{Autonomous decision making}.
In accordance with our overall philosophy,
we will design and share a baseline for AV decision making from intermediate semantic representations.
The output of this Task should be the decision made by the AV in response to a road situation \cite{hubmann2017decision}, represented as a collection of events as defined in this paper. As the action performed by the AV at any given time is part of the annotation, the necessary meta-data is already there. Although we did provide a simple temporal segmentation baseline for this task seen as a classification problem, we intend in the near future to propose a baseline from a decision making point of view, making use of the intermediate semantic representations produced by the detectors.

\emph{Machine theory of mind} \cite{rabinowitz2018machine} refers to the attempt to provide machines with (limited) ability to guess the reasoning process of other intelligent agents they share the environment with. Building on our efforts in this area \cite{Cuzzolin2020tom}, we will work with teams of psychologists and neuroscientists to provide annotations in terms of mental states and reasoning processes for the road agents present in ROAD. Note that theory of mind models can also be validated in terms of how close the predictions of agent behaviour they are capable of generating are to their actual observed behaviour.
Assuming that the output of a theory of mind model is intention (which is observable and annotated) the same baseline as for event anticipation can be employed.

\emph{Continual event detection}.
ROAD's conceptual setting is intrinsically incremental, one in which the autonomous vehicle keeps learning from the data it observes, in particular by updating the models used to estimate the intermediate semantic representations. The videos forming the dataset are particularly suitable, as they last 8 minutes each, providing a long string of events and data to learn from.
To this end, we plan to set a protocol for the continual learning of event classifiers and detectors and propose ROAD as the first continual learning benchmark in this area \cite{parisi2019continual}.

\section{CONCLUSIONS} \label{sec:conclusion}

\REBUT{This paper proposed a strategy for situation awareness in autonomous driving based on the notion of road events, and contributed a new ROad event Awareness Dataset for Autonomous Driving (ROAD) as a benchmark for this area of research. The dataset, built on top of videos captured as part of the Oxford RobotCar dataset~\cite{maddern20171}, has unique features in the field. Its rich annotation follows a multi--label philosophy in which road agents (including the AV), their locations and the action(s) they perform are all labelled, and road events can be obtained by simply composing labels of the three types. The dataset contains}
{22 videos with 122K annotated video frames, for a total of 560K detection bounding boxes associated with 1.7M individual labels.}

Baseline tests were conducted on ROAD using a new 3D-RetinaNet architecture, \REBUT{as well as a Slowfast backbone and a YOLOv5 model (for agent detection)}.
Both frame--mAP and video--mAP were evaluated. 
Our preliminary results 
\REBUT{highlight the challenging nature of ROAD, with the Slowfast baseline achieving a video-mAP on the three main tasks comprised between 20\% and 30\%, at low localisation precision (20\% overlap). YOLOv5, however, was able to achieve significantly better performance. These findings were reinforced by the results of the ROAD @ ICCV 2021 challenge,} 
and support the need for an even broader analysis, while highlighting the significant challenges specific to situation awareness in road scenarios.

Our dataset is extensible to a number of challenging tasks associated with situation awareness in autonomous driving, such as event prediction, trajectory prediction, continual learning and machine theory of mind, and we pledge to further enrich it in the near future by extending ROAD-like annotation to major datasets such as PIE and Waymo.

%% file: text/supp.tex
\section{ADDITIONAL DETAILS} \label{sec:supplementary}

\REBUT{In this section we provide some additional details on the annotation tool, class lists, number of instances, and the nature of composite labels.}

\begin{figure*}[ht!]
\centering{
    \includegraphics[width=0.9 \textwidth]{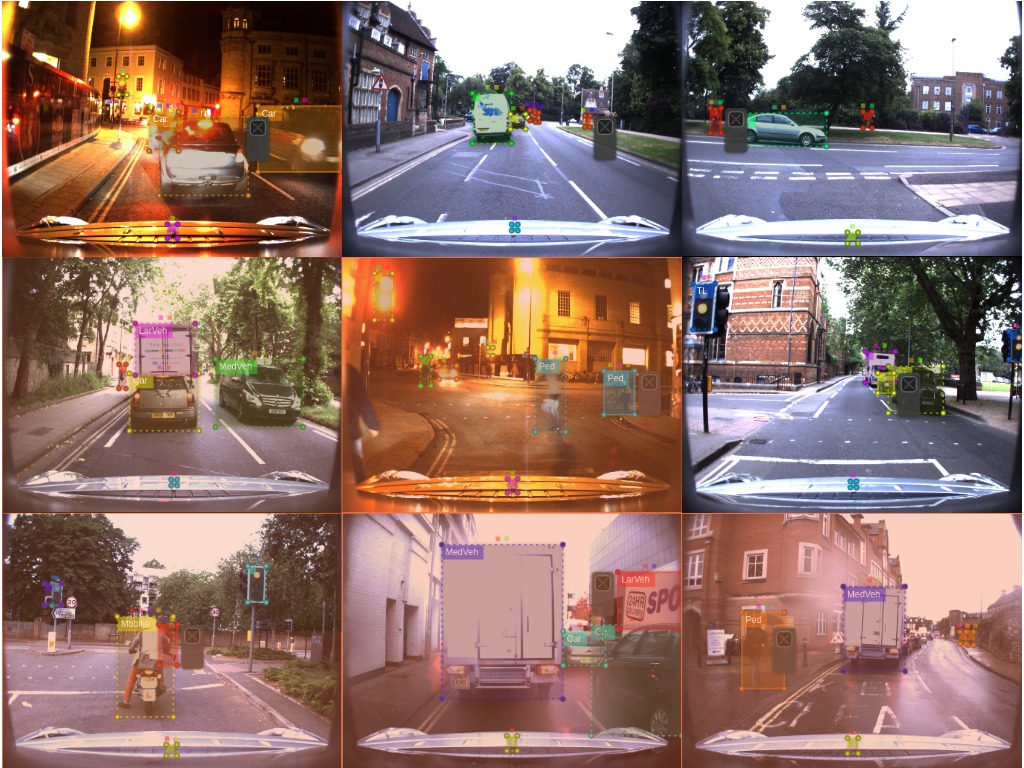}
\caption{Sample frames and annotation. ROAD's annotated frames cover multiple agents and actions, recorded under different weather conditions (overcast, sun, rain) at different times of the day (morning, afternoon and night). Ground truth bounding boxes and labels \REBUT{are also visible}. 
}
\label{fig:dataset-samples}
}
\end{figure*}

\subsection{Annotation tool}

\REBUT{VoTT provides a user-friendly graphical interface which allows annotators to draw boxes around the agents of interest and select the labels they want to associate with them from a predefined list at the bottom. After saving the annotations, the information is stored in a \texttt{json} file having the same name as the video. The file structure contains the bounding boxes' coordinates and the associated labels per frame; a unique ID (UID) helps identify boxes belonging to different frames which are part of the same tube. 
This is important as it is possible to have several instances related to the same kind of action. 
As a result, the temporal connections between boxes can be easily extracted from this file which is, in turn, crucial to measure performance in terms of video-mAP (see Main paper, Experiments).
It is important to note that tubes are built for each active agent, while the action label associated with a tube can in fact change over time, allowing us to model the complexity of an agent's road behaviour as it evolves over time.
A number of examples of annotated frames from videos are shown in Fig. \ref{fig:dataset-samples}, one captured during the day and one at night.}

\subsection{Class names and descriptions}

\REBUT{The class names for the different types of labels are listed here in a series of tables. Agent types classes are shown in Table~\ref{tab:agent}. Similarly, the class names and their description for the action, location, and AV-action labels are are given in Table~\ref{tab:action}, Table~\ref{tab:location} and Table~\ref{tab:av-actions}, respectively. }

\begin{table}[ht!]
\caption{\REBUT{List of ROAD active agent classes, with description.} \label{tab:agent}}
\centering
{
\footnotesize \scalebox{0.95}
{
\begin{tabular}{l l}
\toprule 
Label name & Description 
\\
\midrule 
Autonomous-vehicle & The autonomous vehicle itself 
\\
Car & A car up to the size of a multi-purpose vehicle 
\\
Medium vehicle & Vehicle larger than a car, such as van
\\
Large vehicle & Vehicle larger than a van, such as a lorry 
\\
Bus & A single or double-decker bus or coach 
\\
Motorbike & Motorbike, dirt bike, scooter with 2/3 wheels 
\\
Emergency vehicle & Ambulance, police car, fire engine, etc.
\\
Pedestrian & A person including children 
\\
Cyclist & A person is riding a push/electric bicycle 
\\
Vehicle traffic light & Traffic light related to the AV lane
\\
Other traffic light & Traffic light not related to the AV lane 
\\
\bottomrule
\end{tabular}
}
}
\end{table}

\begin{table*}[t!]
  \caption{\REBUT{List of ROAD action labels, with description.} \label{tab:action}}
  \centering
  \begin{tabular}{l l}
  \toprule
  Label name & Description 
  \\
  \midrule
  Moving away & Agent moving in a direction that increases the distance between Agent and AV. 
  \\
  Moving towards & Agent moving in a direction that decreases the distance between Agent and AV. 
  \\
  Moving & Agent moving perpendicular to the traffic flow or vehicle lane. 
  \\
  Reversing & Agent is moving backwards. 
  \\
  Braking & Agent is slowing down, vehicle braking lights are lit. 
  \\
  Stopped & Agent stationary but in ready position to move. 
  \\
  Indicating left & Agent indicating left by flashing left indicator light, or using a hand signal. 
  \\
  Indicating right & Agent indicating right by flashing right indicator light, or using a hand signal. 
  \\
  Hazard lights on & Hazards lights are flashing on a vehicle. 
  \\
  Turning left & Agent is turning in left direction.
  \\
  Turning right & Agent is turning in right direction. 
  \\
  Moving right & Moving lanes from the current one to the right one.  
  \\
  Moving left & Moving lanes from the current one to the left one. 
  \\
  Overtaking & Agent is moving around a slow-moving user, often switching lanes to overtake. 
  \\
  Waiting to cross & Agent on a pavement, stationary, facing in the direction of the road. 
  \\
  Crossing road from left & Agent crossing road, starting from the left and moving towards the right of AV. 
  \\
  Crossing road from right & Agent crossing road, starting from the right pavement and moving towards the left pavement. 
  \\
  Crossing & Agent crossing road.
  \\
  Pushing object & Agent pushing object, such as trolley or pushchair, wheelchair or bicycle. 
  \\
  Traffic light red & Traffic light with red light lit. 
  \\
  Traffic light amber & Traffic light with amber light lit.
  \\
  Traffic light green & Traffic light with green light lit. 
  \\
  Traffic light black & Traffic light with no lights lit or covered with an out-of-order bag. 
  \\
  \bottomrule
  \end{tabular}
 \end{table*}

  \begin{table*}[t]
    \caption{\REBUT{List of ROAD location labels, with description.}}
    \label{tab:location}
    \centering
    \begin{tabular}{l l}
    \toprule
    Label name & Description \\
    \midrule
    In vehicle lane & Agent in same road lane as AV. \\
    In outgoing lane & Agent in road lane that should be flowing in the same direction as vehicle lane. \\
    In incoming lane & Agent in road lane that should be flowing in the opposite direction as vehicle lane. \\
    In outgoing bus lane & Agent in the bus lane that should be flowing in the same direction as AV. \\
    In incoming bus lane & Agent in the bus lane that should be flowing in the opposite direction as AV. \\
    In outgoing cycle lane & Agent in the cycle lane that should be flowing in the same direction as AV. \\
    In incoming cycle lane & Agent in the cycle lane that should be flowing in the opposite direction as AV. \\
    On left pavement & Pavement to the left side of AV. \\
    On right pavement & Pavement to the right side of AV. \\
    On pavement & A pavement that is perpendicular to the movement of the AV. \\ 
    At junction & Road linked. \\
    At crossing & A marked section of road for cross, such as zebra or pelican crossing. \\
    At bus stop & A marked bus stop area on road, or a section of pavement next to a bus stop sign. \\
    At left parking & A marked parking area on left side of the road. \\
    At right parking & A marked parking area on right side of the road. \\
    \bottomrule
    \end{tabular}
  \end{table*}

\begin{table}[ht!]
  \centering
  \setlength{\tabcolsep}{4pt}
  \caption{\REBUT{AV-related action classes and number of frames labelled per class.}}
  {
  \footnotesize
  \scalebox{1.0}
  {
  \begin{tabular}{llc}
  \toprule
  Class label & Description & No instances \\ 
  \midrule
 \midrule
  Av-move                   & AV on the move  & 81,196    \\ 
  Av-stop                   & AV not moving  & 31,801     \\ 
  Av-turn-right             & Av turning right  & 3,826       \\ 
  Av-turn-left              & AV turning left  & 3,787       \\ 
  Av-overtake               & AV overtaking another vehicle  & 599        \\ 
  Av-move-left              & AV moving towards left  & 537        \\ 
  Av-move-right             & AV moving towards right  & 408        \\ 
  \textbf{Total}                &  --  & \textbf{122,154}    \\ 
 \bottomrule
  \end{tabular}
  }
  }
\label{tab:av-actions} 
\end{table}

\subsection{Number of instances}

We annotated $7K$ tubes \REBUT{associated with} individual agents. 
Each tube consists, on average, of approximately 80 bounding boxes linked over time, resulting in $559K$ bounding box-level agent labels. We also labelled $9.8K$ and $8K$ action and location tubes, respectively, resulting in $641K$ and $498K$ bounding box-level action and location labels, respectively. Overall, we generated circa $1.7M$ bounding box-level labels. In addition, $122K$ frame-level instances of actions by the autonomous vehicle were recorded. 

\subsection{Composite labels}

As explained in the paper, we considered \REBUT{in our analysis} pairs combining agent and action labels. Event labels were constructed by forming triplets composed of agent, action and location labels. Tables~\ref{table:duplex_count} and~\ref{table:event_count} show the number of instances of composite labels used in this study. 

We only considered a proper subset of all the possible duplex and event label combinations, on the basis of their actual occurrence. Namely, the above tables report the number of duplex and event labels associated with at least one tube instance in each of the training, validation and testing folds of each Split. 
This selection process resulted in 39 agent-action pair classes and 68 event classes, 
out of the 152 agent-action combinations and 1,620 event classes that are theoretically possible.

\subsection{Additional classes}

When defining the list of agent classes for annotion we originally included the class 
\textit{Small vehicle} which, however, does not appear in current version of the dataset (although it might appear in future extensions). Similarly, only 19 out of the 23 action classes in our list are actually present in the current version of ROAD. 

The number of instances per class for each label type is reported in a number of Tables below: 
Table~\ref{table:agent_count} (agent classes), 
Table~\ref{table:action_count} (action classes), 
Table~\ref{table:loc_count} (location classes), 
Table~\ref{table:av_action_count} (AV-action classes), 
Table~\ref{table:duplex_count} (classes of agent-action pairs) and 
Table~\ref{table:event_count} (road events).

\begin{table*}[!t]
    \centering
    \setlength{\tabcolsep}{2pt}
    \caption{Comparison of number \emph{agent} instances (excluding \textit{Autonomous-vehicle} and including the `ghost' class \textit{Small vehicle}) present at tube$/$box-level. Some self-evident abbreviations are used for class names (see Main paper, Table 1).}
    {\footnotesize
    \scalebox{0.9}{
    \begin{tabular}{lcccccccccccc}
    \toprule
    \textbf{Fold} & Pedestrian & Car & Cyclist & Mobike & SmalVeh & MedVeh & LarVeh & Bus & EmVeh & VehTL & OthTL & \textbf{Total no} 
    \\ 
    \midrule
    \em All & 228757/4212 & 120955/1318 & 63186/738 & 3163/22 & 0/0 & 29514/229 & 9534/65 & 12826/113 & 352/4 & 70199/261 & 20656/67 & 559142/7029  
    \\ 
    \em Test & 30217/645 & 17029/214 & 10587/160 & 25/1 & 0/0 & 4246/40 & 3341/14 & 831/12 & 0/0 & 14180/46 & 2009/6 & 82465/1138 
    \\ 
    \em Train-1 & 179591/3169 & 92894/949 & 46460/483 & 1017/15 & 0/0 & 15373/131 & 5560/42 & 10161/86 & 352/4 & 48329/177 & 16837/54 & 416574/5110 
    \\ 
    \em Val-1 & 18949/398 & 11032/155 & 6139/95 & 2121/6 & 0/0 & 9895/58 & 633/9 & 1834/15 & 0/0 & 7690/38 & 1810/7 & 60103/781  
    \\ 
    \em Train-2 & 169182/3138 & 84080/904 & 43337/516 & 3089/20 & 0/0 & 24014/177 & 4748/49 & 10472/90 & 352/4 & 42546/182 & 15738/50 & 397558/5130  
    \\ 
    \em Val-2 & 29358/429 & 19846/200 & 9262/62 & 49/1 & 0/0 & 1254/12 & 1445/2 & 1523/11 & 0/0 & 13473/33 & 2909/11 & 79119/761  
    \\ 
    \em Train-3 & 168572/3169 & 80023/917 & 44839/489 & 2511/18 & 0/0 & 19810/153 & 3819/30 & 9943/79 & 155/2 & 48210/176 & 15045/49 & 392927/5082  
    \\ 
    \em Val-3 & 29968/398 & 23903/187 & 7760/89 & 627/3 & 0/0 & 5458/36 & 2374/21 & 2052/22 & 197/2 & 7809/39 & 3602/12 & 83750/809  
    \\ 
    \bottomrule
    \end{tabular}
    }
    }
    \label{table:agent_count} 
\end{table*}

\begin{table*}[!t]
    \centering
    \setlength{\tabcolsep}{2pt}
    \caption{Comparison of number \emph{action} instances present at tube$/$box-level. Some self-explanatory abbreviations are used for class names (see Main paper, Table 2).} 
    {\footnotesize
    \scalebox{0.9}{
    \begin{tabular}{l cccc cccc cccc} 
    \toprule
    \multirow{2}{*}{\textbf{Fold}} & TLRed & TLAmber & TLGreen & MovAway & MovTow & Moving & Rev & Braking & Stopped & IncatLft & IncatRht & HazLit \\
      & & TurLft & TurRht & MovRht & MovLft & Ovtak & Wait2X & XingFmLft & XingFmRht & Crossing & PushObj & \textbf{Total no}\\ 
    \midrule
    \multirow{2}{*}{\em All} & 56874/183 & 7165/164 & 26134/251 & 163856/2271 & 142771/3155 & 11896/309 & 469/4 & 21812/197 & 115014/1359 & 6857/94 & 8113/114 & 3822/45 \\
      & & 5541/167 & 7546/206 & 515/17 & 473/11 & 1119/21 & 12717/247 & 21574/442 & 14957/354 & 5720/133 & 6252/103 & 641197/9847  \\ 
    \multirow{2}{*}{\em Test} & 11159/33 & 976/22 & 4014/36 & 22143/355 & 17628/460 & 2527/85 & 284/1 & 2757/13 & 17215/204 & 1810/11 & 706/13 & 682/3 \\
      & & 816/31 & 984/35 & 148/3 & 90/2 & 170/2 & 2203/40 & 3785/83 & 2784/69 & 583/19 & 1727/34 & 95191/1554 \\ 
    \multirow{2}{*}{\em Train-1} & 39406/132 & 5808/130 & 19310/178 & 127014/1697 & 111222/2370 & 7782/180 & 39/2 & 17958/161 & 80904/990 & 3163/73 & 5392/87 & 2140/34 \\
      & & 3953/109 & 5858/147 & 253/10 & 230/6 & 747/15 & 9695/191 & 15687/305 & 10813/250 & 4800/106 & 3896/58 & 476070/7231   \\ 
    \multirow{2}{*}{\em Val-1} & 6309/18 & 381/12 & 2810/37 & 14699/219 & 13921/325 & 1587/44 & 146/1 & 1097/23 & 16895/165 & 1884/10 & 2015/14 & 1000/8 \\
      & & 772/27 & 704/24 & 114/4 & 153/3 & 202/4 & 819/16 & 2102/54 & 1360/35 & 337/8 & 629/11 & 69936/1062  \\ 
    \multirow{2}{*}{\em Train-2} & 35548/118 & 4644/113 & 17450/179 & 119567/1697 & 106022/2352 & 7829/204 & 185/3 & 15565/148 & 81226/971 & 4503/72 & 6344/92 & 2985/41 \\ & & 3533/112 & 5739/155 & 367/14 & 284/8 & 949/19 & 8363/175 & 15607/318 & 11360/266 & 4906/111 & 3789/59 & 456765/7227  \\ 
    \multirow{2}{*}{\em Val-2} & 10167/32 & 1545/29 & 4670/36 & 22146/219 & 19121/343 & 1540/20 & 0/0 & 3490/36 & 16573/184 & 544/11 & 1063/9 & 155/1 \\ 
      & & 1192/24 & 823/16 & 0/0 & 99/1 & 0/0 & 2151/32 & 2182/41 & 813/19 & 231/3 & 736/10 & 89241/1066  \\ 
    \multirow{2}{*}{\em Train-3} & 40301/122 & 5304/122 & 17508/178 & 113082/1659 & 106515/2370 & 8675/202 & 185/3 & 12404/142 & 76562/966 & 3744/61 & 6866/88 & 2591/39 \\ & & 4162/116 & 6253/164 & 334/11 & 383/9 & 821/17 & 8658/177 & 16157/306 & 10674/254 & 4966/110 & 4068/63 & 450213/7179  \\ 
    \multirow{2}{*}{\em Val-3} & 5414/28 & 885/20 & 4612/37 & 28631/257 & 18628/325 & 694/22 & 0/0 & 6651/42 & 21237/189 & 1303/22 & 541/13 & 549/3 \\ 
    & & 563/20 & 309/7 & 33/3 & 0/0 & 128/2 & 1856/30 & 1632/53 & 1499/31 & 171/4 & 457/6 & 95793/1114  \\ 
 \bottomrule
    \end{tabular}
    }
    }
    \label{table:action_count} 
\end{table*}

\begin{table*}[!t] 
    \centering
    \setlength{\tabcolsep}{2pt}
    \caption{Comparison of number \emph{location} instances present at tube$/$box-level. Some self-explanatory abbreviations are used for class names (see Main paper, Table 3).}
    {\footnotesize
    \scalebox{0.85}{
    \begin{tabular}{lcc cccc cccc ccc}
    \toprule
    \textbf{Fold} & VehLane & OutgoLane & OutgoCycLane & IncomLane & IncomCycLane & Pav & LftPav & RhtPav & At junction & At crossing & BusStop & At parking & \textbf{Total}
    \\ 
    \midrule
\em All & 93984/437 & 14816/155 & 12041/116 & 73168/1307 & 7355/177 & 19646/442 & 104741/1929 & 77702/1811 & 76485/1306 & 14261/274 & 3599/66 & 768/20 & 498566/8040  
\\ 
\em Test & 13966/57 & 669/22 & 2909/29 & 9414/177 & 1587/41 & 3516/108 & 9950/229 & 11713/281 & 12937/276 & 3618/68 & 174/6 & 20/1 & 70473/1295  
\\ 
\em Train-1 & 68123/341 & 12096/121 & 8112/78 & 52842/948 & 5141/120 & 14178/283 & 86065/1517 & 58897/1359 & 51765/852 & 10501/204 & 3097/55 & 682/16 & 371499/5894  
\\ 
\em Val-1 & 11895/39 & 2051/12 & 1020/9 & 10912/182 & 627/16 & 1952/51 & 8726/183 & 7092/171 & 11783/178 & 142/2 & 328/5 & 66/3 & 56594/851 
\\ 
\em Train-2 & 64568/305 & 13799/118 & 7581/74 & 49471/906 & 5282/127 & 13992/306 & 78573/1482 & 56596/1339 & 57373/959 & 9859/195 & 3223/56 & 660/14 & 360977/5881  
\\ 
\em Val-2 & 15450/75 & 348/15 & 1551/13 & 14283/224 & 486/9 & 2138/28 & 16218/218 & 9393/191 & 6175/71 & 784/11 & 202/4 & 88/5 & 67116/864  
\\ 
\em Train-3 & 58645/320 & 10078/102 & 8109/72 & 52367/949 & 4618/109 & 13370/258 & 79053/1515 & 55755/1375 & 55566/877 & 9740/181 & 2960/54 & 748/19 & 351009/5831  
\\ 
\em Val-3 & 21373/60 & 4069/31 & 1023/15 & 11387/181 & 1150/27 & 2760/76 & 15738/185 & 10234/155 & 7982/153 & 903/25 & 465/6 & 0/0 & 77084/914  
\\ 
 \bottomrule
    \end{tabular}
    }
    }
    \label{table:loc_count} 
\end{table*}

\begin{table*}[!t]
    \centering
    \setlength{\tabcolsep}{2pt}
    \caption{Comparison of number \emph{AV-actions} instances present at tube$/$box-level.}
    {\footnotesize
    \scalebox{0.9}{
    \begin{tabular}{lcccccccc}
    \toprule
    \textbf{Fold} & AV-stop & AV-move & AV-TurRht & AV-TurLft & AV-MovRht & AV-MovLft & AV-Ovtak & \textbf{Total}
    \\ 
    \midrule
\em All & 31801/108 & 81196/233 & 3826/50 & 3787/56 & 408/16 & 537/15 & 599/12 & 122154/490  
\\ 
\em Test & 7290/20 & 11769/41 & 724/9 & 438/7 & 18/2 & 42/1 & 93/2 & 20374/82  
\\ 
\em Train-1 & 18958/75 & 58677/162 & 2496/34 & 2437/38 & 327/12 & 477/12 & 479/8 & 83851/341  
\\ 
\em Val-1 & 5553/13 & 10750/30 & 606/7 & 912/11 & 63/2 & 18/2 & 27/2 & 17929/67  
\\ 
\em Train-2 & 18522/75 & 58383/165 & 2649/35 & 2888/41 & 360/13 & 471/13 & 506/10 & 83779/352  
\\ 
\em Val-2 & 5989/13 & 11044/27 & 453/6 & 461/8 & 30/1 & 24/1 & 0/0 & 18001/56  
\\ 
\em Train-3 & 20519/68 & 57958/150 & 2580/33 & 2848/42 & 357/11 & 405/12 & 413/7 & 85080/323  
\\ 
\em Val-3 & 3992/20 & 11469/42 & 522/8 & 501/7 & 33/3 & 90/2 & 93/3 & 16700/85  
\\ 
   \bottomrule
    \end{tabular}
    }
    }
    \label{table:av_action_count} 
\end{table*}

\begin{table*}[!t]
    \centering
    \setlength{\tabcolsep}{2pt}
    \caption{Comparison of number \emph{duplex} (agent-action pair) instances present at tube$/$box-level. }
    {\footnotesize
    \scalebox{0.9}{
    \begin{tabular}{lcccccccc}
    \toprule
    \textbf{Class pair \textbackslash Split} & \em All & \em Test & \em Train-1 & \em Val-1 & \em Train-2 & \em Val-2 & \em Train-3 & \em Val-3 
    \\ 
    \midrule
Bus-MovAway & 3365/28 & 73/3 & 2669/20 & 623/5 & 2506/23 & 786/2 & 2553/16 & 739/9\\
Bus-MovTow & 3556/49 & 224/4 & 3049/41 & 283/4 & 3109/41 & 223/4 & 2686/36 & 646/9\\
Bus-Stop & 4419/37 & 399/3 & 3345/28 & 675/6 & 3819/32 & 201/2 & 3536/30 & 484/4\\
Bus-XingFmLft & 851/16 & 43/1 & 531/9 & 277/6 & 698/13 & 110/2 & 685/11 & 123/4\\
Car-Brake & 18661/159 & 2248/10 & 15747/138 & 666/11 & 13485/121 & 2928/28 & 10745/112 & 5668/37\\
Car-IncatLft & 3715/61 & 1567/6 & 1633/50 & 515/5 & 1807/48 & 341/7 & 1474/38 & 674/17\\
Car-IncatRht & 4686/86 & 582/10 & 3816/68 & 288/8 & 3760/70 & 344/6 & 3693/67 & 411/9\\
Car-MovAway & 37398/340 & 5420/65 & 30281/245 & 1697/30 & 26344/242 & 5634/33 & 22962/225 & 9016/50\\
Car-MovTow & 30134/677 & 4460/125 & 22701/486 & 2973/66 & 21086/475 & 4588/77 & 21922/475 & 3752/77\\
Car-Stop & 44975/435 & 5136/34 & 34742/352 & 5097/49 & 30756/298 & 9083/103 & 29942/328 & 9897/73\\
Car-TurLft & 2994/91 & 418/17 & 2255/62 & 321/12 & 1909/61 & 667/13 & 2275/64 & 301/10\\
Car-TurRht & 4444/123 & 456/19 & 3671/93 & 317/11 & 3653/97 & 335/7 & 3736/98 & 252/6\\
Car-XingFmLft & 6046/161 & 1387/37 & 4034/106 & 625/18 & 4451/113 & 208/11 & 3832/93 & 827/31\\
Car-XingFmRht & 4069/117 & 626/19 & 2785/83 & 658/15 & 3348/93 & 95/5 & 3050/87 & 393/11\\
Cyc-MovAway & 27848/236 & 4808/56 & 20949/163 & 2091/17 & 17838/142 & 5202/38 & 19617/150 & 3423/30\\
Cyc-MovTow & 15164/407 & 2595/73 & 10736/282 & 1833/52 & 11377/307 & 1192/27 & 10667/290 & 1902/44\\
Cyc-Stop & 16021/139 & 2254/26 & 12046/94 & 1721/19 & 11235/97 & 2532/16 & 11465/92 & 2302/21\\
Cyc-TurLft & 1463/49 & 248/10 & 1043/32 & 172/7 & 838/30 & 377/9 & 1151/34 & 64/5\\
Cyc-XingFmLft & 1776/41 & 395/9 & 1048/22 & 333/10 & 1200/29 & 181/3 & 1311/31 & 70/1\\
Cyc-XingFmRht & 2458/60 & 544/17 & 1753/36 & 161/7 & 1759/39 & 155/4 & 1851/41 & 63/2\\
LarVeh-Stop & 4131/32 & 1144/6 & 2715/22 & 272/4 & 2490/25 & 497/1 & 1556/14 & 1431/12\\
MedVeh-IncatLft & 1756/13 & 62/1 & 339/8 & 1355/4 & 1621/11 & 73/1 & 1652/8 & 42/4\\
MedVeh-MovTow & 6360/119 & 1141/20 & 3901/71 & 1318/28 & 4918/93 & 301/6 & 4311/83 & 908/16\\
MedVeh-Stop & 13496/86 & 1906/11 & 6412/49 & 5178/26 & 10657/67 & 933/8 & 9146/60 & 2444/15\\
MedVeh-TurRht & 1075/34 & 197/6 & 561/17 & 317/11 & 849/27 & 29/1 & 821/27 & 57/1\\
OthTL-Green & 5753/27 & 599/2 & 4631/22 & 523/3 & 4105/22 & 1049/3 & 3941/19 & 1213/6\\
OthTL-Red & 13962/55 & 1372/6 & 11303/44 & 1287/5 & 10787/39 & 1803/10 & 10407/38 & 2183/11\\
Ped-Mov & 11896/309 & 2527/85 & 7782/180 & 1587/44 & 7829/204 & 1540/20 & 8675/202 & 694/22\\
Ped-MovAway & 84177/1574 & 10334/217 & 66719/1212 & 7124/145 & 64267/1213 & 9576/144 & 60948/1208 & 12895/149\\
Ped-MovTow & 85079/1866 & 8189/229 & 69771/1468 & 7119/169 & 64122/1409 & 12768/228 & 65931/1464 & 10959/173\\
Ped-PushObj & 6252/103 & 1727/34 & 3896/58 & 629/11 & 3789/59 & 736/10 & 4068/63 & 457/6\\
Ped-Stop & 29338/618 & 6376/124 & 20670/437 & 2292/57 & 19635/440 & 3327/54 & 19028/433 & 3934/61\\
Ped-Wait2X & 12278/239 & 2071/37 & 9388/186 & 819/16 & 8056/170 & 2151/32 & 8437/174 & 1770/28\\
Ped-Xing & 5508/130 & 583/19 & 4588/103 & 337/8 & 4780/109 & 145/2 & 4754/107 & 171/4\\
Ped-XingFmLft & 11537/185 & 1427/22 & 9663/154 & 447/9 & 8427/138 & 1683/25 & 9590/148 & 520/15\\
Ped-XingFmRht & 6936/134 & 1364/24 & 5192/102 & 380/8 & 5108/103 & 464/7 & 4606/95 & 966/15\\
TL-Amber & 6404/144 & 976/22 & 5047/110 & 381/12 & 3940/94 & 1488/28 & 4749/108 & 679/14\\
TL-Green & 20381/224 & 3415/34 & 14679/156 & 2287/34 & 13345/157 & 3621/33 & 13567/159 & 3399/31\\
TL-Red & 42912/131 & 9787/28 & 28103/90 & 5022/13 & 24761/81 & 8364/22 & 29894/86 & 3231/17
\\
\midrule
\textbf{Total} & 641879/9863 & 95231/1558 & 476712/7243 & 69936/1062 & 457407/7239 & 89241/1066 & 450355/7189 & 96293/1116
\\
  \bottomrule
    \end{tabular}
    }
    }
    \label{table:duplex_count} 
\end{table*}

\begin{table*}[!t]
    \centering
    \setlength{\tabcolsep}{2pt}
    \caption{Comparison of number \emph{event} instances present at tube$/$box-level.}
    {\footnotesize
    \scalebox{1}{
    \begin{tabular}{lcccccccc}
    \toprule
    \textbf{Event label \textbackslash Split} & \em All & \em Test & \em Train-1 & \em Val-1 & \em Train-2 & \em Val-2 & \em Train-3 & \em Val-3
    \\ 
    \midrule
Bus-MovTow-IncomLane & 2858/37 & 204/3 & 2384/31 & 270/3 & 2487/30 & 167/4 & 2110/27 & 544/7\\
Bus-MovTow-Jun & 751/18 & 47/2 & 654/15 & 50/1 & 648/15 & 56/1 & 510/12 & 194/4\\
Bus-Stop-IncomLane & 1267/16 & 94/1 & 1082/14 & 91/1 & 1129/14 & 44/1 & 919/13 & 254/2\\
Bus-Stop-VehLane & 1351/9 & 305/2 & 735/5 & 311/2 & 889/6 & 157/1 & 864/5 & 182/2\\
Bus-XingFmLft-Jun & 851/16 & 43/1 & 531/9 & 277/6 & 698/13 & 110/2 & 685/11 & 123/4\\
Car-Brake-Jun & 2679/37 & 18/2 & 2313/28 & 348/7 & 2613/33 & 48/2 & 1518/27 & 1143/8\\
Car-Brake-VehLane & 15494/108 & 2230/8 & 12774/90 & 490/10 & 10446/81 & 2818/19 & 7839/70 & 5425/30\\
Car-IncatLft-Jun & 1843/36 & 716/7 & 1005/27 & 122/2 & 940/25 & 187/4 & 596/18 & 531/11\\
Car-IncatLft-VehLane & 2213/22 & 1454/3 & 457/17 & 302/2 & 672/15 & 87/4 & 582/12 & 177/7\\
Car-IncatRht-IncomLane & 1096/26 & 44/2 & 871/17 & 181/7 & 1008/21 & 44/3 & 859/21 & 193/3\\
Car-IncatRht-Jun & 3110/65 & 297/7 & 2706/55 & 107/3 & 2582/53 & 231/5 & 2685/52 & 128/6\\
Car-MovAway-Jun & 9770/236 & 1454/53 & 7371/162 & 945/21 & 7503/169 & 813/14 & 6325/139 & 1991/44\\
Car-MovAway-OutgoLane & 3037/67 & 189/9 & 2766/54 & 82/4 & 2719/53 & 129/5 & 2229/46 & 619/12\\
Car-MovAway-VehLane & 27319/179 & 4106/29 & 22492/139 & 721/11 & 18399/125 & 4814/25 & 15205/123 & 8008/27\\
Car-MovTow-IncomLane & 25237/543 & 3573/89 & 18947/394 & 2717/60 & 17464/380 & 4200/74 & 18061/387 & 3603/67\\
Car-MovTow-Jun & 6529/238 & 1349/58 & 4841/165 & 339/15 & 4701/164 & 479/16 & 4841/161 & 339/19\\
Car-Stop-IncomLane & 16995/247 & 1281/19 & 13100/199 & 2614/29 & 9322/142 & 6392/86 & 13168/192 & 2546/36\\
Car-Stop-Jun & 10641/98 & 1274/7 & 8175/76 & 1192/15 & 8153/83 & 1214/8 & 7189/73 & 2178/18\\
Car-Stop-VehLane & 17083/92 & 3577/12 & 12468/75 & 1038/5 & 11348/70 & 2158/10 & 8249/59 & 5257/21\\
Car-TurLft-Jun & 2413/85 & 348/18 & 1782/55 & 283/12 & 1463/55 & 602/12 & 1797/57 & 268/10\\
Car-TurLft-VehLane & 913/28 & 105/3 & 775/23 & 33/2 & 722/23 & 86/2 & 724/22 & 84/3\\
Car-TurRht-IncomLane & 295/22 & 73/6 & 210/14 & 12/2 & 208/15 & 14/1 & 168/15 & 54/1\\
Car-TurRht-Jun & 3905/115 & 344/16 & 3256/88 & 305/11 & 3284/93 & 277/6 & 3363/94 & 198/5\\
Car-XingFmLft-Jun & 5953/157 & 1387/37 & 3941/102 & 625/18 & 4442/113 & 124/7 & 3739/89 & 827/31\\
Cyc-MovAway-Jun & 5221/123 & 772/32 & 3917/80 & 532/11 & 3228/76 & 1221/15 & 4264/84 & 185/7\\
Cyc-MovAway-LftPav & 1318/14 & 85/2 & 1219/11 & 14/1 & 545/9 & 688/3 & 755/7 & 478/5\\
Cyc-MovAway-OutgoCycLane & 9957/99 & 2388/30 & 6907/63 & 662/6 & 6303/56 & 1266/13 & 6550/55 & 1019/14\\
Cyc-MovAway-OutgoLane & 788/24 & 71/5 & 679/18 & 38/1 & 711/18 & 6/1 & 501/14 & 216/5\\
Cyc-MovAway-VehLane & 13103/91 & 1414/11 & 10463/73 & 1226/7 & 9272/59 & 2417/21 & 10018/67 & 1671/13\\
Cyc-MovTow-IncomCycLane & 5487/140 & 1479/37 & 3728/93 & 280/10 & 3828/98 & 180/5 & 3006/80 & 1002/23\\
Cyc-MovTow-IncomLane & 7031/208 & 645/18 & 5133/152 & 1253/38 & 5541/172 & 845/18 & 5809/173 & 577/17\\
Cyc-MovTow-Jun & 2472/83 & 546/19 & 1618/57 & 308/7 & 1827/62 & 99/2 & 1820/59 & 106/5\\
Cyc-MovTow-LftPav & 699/16 & 27/3 & 625/11 & 47/2 & 629/12 & 43/1 & 432/10 & 240/3\\
Cyc-Stop-IncomCycLane & 680/19 & 73/3 & 533/14 & 74/2 & 517/14 & 90/2 & 534/13 & 73/3\\
Cyc-Stop-IncomLane & 1376/30 & 20/1 & 932/21 & 424/8 & 1282/27 & 74/2 & 871/19 & 485/10\\
Cyc-Stop-Jun & 5383/49 & 1568/16 & 2838/26 & 977/7 & 3693/32 & 122/1 & 3641/28 & 174/5\\
Cyc-TurLft-Jun & 1211/43 & 189/8 & 873/29 & 149/6 & 684/27 & 338/8 & 958/30 & 64/5\\
Cyc-XingFmLft-Jun & 1347/33 & 345/8 & 669/15 & 333/10 & 943/24 & 59/1 & 932/24 & 70/1\\
MedVeh-MovTow-IncomLane & 5386/101 & 1019/18 & 3218/58 & 1149/25 & 4121/79 & 246/4 & 3459/67 & 908/16\\
MedVeh-MovTow-Jun & 1176/42 & 254/8 & 543/20 & 379/14 & 817/31 & 105/3 & 892/31 & 30/3\\
MedVeh-Stop-IncomLane & 4004/44 & 399/6 & 2388/29 & 1217/9 & 2675/29 & 930/9 & 3003/30 & 602/8\\
MedVeh-Stop-Jun & 1940/18 & 41/2 & 916/10 & 983/6 & 1298/13 & 601/3 & 1871/15 & 28/1\\
MedVeh-TurRht-Jun & 898/32 & 114/6 & 491/15 & 293/11 & 761/25 & 23/1 & 733/25 & 51/1\\
Ped-Mov-Pav & 9549/227 & 2353/77 & 5735/113 & 1461/37 & 5761/133 & 1435/17 & 6502/128 & 694/22\\
Ped-MovAway-LftPav & 44539/750 & 4689/90 & 35787/588 & 4063/72 & 34461/592 & 5389/68 & 31522/586 & 8328/74\\
Ped-MovAway-Pav & 2662/100 & 385/15 & 2254/81 & 23/4 & 1947/75 & 330/10 & 1618/63 & 659/22\\
Ped-MovAway-RhtPav & 31203/711 & 4807/107 & 23525/534 & 2871/70 & 22796/538 & 3600/66 & 23574/553 & 2822/51\\
Ped-MovTow-IncomLane & 473/10 & 186/2 & 282/7 & 5/1 & 233/6 & 54/2 & 252/7 & 35/1\\
Ped-MovTow-LftPav & 43711/905 & 3996/99 & 36114/721 & 3601/85 & 32188/686 & 7527/120 & 35078/725 & 4637/81\\
Ped-MovTow-RhtPav & 34452/844 & 3603/112 & 27615/653 & 3234/79 & 25681/624 & 5168/108 & 25759/659 & 5090/73\\
Ped-MovTow-VehLane & 595/15 & 61/2 & 449/11 & 85/2 & 515/12 & 19/1 & 467/12 & 67/1\\
Ped-PushObj-LftPav & 2745/41 & 509/7 & 1797/27 & 439/7 & 1708/28 & 528/6 & 2038/31 & 198/3\\
Ped-PushObj-RhtPav & 1637/46 & 709/19 & 803/25 & 125/2 & 794/23 & 134/4 & 843/25 & 85/2\\
Ped-Stop-BusStop & 1746/48 & 174/6 & 1553/41 & 19/1 & 1448/39 & 124/3 & 1107/36 & 465/6\\
Ped-Stop-LftPav & 11149/248 & 1054/33 & 9194/190 & 901/25 & 7832/185 & 2263/30 & 8369/192 & 1726/23\\
Ped-Stop-Pav & 4059/69 & 459/8 & 3203/51 & 397/10 & 3227/57 & 373/4 & 3364/54 & 236/7\\
Ped-Stop-RhtPav & 9335/222 & 2768/48 & 5683/151 & 884/23 & 6144/159 & 423/15 & 4957/147 & 1610/27\\
Ped-Stop-VehLane & 1171/16 & 501/4 & 586/10 & 84/2 & 437/8 & 233/4 & 519/11 & 151/1\\
Ped-Wait2X-LftPav & 5165/86 & 301/5 & 4284/73 & 580/8 & 3234/60 & 1630/21 & 3964/73 & 900/8\\
Ped-Wait2X-RhtPav & 2045/58 & 218/2 & 1761/52 & 66/4 & 1539/46 & 288/10 & 1147/39 & 680/17\\
Ped-XingFmLft-IncomLane & 1283/35 & 65/2 & 1191/32 & 27/1 & 606/17 & 612/16 & 1135/32 & 83/1\\
Ped-XingFmLft-Jun & 1933/26 & 76/2 & 1516/19 & 341/5 & 1795/22 & 62/2 & 1803/23 & 54/1\\
Ped-XingFmLft-VehLane & 1386/36 & 77/2 & 1214/31 & 95/3 & 738/17 & 571/17 & 1238/33 & 71/1\\
Ped-XingFmLft-xing & 6828/119 & 1105/18 & 5581/99 & 142/2 & 5407/96 & 316/5 & 5418/89 & 305/12\\
Ped-XingFmRht-IncomLane & 513/15 & 115/3 & 305/9 & 93/3 & 289/9 & 109/3 & 241/8 & 157/4\\
Ped-XingFmRht-Jun & 1135/28 & 144/3 & 939/22 & 52/3 & 928/23 & 63/2 & 935/24 & 56/1\\
Ped-XingFmRht-RhtPav & 195/9 & 73/3 & 119/5 & 3/1 & 98/5 & 24/1 & 32/3 & 90/3\\
Ped-XingFmRht-VehLane & 1037/24 & 156/5 & 696/16 & 185/3 & 795/16 & 86/3 & 773/17 & 108/2
\\
\midrule
\textbf{Total} & 594040/10849 & 85058/1738 & 439196/7941 & 69786/1170 & 430026/7976 & 78956/1135 & 417315/7880 & 91667/1231\\
    \bottomrule
    \end{tabular}
    }
    }
    \label{table:event_count} 
\end{table*}

\section{ADDITIONAL RESULTS}

\REBUT{Here we report both the complete class-wise results for each task, and some qualitative results showing success and failure modes of our 3D-RetinaNet baseline}.

\subsection{Class-wise results}

We provide class-wise detection results for all label types (simple and composite) under the different splits. Table~\ref{tab:classwise-splitwise-primary-labels-supp} shows the class-wise and split-wise results for individual labels. Class-wise and split-wise results for duplex and event labels are given in Table \ref{tab:classwise-splitwise-duplex} and Table~\ref{tab:classwise-splitwise-triplets}, respectively.

Similarly, a class-wise comparison of the results averaged over the three training split for the joint and the product of marginals approaches is proposed in Tables \ref{tab:classwise-marginals-duplexes} and \ref{tab:classwise-marginals-event} for duplex and event detection, respectively.

\subsection{Qualitative results}

\begin{figure*}[ht!]
\centering{
    \includegraphics[width=1.0 \textwidth]{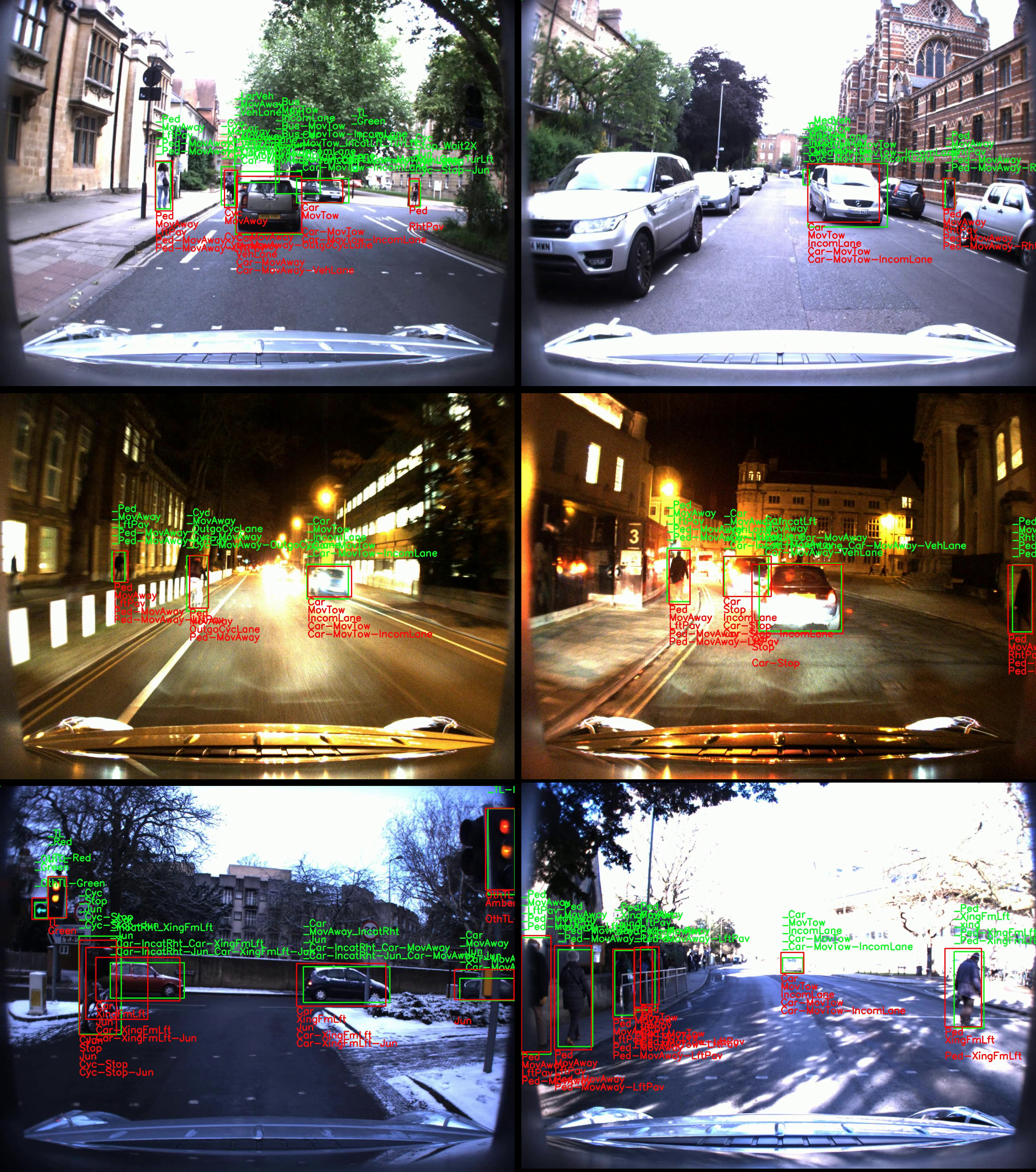}
\caption{\REBUT{Success cases in which our model detects the actions and locations correctly, and only for those agents which are active. Ground truth bounding boxes and labels are shown in green, while the predictions of our model are shown in red. 
}}
\label{fig:Success}
}
\end{figure*}

\REBUT{Finally, we provide some qualitative results of our baseline model in terms of success and failure modes. Cases in which the baseline work accurately are illustrated in Figure~\ref{fig:Success}, where the model is shown to detect only those agents which are active (i.e., are performing some actions) and ignore all the inactive agents (namely, parked vehicles). Agent prediction is very stable across all the examples, whereas action and location prediction show some weakness in some case: for instance, the night-time example in the second row of the second column, where both the cars in front are moving away in the outgoing lane but our method fails to label their location correctly.}

\begin{figure*}[ht!]
\centering{
    \includegraphics[width=1.0 \textwidth]{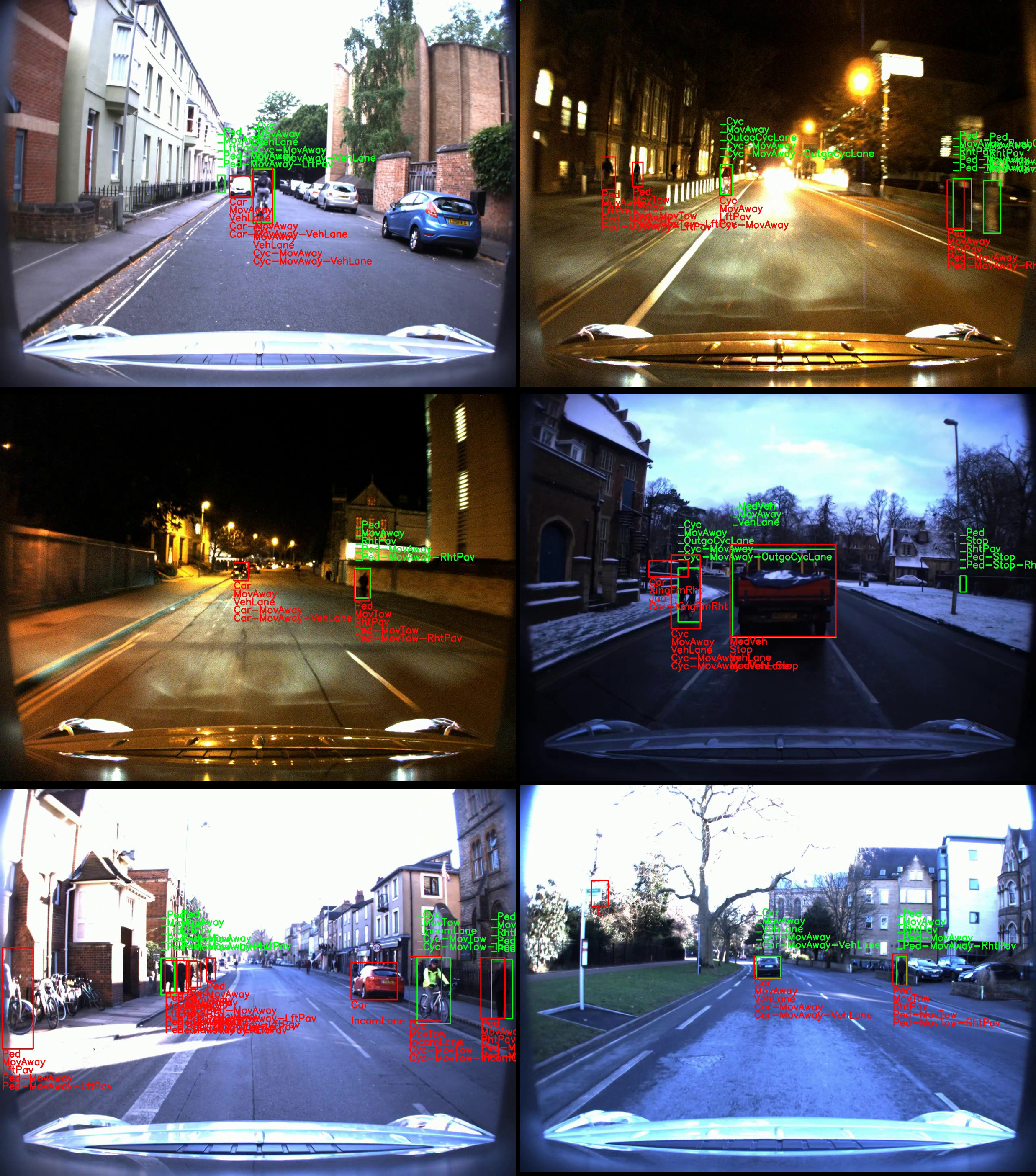}
\caption{\REBUT{Some of the failure modes of our model, shown detecting  inactive agents and/or assigning incorrect action and location labels. Ground truth bounding boxes and labels are shown in green, while the predictions of our model are shown in red.
}}
\label{fig:failure}
}
\end{figure*}

\REBUT{In contrast, the failure modes illustrated in Figure~\ref{fig:failure} are cases in which the model fails to assign to agents the correct label, and also detects agents which are not active (e.g. often parked cars, see the white vehicle in the top row, first column or the red vehicle in the third row, first column). 
}

\begin{table*}[t]
\centering
\setlength{\tabcolsep}{4pt}
\caption{Number of video- and frame-level instances for each individual label on the left. 
Corresponding frame-/video-level results (mAP@$\%$) for each of the three ROAD splits (right). {Val-$n$ denotes the validation set for Split $n$}. The results presented here are generated with I3D backbone.} 
{\footnotesize
\scalebox{0.75}{
\begin{tabular}{lccccc | cccccc cc}
\toprule
& \multicolumn{5}{c}{\textbf{Number of instances}}  & \multicolumn{8}{c}{\textbf{Frame-mAP@$0.5$/Video-mAP@$0.2$}} \\ 
\midrule
& \multicolumn{5}{c}{\#Boxes/\#Tubes} & \multicolumn{2}{c}{Train-1}  & \multicolumn{2}{c}{Train-2}  & \multicolumn{2}{c}{Train-3} & \multicolumn{2}{c}{Averaged} \\ 
\midrule 
Eval subset & \em All & \em Val-1 & \em Val-2  & \em Val-3 & \em Test  & \em Val-1 & \em Test  & \em Val-2 & \em Test  & \em Val-3 & \em Test & \em Val & \em Test \\ 
\midrule
\multicolumn{2}{l}{\textbf{Agent detection}} \\ 
\midrule
Ped                       & 228757/4212  & 18949/398    & 29358/429    & 29968/398    & 30217/645    & 66.3/51.0    & 55.4/33.7    & 29.8/20.8    & 54.6/34.9    & 41.2/31.7    & 56.2/33.5   & 45.8/34.5    & 55.4/34.0   \\ 
Car                       & 120955/1318  & 11032/155    & 19846/200    & 23903/187    & 17029/214    & 75.1/68.6    & 78.6/64.4    & 38.4/38.3    & 60.0/57.8    & 69.3/55.2    & 76.3/60.6   & 60.9/54.0    & 71.7/60.9   \\ 
Tl                        & 70199/261    & 7690/38      & 13473/33     & 7809/39      & 14180/46     & 86.7/7.6     & 65.6/14.6    & 28.4/4.4     & 63.8/16.7    & 47.0/27.3    & 67.8/16.3   & 54.0/13.1    & 65.7/15.9   \\ 
Cyc                       & 63186/738    & 6139/95      & 9262/62      & 7760/89      & 10587/160    & 72.4/43.3    & 68.4/44.1    & 41.2/20.0    & 65.2/36.8    & 47.1/30.7    & 71.0/44.1   & 53.6/31.4    & 68.2/41.7   \\ 
Medveh                    & 29514/229    & 9895/58      & 1254/12      & 5458/36      & 4246/40      & 27.3/43.5    & 31.4/52.0    & 1.1/4.5      & 41.7/36.2    & 16.4/37.9    & 66.5/49.7   & 14.9/28.7    & 46.5/46.0   \\ 
Othtl                     & 20656/67     & 1810/7       & 2909/11      & 3602/12      & 2009/6       & 1.5/15.3     & 3.6/0.2      & 0.2/0.0      & 72.4/28.0    & 24.9/1.4     & 39.7/7.4    & 8.8/5.6      & 38.5/11.8   \\ 
Bus                       & 12826/113    & 1834/15      & 1523/11      & 2052/22      & 831/12       & 66.9/63.2    & 23.9/31.8    & 33.1/71.8    & 22.8/35.1    & 41.6/41.5    & 28.5/31.0   & 47.2/58.8    & 25.1/32.6   \\ 
Larveh                    & 9534/65      & 633/9        & 1445/2       & 2374/21      & 3341/14      & 2.0/4.6      & 13.3/16.2    & 0.0/0.0      & 27.9/27.3    & 19.7/31.8    & 19.6/32.0   & 7.3/12.1     & 20.3/25.1   \\ 
Mobike                    & 3163/22      & 2121/6       & 49/1         & 627/3        & 25/1         & 47.2/3.4     & 0.1/0.1      & 0.0/0.0      & 0.5/1.7      & 45.6/13.0    & 0.3/0.5     & 30.9/5.5     & 0.3/0.8     \\ 
Emveh                     & 352/4        & 0/0          & 0/0          & 197/2        & 0/0          & 0.0/0.0      & 0.0/0.0      & 0.0/0.0      & 0.0/0.0      & 0.1/0.3      & 0.0/0.0     & 0.0/0.1      & 0.0/0.0     \\ 
\midrule
\emph{Total/Mean} & 559142/7029  & 60103/781    & 79119/761    & 83750/809    & 82465/1138   & 44.5/30.1    & 34.0/25.7    & 17.2/16.0    & 40.9/27.4    & 35.3/27.1    & 42.6/27.5   & 32.3/24.4    & 39.2/26.9   \\ 
\midrule
\multicolumn{2}{l}{\textbf{Action detection}} \\ 
\midrule
Movaway                   & 163856/2271  & 14699/219    & 22146/219    & 28631/257    & 22143/355    & 41.9/30.9    & 48.9/23.0    & 18.4/12.5    & 43.2/26.0    & 41.7/23.6    & 44.0/23.4   & 34.0/22.3    & 45.4/24.1    \\ 
Movtow                    & 142771/3155  & 13921/325    & 19121/343    & 18628/325    & 17628/460    & 58.5/58.3    & 49.7/49.1    & 22.7/27.2    & 46.9/44.6    & 42.6/48.0    & 50.1/44.4   & 41.2/44.5    & 48.9/46.0    \\ 
Stop                      & 115014/1359  & 16895/165    & 16573/184    & 21237/189    & 17215/204    & 38.3/27.3    & 39.3/13.5    & 19.7/16.7    & 34.6/16.4    & 40.7/24.6    & 40.2/17.4   & 32.9/22.9    & 38.0/15.8    \\ 
Red                       & 56874/183    & 6309/18      & 10167/32     & 5414/28      & 11159/33     & 74.0/20.0    & 65.7/18.0    & 18.3/2.6     & 59.5/26.9    & 42.5/12.1    & 58.3/24.4   & 44.9/11.6    & 61.2/23.1    \\ 
Green                     & 26134/251    & 2810/37      & 4670/36      & 4612/37      & 4014/36      & 59.6/8.4     & 48.0/15.4    & 25.3/1.7     & 41.0/3.0     & 43.7/30.8    & 45.9/7.7    & 42.9/13.6    & 45.0/8.7     \\ 
Brake                     & 21812/197    & 1097/23      & 3490/36      & 6651/42      & 2757/13      & 7.2/2.0      & 36.2/16.0    & 11.3/1.7     & 29.2/22.6    & 43.0/11.4    & 31.3/27.3   & 20.5/5.0     & 32.2/22.0    \\ 
Xingfmlft                 & 21574/442    & 2102/54      & 2182/41      & 1632/53      & 3785/83      & 35.6/28.5    & 58.1/52.7    & 16.6/31.3    & 60.5/53.8    & 36.5/24.4    & 63.6/56.8   & 29.6/28.1    & 60.8/54.4    \\ 
Xingfmrht                 & 14957/354    & 1360/35      & 813/19       & 1499/31      & 2784/69      & 36.8/38.9    & 48.2/45.0    & 23.7/35.2    & 44.6/46.7    & 29.9/42.2    & 43.7/49.6   & 30.1/38.7    & 45.5/47.1    \\ 
 Wait2x                    & 12717/247    & 819/16       & 2151/32      & 1856/30      & 2203/40      & 17.1/15.8    & 12.4/5.3     & 0.7/0.1      & 8.0/8.6      & 15.4/11.1    & 10.0/9.5   & 11.1/9.0     & 10.1/7.8      \\ 
 Mov                       & 11896/309    & 1587/44      & 1540/20      & 694/22       & 2527/85      & 32.7/16.1    & 32.3/25.2    & 20.5/15.8    & 38.5/30.7    & 2.0/5.7      & 32.7/31.0  & 18.4/12.5    & 34.5/28.9     \\ 
 Incatrht                  & 8113/114     & 2015/14      & 1063/9       & 541/13       & 706/13       & 4.4/10.4     & 4.1/5.7      & 2.2/15.6     & 3.0/2.6      & 1.6/1.0      & 3.2/2.4    & 2.7/9.0      & 3.4/3.5       \\ 
 Turrht                    & 7546/206     & 704/24       & 823/16       & 309/7        & 984/35       & 11.5/9.5     & 11.8/9.9     & 18.1/25.6    & 9.8/8.2      & 5.9/4.0      & 13.0/10.0  & 11.8/13.1    & 11.5/9.4      \\ 
 Amber                     & 7165/164     & 381/12       & 1545/29      & 885/20       & 976/22       & 27.0/2.0     & 27.7/2.0     & 13.6/1.9     & 32.2/8.4     & 26.6/2.7     & 22.2/2.2   & 22.4/2.2     & 27.4/4.2      \\ 
 Incatlft                  & 6857/94      & 1884/10      & 544/11       & 1303/22      & 1810/11      & 3.4/4.4      & 6.4/8.5      & 2.3/4.2      & 7.9/4.9      & 5.3/2.2      & 11.9/9.7   & 3.7/3.6      & 8.7/7.7       \\ 
 Pushobj                   & 6252/103     & 629/11       & 736/10       & 457/6        & 1727/34      & 5.4/12.6     & 6.2/4.6      & 0.2/0.2      & 5.2/3.3      & 2.6/13.9     & 8.9/10.2   & 2.7/8.9      & 6.8/6.0       \\ 
 Xing                      & 5720/133     & 337/8        & 231/3        & 171/4        & 583/19       & 18.6/16.2    & 1.3/1.3      & 0.0/0.1      & 2.7/6.8      & 0.6/1.5      & 1.5/1.6    & 6.4/5.9      & 1.8/3.2       \\ 
 Turlft                    & 5541/167     & 772/27       & 1192/24      & 563/20       & 816/31       & 14.4/4.9     & 4.9/1.8      & 6.4/12.5     & 7.2/3.1      & 8.0/1.4      & 6.7/4.2    & 9.6/6.2      & 6.3/3.1       \\ 
 Hazlit                    & 3822/45      & 1000/8       & 155/1        & 549/3        & 682/3        & 8.6/8.6      & 4.7/33.6     & 2.9/12.5     & 5.8/11.4     & 11.8/14.0    & 1.6/7.6    & 7.8/11.7     & 4.0/17.5      \\ 
 Ovtak                     & 1119/21      & 202/4        & 0/0          & 128/2        & 170/2        & 2.7/7.8      & 0.1/0.0      & 0.0/0.0      & 0.3/0.0      & 1.5/2.3      & 0.2/0.0    & 1.4/3.4      & 0.2/0.0       \\ 
\midrule
\emph{Total/Mean}               & 639740/9815  & 69523/1054   & 89142/1065   & 95760/1111   & 94669/1548   & 26.2/17.0    & 26.6/17.4    & 11.7/11.4    & 25.3/17.3    & 21.2/14.6    & 25.7/17.9   & 19.7/14.3    & 25.9/17.5    \\ 
 \midrule
  \multicolumn{2}{l}{\textbf{Location detection}} \\ 
 \midrule
 Lftpav                    & 104741/1929  & 8726/183     & 16218/218    & 15738/185    & 9950/229     & 67.9/51.4    & 60.3/38.7    & 33.7/26.8    & 59.4/40.1    & 35.1/32.8    & 64.0/39.5    & 45.6/37.0    & 61.2/39.4    \\ 
 Vehlane                   & 93984/437    & 11895/39     & 15450/75     & 21373/60     & 13966/57     & 51.4/31.5    & 60.1/31.9    & 41.1/10.1    & 52.1/19.7    & 51.4/28.1    & 69.8/28.1    & 47.9/23.2    & 60.7/26.6    \\ 
 Rhtpav                    & 77702/1811   & 7092/171     & 9393/191     & 10234/155    & 11713/281    & 61.5/48.0    & 37.9/30.2    & 20.3/21.1    & 37.5/30.1    & 31.9/28.9    & 38.3/29.2    & 37.9/32.7    & 37.9/29.8    \\ 
 Jun                       & 76485/1306   & 11783/178    & 6175/71      & 7982/153     & 12937/276    & 44.1/37.3    & 49.0/31.5    & 11.3/20.3    & 44.9/37.8    & 27.2/25.2    & 52.1/39.1    & 27.5/27.6    & 48.7/36.1    \\ 
 Incomlane                 & 73168/1307   & 10912/182    & 14283/224    & 11387/181    & 9414/177     & 74.1/76.3    & 65.6/71.1    & 35.1/46.6    & 64.0/67.2    & 55.5/62.8    & 62.2/63.3    & 54.9/61.9    & 63.9/67.2    \\ 
 Pavement                  & 19646/442    & 1952/51      & 2138/28      & 2760/76      & 3516/108     & 35.3/21.3    & 23.7/15.9    & 13.7/6.4     & 29.5/21.6    & 3.2/4.3      & 28.3/25.3    & 17.4/10.7    & 27.2/20.9    \\ 
 Outgolane                 & 14816/155    & 2051/12      & 348/15       & 4069/31      & 669/22       & 12.9/16.8    & 2.4/0.7      & 0.1/0.4      & 2.6/0.3      & 14.0/18.5    & 3.2/0.5      & 9.0/11.9     & 2.7/0.5      \\ 
 Xing                      & 14261/274    & 142/2        & 784/11       & 903/25       & 3618/68      & 4.8/1.2      & 21.7/12.5    & 2.7/3.5      & 25.0/19.2    & 17.0/14.0    & 24.7/15.9    & 8.2/6.2      & 23.8/15.9    \\ 
 Outgocyclane              & 12041/116    & 1020/9       & 1551/13      & 1023/15      & 2909/29      & 45.2/32.3    & 57.4/26.6    & 5.1/8.6      & 55.3/42.0    & 36.6/20.7    & 56.2/48.1    & 29.0/20.5    & 56.3/38.9    \\ 
 Incomcyclane              & 7355/177     & 627/16       & 486/9        & 1150/27      & 1587/41      & 17.0/23.7    & 43.6/57.7    & 0.5/0.9      & 36.2/35.6    & 32.6/43.4    & 40.6/51.4    & 16.7/22.6    & 40.1/48.3    \\ 
 Busstop                   & 3599/66      & 328/5        & 202/4        & 465/6        & 174/6        & 4.9/3.0      & 0.6/0.5      & 0.2/0.7      & 0.5/1.4      & 0.2/0.2      & 1.0/3.2      & 1.8/1.3      & 0.7/1.7      \\ 
 Parking                   & 768/20       & 66/3         & 88/5         & 0/0          & 20/1         & 0.0/0.0      & 0.0/0.0      & 0.3/0.0      & 0.0/0.0      & 0.0/0.0      & 0.0/0.0      & 0.1/0.0      & 0.0/0.0      \\ 
\midrule
\emph{Total/Mean}                & 498566/8040  & 56594/851    & 67116/864    & 77084/914    & 70473/1295   & 34.9/28.6    & 35.2/26.4    & 13.7/12.1    & 33.9/26.3    & 25.4/23.2    & 36.7/28.6    & 24.7/21.3    & 35.3/27.1    \\ 
 \midrule
 \multicolumn{11}{l}{\textbf{AV-action segmentation}} \\ 
 \midrule
 Av-mov                   & 81196/233    & 10750/30     & 11044/27     & 11469/42     & 11769/41     & 98.8/0.0     & 97.4/0.0     & 79.1/0.0     & 94.7/0.0     & 98.3/0.0     & 97.9/0.0     & 92.0/0.0     & 96.6/0.0     \\ 
Av-stop                   & 31801/108    & 5553/13      & 5989/13      & 3992/20      & 7290/20      & 99.6/0.0     & 99.1/0.0     & 80.3/0.0     & 97.4/0.0     & 96.9/0.0     & 99.1/0.0     & 92.2/0.0     & 98.5/0.0     \\ 
Av-turrht                 & 3826/50      & 606/7        & 453/6        & 522/8        & 724/9        & 80.2/0.0     & 63.6/0.0     & 18.5/0.0     & 49.8/0.0     & 39.5/0.0     & 75.7/0.0     & 46.1/0.0     & 63.0/0.0     \\ 
Av-turlft                 & 3787/56      & 912/11       & 461/8        & 501/7        & 438/7        & 90.5/0.0     & 57.8/0.0     & 56.1/0.0     & 58.7/0.0     & 60.6/0.0     & 63.0/0.0     & 69.0/0.0     & 59.8/0.0     \\ 
Av-ovtak                  & 599/12       & 27/2         & 0/0          & 93/3         & 93/2         & 6.0/0.0      & 0.6/0.0      & 0.0/0.0      & 1.9/0.0      & 8.7/0.0      & 0.9/0.0      & 4.9/0.0      & 1.1/0.0      \\ 
Av-movlft                 & 537/15       & 18/2         & 24/1         & 90/2         & 42/1         & 0.5/0.0      & 0.8/0.0      & 0.2/0.0      & 1.4/0.0      & 0.9/0.0      & 0.4/0.0      & 0.5/0.0      & 0.8/0.0      \\ 
Av-movrht                 & 408/16       & 63/2         & 30/1         & 33/3         & 18/2         & 29.7/0.0     & 0.3/0.0      & 0.6/0.0      & 1.3/0.0      & 1.1/0.0      & 0.2/0.0      & 10.5/0.0     & 0.6/0.0      \\ 
\midrule
\emph{Total/Mean}               & 122154/490   & 17929/67     & 18001/56     & 16700/85     & 20374/82     & 57.9/0.0     & 45.7/0.0     & 33.5/0.0     & 43.6/0.0     & 43.7/0.0     & 48.2/0.0     & 45.0/0.0     & 45.8/0.0     \\
 \bottomrule
  \end{tabular}
  }
  }
\label{tab:classwise-splitwise-primary-labels-supp} 
\end{table*}

\begin{table*}[t]
\centering
\setlength{\tabcolsep}{4pt}
\caption{Number of video- and frame-level instances for duplexes (agent-action pairs), left. 
Corresponding frame-/video-level results (mAP@$\%$) for each of the three ROAD splits (right). {Val-$n$ denotes the validation set for Split $n$}. The results presented here are generated using the I3D backbone.} 
{\footnotesize
\scalebox{0.75}{
\begin{tabular}{lccccc | cccccc cc}
\toprule
& \multicolumn{5}{c}{\textbf{Number of instances}}  & \multicolumn{8}{c}{\textbf{Frame-mAP@$0.5$/Video-mAP@$0.2$}} \\ 
\midrule
& \multicolumn{5}{c}{\#Boxes/\#Tubes} & \multicolumn{2}{c}{Train-1}  & \multicolumn{2}{c}{Train-2}  & \multicolumn{2}{c}{Train-3} &  \multicolumn{2}{c}{Averaged} \\ 
\midrule 
\textbf{Class \textbackslash Eval subset} 
& \em All & \em Val-1 & \em Val-2  & \em Val-3 & \em Test  & \em Val-1 & \em Test  & \em Val-2 & \em Test  & \em Val-3 & \em Test  & \em Val & \em Test \\ 
\midrule
Ped-movtow                & 85079/1866   & 7119/169     & 12768/228    & 10959/173    & 8189/229     & 48.1/41.0    & 26.3/24.1    & 18.8/17.1    & 29.4/27.2    & 23.7/27.9    & 29.0/24.1    & 30.2/28.7    & 28.2/25.2    \\ 
Ped-movaway               & 84177/1574   & 7124/145     & 9576/144     & 12895/149    & 10334/217    & 51.0/41.0    & 34.4/20.1    & 13.0/9.8     & 35.3/23.8    & 25.8/23.6    & 31.6/21.4    & 29.9/24.8    & 33.7/21.8    \\ 
Car-stop                  & 44975/435    & 5097/49      & 9083/103     & 9897/73      & 5136/34      & 60.2/50.4    & 59.1/27.8    & 15.2/24.4    & 22.7/22.6    & 46.0/28.1    & 44.9/33.1    & 40.5/34.3    & 42.2/27.9    \\ 
Tl-red                    & 42912/131    & 5022/13      & 8364/22      & 3231/17      & 9787/28      & 90.4/18.0    & 72.5/22.8    & 20.1/7.9     & 64.6/24.7    & 55.9/18.1    & 66.6/27.9    & 55.5/14.6    & 67.9/25.1    \\ 
Car-movaway               & 37398/340    & 1697/30      & 5634/33      & 9016/50      & 5420/65      & 15.3/13.2    & 59.3/24.1    & 16.7/22.6    & 44.2/26.6    & 62.7/26.3    & 54.1/24.9    & 31.6/20.7    & 52.5/25.2    \\ 
Car-movtow                & 30134/677    & 2973/66      & 4588/77      & 3752/77      & 4460/125     & 64.0/72.5    & 64.1/66.8    & 25.0/50.4    & 56.4/54.3    & 68.4/76.7    & 63.1/60.6    & 52.5/66.5    & 61.2/60.6    \\ 
Ped-stop                  & 29338/618    & 2292/57      & 3327/54      & 3934/61      & 6376/124     & 8.9/9.5      & 11.2/5.2     & 2.7/2.2      & 14.1/10.7    & 12.1/10.9    & 11.5/8.4     & 7.9/7.5      & 12.2/8.1     \\ 
Cyc-movaway               & 27848/236    & 2091/17      & 5202/38      & 3423/30      & 4808/56      & 75.3/25.3    & 55.6/42.2    & 28.2/19.5    & 56.2/36.1    & 40.4/21.6    & 63.4/43.9    & 48.0/22.1    & 58.4/40.7    \\ 
Tl-green                  & 20381/224    & 2287/34      & 3621/33      & 3399/31      & 3415/34      & 68.6/12.1    & 38.7/15.4    & 33.0/3.2     & 38.6/4.6     & 38.9/32.0    & 40.8/3.7     & 46.8/15.8    & 39.3/7.9     \\ 
Car-brake                 & 18661/159    & 666/11       & 2928/28      & 5668/37      & 2248/10      & 11.8/4.2     & 38.1/15.4    & 10.4/0.8     & 40.0/26.2    & 43.5/11.0    & 29.0/32.9    & 21.9/5.3     & 35.7/24.9    \\ 
Cyc-stop                  & 16021/139    & 1721/19      & 2532/16      & 2302/21      & 2254/26      & 21.6/15.9    & 44.9/7.1     & 27.1/16.2    & 59.5/8.4     & 42.2/23.7    & 59.4/12.7    & 30.3/18.6    & 54.6/9.4     \\ 
Cyc-movtow                & 15164/407    & 1833/52      & 1192/27      & 1902/44      & 2595/73      & 69.3/69.5    & 51.3/55.6    & 28.2/35.3    & 35.8/38.5    & 37.8/41.9    & 43.2/46.0    & 45.1/48.9    & 43.4/46.7    \\ 
Othtl-red                 & 13962/55     & 1287/5       & 1803/10      & 2183/11      & 1372/6       & 1.0/4.1      & 2.6/1.1      & 0.1/0.0      & 80.8/45.1    & 10.0/9.1     & 24.8/9.8     & 3.7/4.4      & 36.1/18.7    \\ 
Medveh-stop               & 13496/86     & 5178/26      & 933/8        & 2444/15      & 1906/11      & 6.0/18.1     & 11.2/38.8    & 0.5/2.6      & 22.9/36.3    & 6.4/25.0     & 52.2/51.7    & 4.3/15.3     & 28.8/42.3    \\ 
Ped-wait2x                & 12278/239    & 819/16       & 2151/32      & 1770/28      & 2071/37      & 15.5/21.3    & 15.1/5.9     & 0.8/0.2      & 8.5/8.5      & 14.0/11.9    & 10.6/10.2    & 10.1/11.1    & 11.4/8.2     \\ 
Ped-mov                   & 11896/309    & 1587/44      & 1540/20      & 694/22       & 2527/85      & 32.0/17.0    & 31.4/23.5    & 20.8/15.8    & 39.3/30.1    & 2.1/5.9      & 32.2/29.2    & 18.3/12.9    & 34.3/27.6    \\ 
Ped-xingfmlft             & 11537/185    & 447/9        & 1683/25      & 520/15       & 1427/22      & 10.0/2.8     & 43.8/32.2    & 20.5/21.0    & 50.1/54.3    & 23.3/18.1    & 53.0/37.0    & 17.9/14.0    & 49.0/41.2    \\ 
Ped-xingfmrht             & 6936/134     & 380/8        & 464/7        & 966/15       & 1364/24      & 5.9/16.1     & 28.3/22.1    & 19.7/14.6    & 21.1/18.8    & 27.8/33.7    & 22.8/26.1    & 17.8/21.5    & 24.1/22.3    \\ 
Tl-amber                  & 6404/144     & 381/12       & 1488/28      & 679/14       & 976/22       & 28.1/3.3     & 29.1/1.9     & 15.4/3.1     & 33.4/9.2     & 26.3/1.7     & 22.5/3.3     & 23.3/2.7     & 28.3/4.8     \\ 
Medveh-movtow             & 6360/119     & 1318/28      & 301/6        & 908/16       & 1141/20      & 47.2/59.3    & 38.8/56.3    & 2.1/4.2      & 41.0/56.6    & 50.7/71.0    & 47.4/63.4    & 33.3/44.8    & 42.4/58.8    \\ 
Ped-pushobj               & 6252/103     & 629/11       & 736/10       & 457/6        & 1727/34      & 6.3/13.0     & 5.9/4.7      & 0.1/0.1      & 5.3/3.9      & 2.6/23.6     & 8.7/9.9      & 3.0/12.2     & 6.7/6.2      \\ 
Car-xingfmlft             & 6046/161     & 625/18       & 208/11       & 827/31       & 1387/37      & 48.6/45.8    & 76.7/76.6    & 23.0/46.5    & 77.3/74.2    & 50.3/31.9    & 82.4/82.1    & 40.7/41.4    & 78.8/77.6    \\ 
Othtl-green               & 5753/27      & 523/3        & 1049/3       & 1213/6       & 599/2        & 1.4/2.8      & 1.2/0.0      & 0.1/0.0      & 9.3/0.0      & 22.0/3.6     & 41.7/50.0    & 7.8/2.1      & 17.4/16.7    \\ 
Ped-xing                  & 5508/130     & 337/8        & 145/2        & 171/4        & 583/19       & 17.2/14.5    & 1.1/1.3      & 0.0/0.0      & 3.0/7.0      & 0.5/1.2      & 1.6/1.8      & 5.9/5.2      & 1.9/3.4      \\ 
Car-incatrht              & 4686/86      & 288/8        & 344/6        & 411/9        & 582/10       & 19.7/16.8    & 6.5/15.9     & 5.8/19.3     & 5.2/4.5      & 2.0/1.6      & 4.4/4.9      & 9.1/12.5     & 5.4/8.4      \\ 
Car-turrht                & 4444/123     & 317/11       & 335/7        & 252/6        & 456/19       & 20.6/11.6    & 12.8/12.7    & 30.9/40.1    & 5.6/5.6      & 10.8/17.0    & 12.8/14.5    & 20.8/22.9    & 10.4/11.0    \\ 
Bus-stop                  & 4419/37      & 675/6        & 201/2        & 484/4        & 399/3        & 28.6/71.8    & 11.8/40.9    & 21.5/18.0    & 8.2/37.2     & 11.0/17.1    & 16.0/47.8    & 20.4/35.6    & 12.0/42.0    \\ 
Larveh-stop               & 4131/32      & 272/4        & 497/1        & 1431/12      & 1144/6       & 0.7/1.5      & 1.3/1.9      & 0.0/0.0      & 2.9/2.8      & 13.2/28.9    & 1.0/9.3      & 4.6/10.1     & 1.8/4.7      \\ 
Car-xingfmrht             & 4069/117     & 658/15       & 95/5         & 393/11       & 626/19       & 65.7/61.0    & 72.0/63.1    & 49.9/71.7    & 64.2/71.7    & 46.8/62.4    & 60.6/73.2    & 54.1/65.0    & 65.6/69.3    \\ 
Car-incatlft              & 3715/61      & 515/5        & 341/7        & 674/17       & 1567/6       & 6.3/21.5     & 6.9/26.5     & 2.5/4.1      & 24.4/19.6    & 10.8/1.3     & 20.0/27.2    & 6.5/9.0      & 17.1/24.4    \\ 
Bus-movtow                & 3556/49      & 283/4        & 223/4        & 646/9        & 224/4        & 39.4/70.8    & 48.7/55.0    & 5.6/14.6     & 54.8/50.6    & 52.0/69.8    & 66.4/51.0    & 32.3/51.7    & 56.6/52.2    \\ 
Bus-movaway               & 3365/28      & 623/5        & 786/2        & 739/9        & 73/3         & 19.7/34.0    & 0.2/0.0      & 4.9/79.2     & 6.2/0.0      & 20.9/10.5    & 1.5/0.0      & 15.1/41.2    & 2.6/0.0      \\ 
Car-turlft                & 2994/91      & 321/12       & 667/13       & 301/10       & 418/17       & 13.3/2.3     & 10.2/4.8     & 13.3/32.4    & 12.2/7.4     & 6.9/2.2      & 11.1/9.4     & 11.2/12.3    & 11.2/7.2     \\ 
Cyc-xingfmrht             & 2458/60      & 161/7        & 155/4        & 63/2         & 544/17       & 20.5/12.4    & 63.6/59.3    & 8.6/0.0      & 61.9/57.4    & 20.6/50.0    & 64.2/56.7    & 16.6/20.8    & 63.2/57.8    \\ 
Cyc-xingfmlft             & 1776/41      & 333/10       & 181/3        & 70/1         & 395/9        & 23.4/19.5    & 26.3/32.3    & 20.3/33.3    & 34.5/19.5    & 2.8/0.0      & 38.9/60.2    & 15.5/17.6    & 33.2/37.4    \\ 
Medveh-incatlft           & 1756/13      & 1355/4       & 73/1         & 42/4         & 62/1         & 1.3/25.0     & 14.2/0.0     & 0.2/1.2      & 36.0/0.0     & 0.8/0.0      & 29.1/0.0     & 0.8/8.7      & 26.5/0.0     \\ 
Cyc-turlft                & 1463/49      & 172/7        & 377/9        & 64/5         & 248/10       & 2.3/0.4      & 1.0/0.2      & 0.0/0.0      & 0.6/0.0      & 0.0/0.0      & 1.3/2.6      & 0.8/0.1      & 1.0/1.0      \\ 
Medveh-turrht             & 1075/34      & 317/11       & 29/1         & 57/1         & 197/6        & 6.0/8.1      & 2.6/2.9      & 0.6/1.2      & 11.8/19.5    & 0.1/1.4      & 16.1/18.2    & 2.2/3.6      & 10.2/13.5    \\ 
Bus-xingfmlft             & 851/16       & 277/6        & 110/2        & 123/4        & 43/1         & 27.8/38.2    & 2.1/5.0      & 25.1/41.7    & 5.8/25.0     & 0.6/0.7      & 4.9/16.7     & 17.9/26.9    & 4.3/15.6     
\\
\midrule 
\textbf{Total/Mean} & 603274/9335  & 60000/965    & 85730/1032   & 88960/1050   & 89080/1471   & 28.2/25.3    & 28.7/23.4    & 13.6/17.3    & 31.4/24.8    & 23.9/21.6    & 33.0/28.4    & 21.9/21.4    & 31.0/25.5    \\ 
 \bottomrule
  \end{tabular}
  }
  }
\label{tab:classwise-splitwise-duplex} 
\end{table*}

\begin{table*}[t]
\centering
\setlength{\tabcolsep}{4pt}
\caption{Number of video- and frame-level instances of road events (left).
Corresponding frame-/video-level results (mAP@$\%$) for each of the three ROAD splits (right). {Val-$n$ denotes the validation set for Split $n$}. The results presented here are generated using the I3D backbone.} 
{\footnotesize
\scalebox{0.75}{
\begin{tabular}{lccccc | cccccc cc}
\toprule
& \multicolumn{5}{c}{\textbf{Number of instances}}  & \multicolumn{8}{c}{\textbf{Frame-mAP@$0.5$/Video-mAP@$0.2$}} 
\\ 
\midrule
& \multicolumn{5}{c}{\#Boxes/\#Tubes} & \multicolumn{2}{c}{Train-1}  & \multicolumn{2}{c}{Train-2}  & \multicolumn{2}{c}{Train-3} &  \multicolumn{2}{c}{Averaged}\\ 
\midrule 
\textbf{Class \textbackslash Eval subset} & \em All & \em Val-1 & \em Val-2  & \em Val-3 & \em Test  & \em Val-1 & \em Test  & \em Val-2 & \em Test  & \em Val-3 & \em Test & \em Val & \em Test 
\\ 
\midrule
Ped-movaway-lftpav        & 44539/750    & 4063/72      & 5389/68      & 8328/74      & 4689/90      & 51.0/39.6    & 38.5/25.8    & 19.2/18.5    & 40.1/31.0    & 19.4/24.4    & 39.3/30.5    & 29.9/27.5    & 39.3/29.1    \\ 
Ped-movtow-lftpav         & 43711/905    & 3601/85      & 7527/120     & 4637/81      & 3996/99      & 51.2/43.6    & 33.9/25.0    & 22.9/21.0    & 37.9/32.7    & 25.4/30.5    & 39.7/29.1    & 33.2/31.7    & 37.2/29.0    \\ 
Ped-movtow-rhtpav         & 34452/844    & 3234/79      & 5168/108     & 5090/73      & 3603/112     & 45.4/35.4    & 23.1/26.7    & 13.8/13.5    & 23.7/26.6    & 26.9/29.5    & 22.8/22.1    & 28.7/26.2    & 23.2/25.2    \\ 
Ped-movaway-rhtpav        & 31203/711    & 2871/70      & 3600/66      & 2822/51      & 4807/107     & 41.3/38.8    & 27.3/17.8    & 4.8/6.7      & 25.5/17.1    & 10.5/13.1    & 22.0/13.2    & 18.9/19.5    & 24.9/16.0    \\ 
Car-movaway-vehlane       & 27319/179    & 721/11       & 4814/25      & 8008/27      & 4106/29      & 14.4/12.0    & 65.0/25.3    & 17.0/25.6    & 47.7/19.6    & 62.7/28.5    & 61.6/24.3    & 31.4/22.0    & 58.1/23.0    \\ 
Car-movtow-incomlane      & 25237/543    & 2717/60      & 4200/74      & 3603/67      & 3573/89      & 69.1/76.6    & 68.1/77.1    & 25.1/42.9    & 60.6/66.3    & 68.0/83.7    & 66.2/74.4    & 54.1/67.7    & 65.0/72.6    \\ 
Car-stop-vehlane          & 17083/92     & 1038/5       & 2158/10      & 5257/21      & 3577/12      & 28.3/33.1    & 71.9/36.4    & 14.3/4.9     & 28.0/24.0    & 44.7/19.6    & 53.2/41.8    & 29.1/19.2    & 51.1/34.1    \\ 
Car-stop-incomlane        & 16995/247    & 2614/29      & 6392/86      & 2546/36      & 1281/19      & 66.3/74.2    & 27.8/32.8    & 21.9/46.3    & 42.8/54.0    & 38.9/57.7    & 25.4/35.5    & 42.3/59.4    & 32.0/40.7    \\ 
Car-brake-vehlane         & 15494/108    & 490/10       & 2818/19      & 5425/30      & 2230/8       & 13.0/1.8     & 36.1/19.3    & 11.4/3.1     & 40.0/38.3    & 35.4/13.0    & 28.4/43.2    & 19.9/6.0     & 34.8/33.6    \\ 
Cyc-movaway-vehlane       & 13103/91     & 1226/7       & 2417/21      & 1671/13      & 1414/11      & 83.9/47.6    & 31.6/44.1    & 36.2/22.5    & 37.4/42.5    & 25.9/6.0     & 40.3/45.2    & 48.7/25.4    & 36.4/43.9    \\ 
Ped-stop-lftpav           & 11149/248    & 901/25       & 2263/30      & 1726/23      & 1054/33      & 14.0/9.4     & 6.7/3.5      & 5.7/9.7      & 5.2/5.2      & 17.7/18.6    & 5.9/5.4      & 12.4/12.6    & 5.9/4.7      \\ 
Car-stop-jun              & 10641/98     & 1192/15      & 1214/8       & 2178/18      & 1274/7       & 1.0/3.3      & 1.9/1.8      & 0.6/12.5     & 0.8/0.6      & 12.0/2.9     & 3.8/4.8      & 4.5/6.3      & 2.2/2.4      \\ 
Cyc-movaway-outgocyclane  & 9957/99      & 662/6        & 1266/13      & 1019/14      & 2388/30      & 38.6/25.0    & 50.0/32.2    & 4.7/11.4     & 50.0/43.0    & 39.3/17.8    & 50.4/45.9    & 27.6/18.1    & 50.2/40.4    \\ 
Car-movaway-jun           & 9770/236     & 945/21       & 813/14       & 1991/44      & 1454/53      & 14.7/13.0    & 12.4/16.0    & 5.1/11.1     & 14.0/18.1    & 28.0/17.2    & 14.0/16.7    & 15.9/13.8    & 13.5/16.9    \\ 
Ped-mov-pav               & 9549/227     & 1461/37      & 1435/17      & 694/22       & 2353/77      & 32.6/17.7    & 35.8/27.0    & 22.5/18.0    & 42.3/32.9    & 2.8/7.7      & 37.2/35.0    & 19.3/14.4    & 38.4/31.6    \\ 
Ped-stop-rhtpav           & 9335/222     & 884/23       & 423/15       & 1610/27      & 2768/48      & 2.4/2.3      & 2.1/4.5      & 0.4/0.3      & 4.2/10.9     & 6.0/11.6     & 3.9/8.1      & 2.9/4.7      & 3.4/7.8      \\ 
Cyc-movtow-incomlane      & 7031/208     & 1253/38      & 845/18       & 577/17       & 645/18       & 58.7/63.5    & 13.4/18.3    & 31.8/48.5    & 13.0/26.0    & 15.5/15.5    & 21.5/15.6    & 35.3/42.5    & 16.0/20.0    \\ 
Ped-xingfmlft-xing        & 6828/119     & 142/2        & 316/5        & 305/12       & 1105/18      & 9.5/4.2      & 31.1/42.9    & 4.2/32.8     & 40.0/44.5    & 10.8/9.2     & 45.3/45.0    & 8.2/15.4     & 38.8/44.1    \\ 
Car-movtow-jun            & 6529/238     & 339/15       & 479/16       & 339/19       & 1349/58      & 4.3/6.3      & 7.5/6.6      & 0.1/0.2      & 7.5/4.6      & 6.3/8.6      & 9.4/6.9      & 3.6/5.0      & 8.1/6.0      \\ 
Car-xingfmlft-jun         & 5953/157     & 625/18       & 124/7        & 827/31       & 1387/37      & 50.0/49.7    & 75.6/73.1    & 8.6/21.8     & 77.4/73.4    & 48.9/34.0    & 81.4/79.8    & 35.8/35.1    & 78.1/75.5    \\ 
Cyc-movtow-incomcyclane   & 5487/140     & 280/10       & 180/5        & 1002/23      & 1479/37      & 29.1/23.9    & 54.3/68.1    & 0.3/1.2      & 40.5/41.8    & 38.3/52.1    & 48.8/58.5    & 22.6/25.7    & 47.9/56.1    \\ 
Medveh-movtow-incomlane   & 5386/101     & 1149/25      & 246/4        & 908/16       & 1019/18      & 46.8/67.4    & 43.4/63.1    & 2.9/25.0     & 42.6/59.5    & 51.0/71.1    & 50.9/68.2    & 33.6/54.5    & 45.6/63.6    \\ 
Cyc-stop-jun              & 5383/49      & 977/7        & 122/1        & 174/5        & 1568/16      & 17.6/6.0     & 55.4/7.0     & 0.0/0.0      & 53.2/3.2     & 1.1/3.6      & 66.4/3.2     & 6.2/3.2      & 58.4/4.4     \\ 
Cyc-movaway-jun           & 5221/123     & 532/11       & 1221/15      & 185/7        & 772/32       & 13.4/16.7    & 5.6/2.7      & 6.0/7.8      & 10.6/7.1     & 0.2/0.0      & 14.3/8.2     & 6.6/8.2      & 10.2/6.0     \\ 
Ped-wait2x-lftpav         & 5165/86      & 580/8        & 1630/21      & 900/8        & 301/5        & 9.0/16.0     & 1.5/2.8      & 2.6/1.4      & 0.7/3.3      & 18.9/30.6    & 2.2/8.1      & 10.2/16.0    & 1.5/4.8      \\ 
Ped-stop-pav              & 4059/69      & 397/10       & 373/4        & 236/7        & 459/8        & 16.1/21.1    & 3.9/0.0      & 0.6/3.1      & 0.4/0.0      & 0.1/0.0      & 1.1/1.6      & 5.6/8.1      & 1.8/0.5      \\ 
Medveh-stop-incomlane     & 4004/44      & 1217/9       & 930/9        & 602/8        & 399/6        & 20.2/48.8    & 25.6/62.0    & 0.8/4.4      & 36.1/80.4    & 27.1/47.0    & 41.9/76.9    & 16.0/33.4    & 34.6/73.1    \\ 
Car-turrht-jun            & 3905/115     & 305/11       & 277/6        & 198/5        & 344/16       & 10.7/4.4     & 7.2/5.9      & 15.9/13.3    & 4.1/2.8      & 9.0/20.4     & 10.1/9.6     & 11.9/12.7    & 7.2/6.1      \\ 
Car-incatrht-jun          & 3110/65      & 107/3        & 231/5        & 128/6        & 297/7        & 37.4/33.8    & 16.5/25.9    & 12.2/25.3    & 13.0/11.2    & 1.4/0.6      & 11.5/27.1    & 17.0/19.9    & 13.7/21.4    \\ 
Car-movaway-outgolane     & 3037/67      & 82/4         & 129/5        & 619/12       & 189/9        & 0.3/0.0      & 0.5/0.5      & 0.0/0.0      & 0.4/0.0      & 4.6/6.5      & 0.2/0.4      & 1.6/2.2      & 0.4/0.3      \\ 
Bus-movtow-incomlane      & 2858/37      & 270/3        & 167/4        & 544/7        & 204/3        & 36.6/85.0    & 47.0/66.7    & 3.5/21.6     & 52.5/66.7    & 55.0/78.5    & 62.8/66.7    & 31.7/61.7    & 54.1/66.7    \\ 
Ped-pushobj-lftpav        & 2745/41      & 439/7        & 528/6        & 198/3        & 509/7        & 5.4/19.2     & 10.5/25.7    & 0.2/0.3      & 14.3/18.6    & 5.0/33.3     & 12.5/22.2    & 3.5/17.6     & 12.4/22.2    \\ 
Car-brake-jun             & 2679/37      & 348/7        & 48/2         & 1143/8       & 18/2         & 4.6/14.3     & 0.0/0.0      & 0.0/0.0      & 0.0/0.0      & 4.7/0.0      & 0.0/0.0      & 3.1/4.8      & 0.0/0.0      \\ 
Ped-movaway-pav           & 2662/100     & 23/4         & 330/10       & 659/22       & 385/15       & 0.2/0.0      & 0.3/0.2      & 0.0/0.0      & 0.3/0.1      & 0.1/0.2      & 0.4/1.1      & 0.1/0.1      & 0.3/0.5      \\ 
Cyc-movtow-jun            & 2472/83      & 308/7        & 99/2         & 106/5        & 546/19       & 11.3/12.8    & 6.9/6.3      & 0.4/1.4      & 2.8/4.1      & 4.3/6.3      & 4.6/7.0      & 5.3/6.8      & 4.8/5.8      \\ 
Car-turlft-jun            & 2413/85      & 283/12       & 602/12       & 268/10       & 348/18       & 11.5/1.7     & 6.6/0.8      & 8.8/22.0     & 7.6/3.0      & 5.9/2.9      & 11.7/8.3     & 8.7/8.8      & 8.6/4.0      \\ 
Car-incatlft-vehlane      & 2213/22      & 302/2        & 87/4         & 177/7        & 1454/3       & 7.2/0.0      & 6.3/4.2      & 0.1/6.2      & 24.2/33.3    & 2.4/0.0      & 17.7/15.9    & 3.2/2.1      & 16.0/17.8    \\ 
Ped-wait2x-rhtpav         & 2045/58      & 66/4         & 288/10       & 680/17       & 218/2        & 2.5/0.2      & 5.0/50.0     & 0.0/0.0      & 0.1/0.0      & 5.7/6.6      & 0.3/0.4      & 2.8/2.3      & 1.8/16.8     \\ 
Medveh-stop-jun           & 1940/18      & 983/6        & 601/3        & 28/1         & 41/2         & 0.6/1.0      & 0.0/0.0      & 0.0/0.0      & 0.0/0.0      & 0.0/0.0      & 0.0/0.0      & 0.2/0.3      & 0.0/0.0      \\ 
Ped-xingfmlft-jun         & 1933/26      & 341/5        & 62/2         & 54/1         & 76/2         & 12.4/5.0     & 0.8/1.7      & 10.0/52.2    & 0.3/0.0      & 8.9/100.0    & 1.1/1.9      & 10.4/52.4    & 0.7/1.2      \\ 
Car-incatlft-jun          & 1843/36      & 122/2        & 187/4        & 531/11       & 716/7        & 8.6/0.0      & 2.2/0.6      & 4.3/26.7     & 2.2/0.6      & 8.7/3.9      & 2.5/2.4      & 7.2/10.2     & 2.3/1.2      \\ 
Ped-stop-busstop          & 1746/48      & 19/1         & 124/3        & 465/6        & 174/6        & 0.0/0.2      & 0.6/0.4      & 0.7/5.6      & 0.9/2.1      & 0.3/0.5      & 1.3/2.1      & 0.3/2.1      & 1.0/1.6      \\ 
Ped-pushobj-rhtpav        & 1637/46      & 125/2        & 134/4        & 85/2         & 709/19       & 0.1/0.1      & 1.3/1.7      & 0.1/0.1      & 1.4/1.8      & 0.1/0.0      & 3.8/17.7     & 0.1/0.1      & 2.1/7.1      \\ 
Ped-xingfmlft-vehlane     & 1386/36      & 95/3         & 571/17       & 71/1         & 77/2         & 15.5/8.4     & 8.4/0.0      & 16.6/7.5     & 5.1/4.2      & 46.2/100.0   & 5.6/0.0      & 26.1/38.6    & 6.4/1.4      \\ 
Cyc-stop-incomlane        & 1376/30      & 424/8        & 74/2         & 485/10       & 20/1         & 17.9/42.1    & 0.3/0.0      & 0.1/0.0      & 0.4/0.0      & 10.1/21.5    & 0.2/0.0      & 9.3/21.2     & 0.3/0.0      \\ 
Bus-stop-vehlane          & 1351/9       & 311/2        & 157/1        & 182/2        & 305/2        & 37.8/100.0   & 9.8/12.5     & 29.2/100.0   & 11.5/100.0   & 10.3/50.0    & 1.9/10.0     & 25.8/83.3    & 7.7/40.8     \\ 
Cyc-xingfmlft-jun         & 1347/33      & 333/10       & 59/1         & 70/1         & 345/8        & 28.1/20.1    & 11.3/38.0    & 0.0/0.0      & 20.7/21.6    & 1.2/0.0      & 27.0/51.4    & 9.8/6.7      & 19.7/37.0    \\ 
Cyc-movaway-lftpav        & 1318/14      & 14/1         & 688/3        & 478/5        & 85/2         & 0.7/0.0      & 0.2/0.0      & 0.0/0.0      & 0.8/0.0      & 0.2/0.8      & 1.4/0.0      & 0.3/0.3      & 0.8/0.0      \\ 
Ped-xingfmlft-incomlane   & 1283/35      & 27/1         & 612/16       & 83/1         & 65/2         & 3.6/6.2      & 5.9/0.0      & 9.9/7.5      & 4.2/2.8      & 0.0/0.0      & 10.3/0.0     & 4.5/4.6      & 6.8/0.9      \\ 
Bus-stop-incomlane        & 1267/16      & 91/1         & 44/1         & 254/2        & 94/1         & 14.3/25.0    & 29.5/25.0    & 4.4/6.2      & 42.7/100.0   & 7.6/10.6     & 79.8/100.0   & 8.8/14.0     & 50.6/75.0    \\ 
Cyc-turlft-jun            & 1211/43      & 149/6        & 338/8        & 64/5         & 189/8        & 1.4/0.0      & 1.1/0.0      & 0.0/0.0      & 0.3/0.6      & 0.0/0.0      & 0.6/3.5      & 0.5/0.0      & 0.7/1.4      \\ 
Medveh-movtow-jun         & 1176/42      & 379/14       & 105/3        & 30/3         & 254/8        & 8.1/16.7     & 1.1/2.7      & 0.3/0.0      & 3.8/19.1     & 1.6/1.7      & 4.8/8.1      & 3.3/6.1      & 3.2/10.0     \\ 
Ped-stop-vehlane          & 1171/16      & 84/2         & 233/4        & 151/1        & 501/4        & 0.2/2.0      & 3.9/3.1      & 0.0/0.0      & 16.3/25.0    & 0.9/0.0      & 1.8/0.4      & 0.4/0.7      & 7.3/9.5      \\ 
Ped-xingfmrht-jun         & 1135/28      & 52/3         & 63/2         & 56/1         & 144/3        & 7.5/8.3      & 1.6/0.0      & 0.0/0.0      & 0.5/0.0      & 0.2/1.2      & 1.3/1.0      & 2.6/3.2      & 1.1/0.3      \\ 
Car-incatrht-incomlane    & 1096/26      & 181/7        & 44/3         & 193/3        & 44/2         & 10.3/11.1    & 0.5/0.0      & 0.1/0.1      & 4.5/0.0      & 4.4/12.6     & 0.5/0.0      & 4.9/8.0      & 1.8/0.0      \\ 
Ped-xingfmrht-vehlane     & 1037/24      & 185/3        & 86/3         & 108/2        & 156/5        & 1.1/3.0      & 3.7/4.5      & 4.9/33.3     & 0.8/0.2      & 0.2/0.4      & 3.1/5.4      & 2.1/12.3     & 2.6/3.4      \\ 
Car-turlft-vehlane        & 913/28       & 33/2         & 86/2         & 84/3         & 105/3        & 24.1/0.0     & 26.2/2.4     & 4.9/8.3      & 30.9/1.3     & 3.0/0.0      & 12.5/0.6     & 10.6/2.8     & 23.2/1.4     \\ 
Medveh-turrht-jun         & 898/32       & 293/11       & 23/1         & 51/1         & 114/6        & 5.9/9.0      & 6.9/5.2      & 1.1/2.3      & 14.5/18.0    & 0.1/1.4      & 21.0/18.1    & 2.4/4.2      & 14.1/13.8    \\ 
Bus-xingfmlft-jun         & 851/16       & 277/6        & 110/2        & 123/4        & 43/1         & 28.0/32.7    & 2.8/5.6      & 21.2/41.7    & 5.3/25.0     & 0.8/0.8      & 4.5/8.3      & 16.7/25.1    & 4.2/13.0     \\ 
Cyc-movaway-outgolane     & 788/24       & 38/1         & 6/1          & 216/5        & 71/5         & 0.0/0.0      & 0.7/0.0      & 0.0/0.0      & 1.7/0.0      & 8.4/2.0      & 1.2/0.0      & 2.8/0.7      & 1.2/0.0      \\ 
Bus-movtow-jun            & 751/18       & 50/1         & 56/1         & 194/4        & 47/2         & 0.2/0.0      & 2.4/21.2     & 0.8/7.1      & 4.6/13.2     & 10.0/24.4    & 14.1/51.1    & 3.6/10.5     & 7.0/28.5     \\ 
Cyc-movtow-lftpav         & 699/16       & 47/2         & 43/1         & 240/3        & 27/3         & 3.4/5.0      & 0.8/1.0      & 0.0/0.0      & 0.2/0.5      & 28.4/19.4    & 0.1/0.2      & 10.6/8.1     & 0.3/0.6      \\ 
Cyc-stop-incomcyclane     & 680/19       & 74/2         & 90/2         & 73/3         & 73/3         & 0.9/0.0      & 0.1/0.0      & 0.0/0.0      & 0.2/0.0      & 0.8/1.9      & 0.1/0.0      & 0.6/0.6      & 0.1/0.0      \\ 
Ped-movtow-vehlane        & 595/15       & 85/2         & 19/1         & 67/1         & 61/2         & 0.0/0.0      & 0.2/0.0      & 0.0/0.0      & 0.1/0.2      & 0.2/1.0      & 0.1/0.0      & 0.1/0.3      & 0.1/0.1      \\ 
Ped-xingfmrht-incomlane   & 513/15       & 93/3         & 109/3        & 157/4        & 115/3        & 0.1/0.0      & 0.1/0.0      & 1.0/0.0      & 0.0/0.0      & 0.9/5.6      & 0.2/0.0      & 0.7/1.9      & 0.1/0.0      \\ 
Ped-movtow-incomlane      & 473/10       & 5/1          & 54/2         & 35/1         & 186/2        & 0.2/0.0      & 0.1/0.4      & 0.0/0.2      & 0.0/0.2      & 0.5/0.0      & 0.0/0.0      & 0.2/0.1      & 0.0/0.2      \\ 
Car-turrht-incomlane      & 295/22       & 12/2         & 14/1         & 54/1         & 73/6         & 0.3/0.0      & 0.7/0.0      & 0.0/0.0      & 0.3/0.0      & 0.0/0.0      & 0.8/0.0      & 0.1/0.0      & 0.6/0.0      \\ 
Ped-xingfmrht-rhtpav      & 195/9        & 3/1          & 24/1         & 90/3         & 73/3         & 0.0/0.0      & 0.0/0.0      & 0.0/0.0      & 0.0/0.0      & 0.0/0.0      & 0.0/0.0      & 0.0/0.0      & 0.0/0.0      \\ 
\midrule
\textbf{Total/Mean}                    & 453626/8394  & 43569/883    & 65965/963    & 72152/967    & 64545/1301   & 17.7/18.6    & 15.9/15.8    & 6.4/11.8     & 16.4/18.9    & 13.7/17.2    & 18.1/18.9    & 12.6/15.9    & 16.8/17.9    \\ 
 \bottomrule
  \end{tabular}
  }
  }
\label{tab:classwise-splitwise-triplets} 
\end{table*}

\begin{table}[t]
  \centering
  \setlength{\tabcolsep}{4pt}
  \caption{Comparison of joint vs product of marginals approaches for \emph{duplex} detection with I3D backbone.
  Number of video-/frame-level instances for each composite label ('No instances' column) and corresponding frame-/video-level results (mAP@$\%$) averaged across all three splits, on both validation and test sets.}
  {\footnotesize
  \scalebox{0.85}{
  \begin{tabular}{lccccc}
  \toprule
  & \multicolumn{1}{c}{\textbf{No instances}}  & \multicolumn{4}{c}{\textbf{Frame-mAP@$0.5$/Video-mAP@$0.2$}} \\ 
  \midrule
  Eval-method & \multicolumn{1}{c}{} & \multicolumn{2}{c}{Joint} & \multicolumn{2}{c}{Prod. of marginals}  \\ 
  \midrule 
  \textbf{Class \textbackslash Eval subset} & \em All & \em Val & \em Test  & \em Val & \em Test 
  \\ 
  \midrule
 Ped-movtow                & 85079/1866   & 30.2/28.7    & 28.2/25.2    & 30.8/28.8    & 27.8/24.3    \\ 
 Ped-movaway               & 84177/1574   & 29.9/24.8    & 33.7/21.8    & 30.3/24.0    & 34.4/22.2    \\ 
 Car-stop                  & 44975/435    & 40.5/34.3    & 42.2/27.9    & 42.3/34.8    & 43.2/28.9    \\ 
 Tl-red                    & 42912/131    & 55.5/14.6    & 67.9/25.1    & 55.1/14.8    & 69.4/26.6    \\ 
 Car-movaway               & 37398/340    & 31.6/20.7    & 52.5/25.2    & 32.7/21.5    & 53.2/26.1    \\ 
 Car-movtow                & 30134/677    & 52.5/66.5    & 61.2/60.6    & 54.0/66.8    & 60.2/59.9    \\ 
 Ped-stop                  & 29338/618    & 7.9/7.5      & 12.2/8.1     & 7.3/7.1      & 13.3/8.8     \\ 
 Cyc-movaway               & 27848/236    & 48.0/22.1    & 58.4/40.7    & 48.0/21.8    & 57.4/41.1    \\ 
 Tl-green                  & 20381/224    & 46.8/15.8    & 39.3/7.9     & 43.2/14.2    & 32.9/5.8     \\ 
 Car-brake                 & 18661/159    & 21.9/5.3     & 35.7/24.9    & 22.4/5.4     & 39.0/25.7    \\ 
 Cyc-stop                  & 16021/139    & 30.3/18.6    & 54.6/9.4     & 33.0/16.3    & 53.7/8.4     \\ 
 Cyc-movtow                & 15164/407    & 45.1/48.9    & 43.4/46.7    & 44.8/47.3    & 39.6/45.0    \\ 
 Othtl-red                 & 13962/55     & 3.7/4.4      & 36.1/18.7    & 3.0/3.6      & 44.8/17.6    \\ 
 Medveh-stop               & 13496/86     & 4.3/15.3     & 28.8/42.3    & 4.3/14.1     & 32.2/44.6    \\ 
 Ped-wait2x                & 12278/239    & 10.1/11.1    & 11.4/8.2     & 13.2/11.1    & 9.8/8.0      \\ 
 Ped-mov                   & 11896/309    & 18.3/12.9    & 34.3/27.6    & 16.6/11.7    & 29.0/23.2    \\ 
 Ped-xingfmlft             & 11537/185    & 17.9/14.0    & 49.0/41.2    & 20.5/15.3    & 48.3/37.3    \\ 
 Ped-xingfmrht             & 6936/134     & 17.8/21.5    & 24.1/22.3    & 16.0/19.9    & 25.8/21.9    \\ 
 Tl-amber                  & 6404/144     & 23.3/2.7     & 28.3/4.8     & 21.8/2.2     & 24.6/3.6     \\ 
 Medveh-movtow             & 6360/119     & 33.3/44.8    & 42.4/58.8    & 33.3/44.2    & 41.9/57.0    \\ 
 Ped-pushobj               & 6252/103     & 3.0/12.2     & 6.7/6.2      & 2.4/11.4     & 7.4/7.0      \\ 
 Car-xingfmlft             & 6046/161     & 40.7/41.4    & 78.8/77.6    & 37.2/39.2    & 78.8/75.1    \\ 
 Othtl-green               & 5753/27      & 7.8/2.1      & 17.4/16.7    & 7.6/2.8      & 28.6/16.7    \\ 
 Ped-xing                  & 5508/130     & 5.9/5.2      & 1.9/3.4      & 8.1/9.8      & 2.4/3.8      \\ 
 Car-incatrht              & 4686/86      & 9.1/12.5     & 5.4/8.4      & 9.7/13.1     & 5.4/6.5      \\ 
 Car-turrht                & 4444/123     & 20.8/22.9    & 10.4/11.0    & 20.4/21.4    & 11.1/10.1    \\ 
 Bus-stop                  & 4419/37      & 20.4/35.6    & 12.0/42.0    & 21.9/43.7    & 10.9/30.1    \\ 
 Larveh-stop               & 4131/32      & 4.6/10.1     & 1.8/4.7      & 5.8/10.2     & 5.4/6.2      \\ 
 Car-xingfmrht             & 4069/117     & 54.1/65.0    & 65.6/69.3    & 48.2/62.4    & 63.3/70.9    \\ 
 Car-incatlft              & 3715/61      & 6.5/9.0      & 17.1/24.4    & 7.0/8.8      & 14.0/24.3    \\ 
 Bus-movtow                & 3556/49      & 32.3/51.7    & 56.6/52.2    & 31.6/48.9    & 57.2/51.7    \\ 
 Bus-movaway               & 3365/28      & 15.1/41.2    & 2.6/0.0      & 14.4/39.8    & 0.6/0.0      \\ 
 Car-turlft                & 2994/91      & 11.2/12.3    & 11.2/7.2     & 10.7/9.3     & 11.7/6.6     \\ 
 Cyc-xingfmrht             & 2458/60      & 16.6/20.8    & 63.2/57.8    & 11.5/18.9    & 53.8/55.3    \\ 
 Cyc-xingfmlft             & 1776/41      & 15.5/17.6    & 33.2/37.4    & 8.3/14.0     & 21.1/24.5    \\ 
 Medveh-incatlft           & 1756/13      & 0.8/8.7      & 26.5/0.0     & 1.9/8.9      & 33.4/0.0     \\ 
 Cyc-turlft                & 1463/49      & 0.8/0.1      & 1.0/1.0      & 0.8/0.1      & 0.9/0.7      \\ 
 Medveh-turrht             & 1075/34      & 2.2/3.6      & 10.2/13.5    & 1.7/1.4      & 9.7/6.4      \\ 
 Bus-xingfmlft             & 851/16       & 17.9/26.9    & 4.3/15.6     & 20.3/39.4    & 5.7/14.6     \\ 
 \midrule
\textbf{Total/Mean} & 603274/9335  & 21.9/21.4    & 31.0/25.5    & 21.6/21.2    & 30.8/24.3    \\ 
 \bottomrule
  \end{tabular}
  }
  }
\label{tab:classwise-marginals-duplexes} 
\end{table}

\begin{table}[t]
  \centering
  \setlength{\tabcolsep}{4pt}
  \caption{Comparison of joint vs product of marginals approaches for \emph{road event} detection with I3D backbone.
  Number of video-/frame-level instances for each composite label ('No instances' column) and corresponding frame-/video-level results (mAP@$\%$) averaged across all three splits, on both validation and test sets.}
  {\scriptsize
  \scalebox{0.85}{
  \begin{tabular}{lccccc}
  \toprule
  & \multicolumn{1}{c}{\textbf{No instances}}  & \multicolumn{4}{c}{\textbf{Frame-mAP@$0.5$/Video-mAP@$0.2$}} \\ 
  \midrule
  Eval-method & \multicolumn{1}{c}{} & \multicolumn{2}{c}{Joint} & \multicolumn{2}{c}{Prod. of marginals}  \\ 
  \midrule 
  \textbf{Class \textbackslash Eval subset} & \em All & \em Val & \em Test  & \em Val & \em Test  \\ 
  \midrule
  Ped-movaway-lftpav        & 44539/750   & 29.9/27.5    & 39.3/29.1   & 30.4/26.5    & 43.0/30.4    \\ 
Ped-movtow-lftpav         & 43711/905   & 33.2/31.7    & 37.2/29.0   & 33.8/32.7    & 36.6/28.5    \\ 
Ped-movtow-rhtpav         & 34452/844   & 28.7/26.2    & 23.2/25.2   & 27.7/25.4    & 21.0/22.1    \\ 
Ped-movaway-rhtpav        & 31203/711   & 18.9/19.5    & 24.9/16.0   & 19.1/18.7    & 23.2/16.1    \\ 
Car-movaway-vehlane       & 27319/179   & 31.4/22.0    & 58.1/23.0   & 34.9/22.5    & 56.7/22.7    \\ 
Car-movtow-incomlane      & 25237/543   & 54.1/67.7    & 65.0/72.6   & 55.5/68.4    & 63.3/72.2    \\ 
Car-stop-vehlane          & 17083/92    & 29.1/19.2    & 51.1/34.1   & 33.1/18.5    & 51.6/35.6    \\ 
Car-stop-incomlane        & 16995/247   & 42.3/59.4    & 32.0/40.7   & 42.0/58.8    & 32.1/39.8    \\ 
Car-brake-vehlane         & 15494/108   & 19.9/6.0     & 34.8/33.6   & 23.4/5.2     & 42.0/33.5    \\ 
Cyc-movaway-vehlane       & 13103/91    & 48.7/25.4    & 36.4/43.9   & 51.8/26.3    & 29.0/43.5    \\ 
Ped-stop-lftpav           & 11149/248   & 12.4/12.6    & 5.9/4.7     & 10.9/10.5    & 5.3/4.9      \\ 
Car-stop-jun              & 10641/98    & 4.5/6.3      & 2.2/2.4     & 10.0/4.0     & 4.6/5.2      \\ 
Cyc-movaway-outgocyclane  & 9957/99     & 27.6/18.1    & 50.2/40.4   & 25.4/18.3    & 49.7/37.8    \\ 
Car-movaway-jun           & 9770/236    & 15.9/13.8    & 13.5/16.9   & 16.4/14.0    & 16.5/17.6    \\ 
Ped-mov-pav               & 9549/227    & 19.3/14.4    & 38.4/31.6   & 18.0/12.4    & 33.2/27.3    \\ 
Ped-stop-rhtpav           & 9335/222    & 2.9/4.7      & 3.4/7.8     & 3.7/5.1      & 6.0/8.3      \\ 
Cyc-movtow-incomlane      & 7031/208    & 35.3/42.5    & 16.0/20.0   & 36.2/42.6    & 12.8/17.1    \\ 
Ped-xingfmlft-xing        & 6828/119    & 8.2/15.4     & 38.8/44.1   & 11.0/13.2    & 38.7/40.9    \\ 
Car-movtow-jun            & 6529/238    & 3.6/5.0      & 8.1/6.0     & 3.9/4.8      & 8.2/5.8      \\ 
Car-xingfmlft-jun         & 5953/157    & 35.8/35.1    & 78.1/75.5   & 33.2/32.6    & 76.5/72.6    \\ 
Cyc-movtow-incomcyclane   & 5487/140    & 22.6/25.7    & 47.9/56.1   & 19.8/23.0    & 43.8/52.8    \\ 
Medveh-movtow-incomlane   & 5386/101    & 33.6/54.5    & 45.6/63.6   & 32.7/47.1    & 43.7/60.0    \\ 
Cyc-stop-jun              & 5383/49     & 6.2/3.2      & 58.4/4.4    & 12.2/7.3     & 57.4/3.9     \\ 
Cyc-movaway-jun           & 5221/123    & 6.6/8.2      & 10.2/6.0    & 10.0/9.5     & 10.1/6.2     \\ 
Ped-wait2x-lftpav         & 5165/86     & 10.2/16.0    & 1.5/4.8     & 18.3/20.5    & 1.7/5.7      \\ 
Ped-stop-pav              & 4059/69     & 5.6/8.1      & 1.8/0.5     & 4.8/8.5      & 1.3/0.3      \\ 
Medveh-stop-incomlane     & 4004/44     & 16.0/33.4    & 34.6/73.1   & 18.3/32.5    & 30.3/66.6    \\ 
Car-turrht-jun            & 3905/115    & 11.9/12.7    & 7.2/6.1     & 16.7/15.0    & 10.0/9.5     \\ 
Car-incatrht-jun          & 3110/65     & 17.0/19.9    & 13.7/21.4   & 21.4/26.1    & 15.0/21.4    \\ 
Car-movaway-outgolane     & 3037/67     & 1.6/2.2      & 0.4/0.3     & 2.4/4.3      & 0.5/0.4      \\ 
Bus-movtow-incomlane      & 2858/37     & 31.7/61.7    & 54.1/66.7   & 34.3/66.9    & 53.4/66.7    \\ 
Ped-pushobj-lftpav        & 2745/41     & 3.5/17.6     & 12.4/22.2   & 1.2/6.3      & 16.1/19.9    \\ 
Car-brake-jun             & 2679/37     & 3.1/4.8      & 0.0/0.0     & 8.0/4.8      & 0.1/0.0      \\ 
Ped-movaway-pav           & 2662/100    & 0.1/0.1      & 0.3/0.5     & 0.1/0.1      & 0.2/0.4      \\ 
Cyc-movtow-jun            & 2472/83     & 5.3/6.8      & 4.8/5.8     & 7.6/7.9      & 6.6/8.9      \\ 
Car-turlft-jun            & 2413/85     & 8.7/8.8      & 8.6/4.0     & 10.2/8.4     & 8.4/3.6      \\ 
Car-incatlft-vehlane      & 2213/22     & 3.2/2.1      & 16.0/17.8   & 7.5/25.0     & 15.5/15.7    \\ 
Ped-wait2x-rhtpav         & 2045/58     & 2.8/2.3      & 1.8/16.8    & 3.2/4.5      & 1.3/0.0      \\ 
Medveh-stop-jun           & 1940/18     & 0.2/0.3      & 0.0/0.0     & 0.2/0.9      & 0.0/0.0      \\ 
Ped-xingfmlft-jun         & 1933/26     & 10.4/52.4    & 0.7/1.2     & 11.0/22.2    & 0.5/0.6      \\ 
Car-incatlft-jun          & 1843/36     & 7.2/10.2     & 2.3/1.2     & 9.5/6.1      & 2.9/1.2      \\ 
Ped-stop-busstop          & 1746/48     & 0.3/2.1      & 1.0/1.6     & 0.2/0.4      & 4.0/9.0      \\ 
Ped-pushobj-rhtpav        & 1637/46     & 0.1/0.1      & 2.1/7.1     & 0.2/0.1      & 2.3/4.2      \\ 
Ped-xingfmlft-vehlane     & 1386/36     & 26.1/38.6    & 6.4/1.4     & 23.7/36.9    & 9.3/4.2      \\ 
Cyc-stop-incomlane        & 1376/30     & 9.3/21.2     & 0.3/0.0     & 6.8/10.5     & 1.0/0.0      \\ 
Bus-stop-vehlane          & 1351/9      & 25.8/83.3    & 7.7/40.8    & 27.6/58.3    & 4.0/10.7     \\ 
Cyc-xingfmlft-jun         & 1347/33     & 9.8/6.7      & 19.7/37.0   & 4.8/3.0      & 9.1/16.0     \\ 
Cyc-movaway-lftpav        & 1318/14     & 0.3/0.3      & 0.8/0.0     & 0.3/0.7      & 0.8/0.7      \\ 
Ped-xingfmlft-incomlane   & 1283/35     & 4.5/4.6      & 6.8/0.9     & 9.1/8.3      & 13.8/0.0     \\ 
Bus-stop-incomlane        & 1267/16     & 8.8/14.0     & 50.6/75.0   & 8.4/15.6     & 42.4/50.0    \\ 
Cyc-turlft-jun            & 1211/43     & 0.5/0.0      & 0.7/1.4     & 0.7/0.0      & 0.8/1.4      \\ 
Medveh-movtow-jun         & 1176/42     & 3.3/6.1      & 3.2/10.0    & 5.6/10.3     & 4.2/6.7      \\ 
Ped-stop-vehlane          & 1171/16     & 0.4/0.7      & 7.3/9.5     & 0.9/0.8      & 5.2/1.1      \\ 
Ped-xingfmrht-jun         & 1135/28     & 2.6/3.2      & 1.1/0.3     & 2.2/1.5      & 1.0/0.8      \\ 
Car-incatrht-incomlane    & 1096/26     & 4.9/8.0      & 1.8/0.0     & 2.6/3.4      & 0.9/0.0      \\ 
Ped-xingfmrht-vehlane     & 1037/24     & 2.1/12.3     & 2.6/3.4     & 4.6/12.9     & 4.6/4.6      \\ 
Car-turlft-vehlane        & 913/28      & 10.6/2.8     & 23.2/1.4    & 11.2/0.0     & 10.7/0.0     \\ 
Medveh-turrht-jun         & 898/32      & 2.4/4.2      & 14.1/13.8   & 1.7/1.8      & 8.2/11.4     \\ 
Bus-xingfmlft-jun         & 851/16      & 16.7/25.1    & 4.2/13.0    & 22.0/43.4    & 3.9/9.4      \\ 
Cyc-movaway-outgolane     & 788/24      & 2.8/0.7      & 1.2/0.0     & 2.6/1.6      & 1.7/0.0      \\ 
Bus-movtow-jun            & 751/18      & 3.6/10.5     & 7.0/28.5    & 3.6/9.6      & 6.4/39.4     \\ 
Cyc-movtow-lftpav         & 699/16      & 10.6/8.1     & 0.3/0.6     & 7.2/2.3      & 0.2/0.1      \\ 
Cyc-stop-incomcyclane     & 680/19      & 0.6/0.6      & 0.1/0.0     & 1.8/4.2      & 0.1/0.6      \\ 
Ped-movtow-vehlane        & 595/15      & 0.1/0.3      & 0.1/0.1     & 0.2/0.4      & 0.1/0.1      \\ 
Ped-xingfmrht-incomlane   & 513/15      & 0.7/1.9      & 0.1/0.0     & 9.4/12.5     & 0.4/0.0      \\ 
Ped-movtow-incomlane      & 473/10      & 0.2/0.1      & 0.0/0.2     & 1.0/0.1      & 0.0/0.0      \\ 
Car-turrht-incomlane      & 295/22      & 0.1/0.0      & 0.6/0.0     & 0.2/0.0      & 1.3/0.0      \\ 
Ped-xingfmrht-rhtpav      & 195/9       & 0.0/0.0      & 0.0/0.0     & 0.2/0.0      & 1.6/1.5      \\ 
 \midrule
\textbf{Total/Mean}                  & 453626/8394 & 12.6/15.9    & 16.8/17.9   & 13.7/15.4    & 16.3/16.1    \\ 
 \bottomrule
  \end{tabular}
  }
  }
\label{tab:classwise-marginals-event} 
\end{table}